\documentclass{article}

\PassOptionsToPackage{numbers}{natbib}

\usepackage[final]{neurips_2021}

\usepackage[utf8]{inputenc} 
\usepackage[T1]{fontenc}    
\usepackage{hyperref}       
\usepackage{url}            
\usepackage{booktabs}       
\usepackage{amsfonts}       
\usepackage{nicefrac}       
\usepackage{microtype}      

\usepackage{amsmath}
\usepackage{graphicx}
\usepackage{siunitx}
\usepackage[small]{caption}
\usepackage[dvipsnames]{xcolor}
\usepackage[inline]{enumitem}
\usepackage{wrapfig}
\usepackage{caption}
\usepackage{subcaption}
\usepackage[export]{adjustbox}
\usepackage[ruled,vlined]{algorithm2e}

\usepackage{floatrow}
\newfloatcommand{capbtabbox}{table}[][\FBwidth]

\newcommand{\GP}{\mathrm{GP}}

\newcommand{\gauss}{{\cal N}}

\newcommand{\IG}{\mathrm{IG}}

\newcommand{\myvec}[1]{\mathbf{#1}}
\newcommand{\myvecsym}[1]{\boldsymbol{#1}}

\newcommand{\vone}{\myvecsym{1}}

\newcommand{\vmu}{\myvecsym{\mu}}

\newcommand{\vLambda}{\myvecsym{\Lambda}}

\newcommand{\vSigma}{\myvecsym{\Sigma}}

\newcommand{\vrho}{\myvecsym{\rho}}

\newcommand{\vm}{\myvec{m}}

\newcommand{\vr}{\myvec{r}}

\newcommand{\vy}{\myvec{y}}

\newcommand{\vK}{\myvec{K}}

\newcommand{\calD}{{\cal D}}

\newcommand{\calX}{{\cal X}}

\newcommand{\be}{\begin{equation}}
\newcommand{\ee}{\end{equation}}
\newcommand{\bea}{\begin{eqnarray}}
\newcommand{\eea}{\end{eqnarray}}
\newcommand{\beaa}{\begin{eqnarray*}}
\newcommand{\eeaa}{\end{eqnarray*}}

\DeclareMathAlphabet{\mathpzc}{OT1}{pzc}{m}{n}

\newcommand{\dmcontrol}{{dm-control}}
\newcommand{\mpg}{{MPG}}
\newcommand{\OPE}{{\textsc{OPE}}} 
\newcommand{\IndUniform}{{\textsc{Ind+Uniform}}}
\newcommand{\IndUCB}{{\textsc{Ind+UCB}}}
\newcommand{\GPUniform}{{\textsc{GP+Uniform}}}
\newcommand{\GPUCB}{{\textsc{GP+UCB}}}
\newcommand{\IndUniformOPE}{{\textsc{Ind+Uniform+OPE}}}
\newcommand{\IndUCBOPE}{{\textsc{Ind+UCB+OPE}}}
\newcommand{\GPUniformOPE}{{\textsc{GP+Uniform+OPE}}}
\newcommand{\GPUCBOPE}{{\textsc{GP+UCB+OPE}}}
\newcommand{\ours}{{\textsc{A-ops}}}

\makeatletter
\newcommand{\printfnsymbol}[1]{%
  \textsuperscript{\@fnsymbol{#1}}%
}
\makeatother

\title{Active Offline Policy Selection}

\author{%
  Ksenia Konyushkova\thanks{equal contribution} \\
  DeepMind\\
  \texttt{kksenia@deepmind.com} \\
  \And
  Yutian Chen\printfnsymbol{1} \\
  DeepMind\\
  \texttt{yutianc@deepmind.com} \\
  \And
  Tom Le Paine \\
  DeepMind\\
  \texttt{tpaine@deepmind.com} \\
  \And
  Caglar Gulcehre \\
  DeepMind\\
  \texttt{caglarg@deepmind.com} \\
  \And
  Cosmin Paduraru \\
  DeepMind\\
  \texttt{paduraru@deepmind.com} \\
  \And
  Daniel J Mankowitz \\
  DeepMind\\
  \texttt{dmankowitz@deepmind.com} \\
  \And
  Misha Denil \\
  DeepMind\\
  \texttt{mdenil@deepmind.com} \\
  \And
  Nando de Freitas \\
  DeepMind\\
  \texttt{nandodefreitas@deepmind.com}
}

\begin{document}

\maketitle

\begin{abstract}

This paper addresses the problem of policy selection in domains with abundant logged data, but with a restricted interaction budget. 
Solving this problem would enable safe evaluation and deployment of offline reinforcement learning policies in industry, robotics, and recommendation domains among others. 
Several off-policy evaluation (OPE) techniques have been proposed to assess the value of policies using only logged data. 
However, there is still a big gap between the evaluation by OPE and the full online evaluation.
Yet, large amounts of online interactions are often not possible in practice.
To overcome this problem, we introduce \emph{active offline policy selection} --- a novel sequential decision approach that combines logged data with online interaction to identify the best policy.
We use OPE estimates to warm start the online evaluation. 
Then, in order to utilize the limited environment interactions wisely we decide which policy to evaluate next based on a Bayesian optimization method with a kernel that represents policy similarity. 
We use multiple benchmarks, including real-world robotics, with a large number of candidate policies to show that the proposed approach improves upon state-of-the-art OPE estimates and pure online policy evaluation \footnote{The paper website is at \url{https://sites.google.com/corp/view/active-ops} and the code is at \url{https://github.com/deepmind/active_ops}.}.

\end{abstract}
\section{Introduction}
\label{sec:intro}

\begin{figure}[t!]
    \centering
    \includegraphics[width=0.65\textwidth]{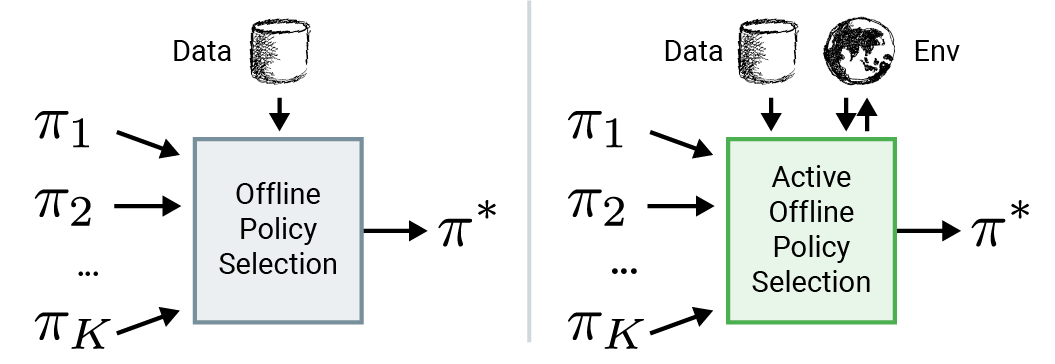}
    \caption{Left: Offline policy selection attempts to choose the best policy from a set of policies, given only a dataset of logged experiences. Right: \emph{Active offline policy selection} additionally assumes a restricted environment interaction budget which is assigned intelligently to evaluate the most promising policies. }
    \label{fig:aops}
\end{figure}

Reinforcement learning (RL) has recently proven to be successful in a range of applications from computer and board games \citep{dqn, muzero,alphago} to robotics and chip design \citep{levine2016end, cabi2020sketchy, chips}.
However, many challenges of real-world systems still prevent RL from being applied at scale in practice \citep{challenges}.
One of the limiting factors in real applications is that environment interactions are often expensive. 
Thus, training and evaluating agents becomes prohibitively slow and costly.
Offline RL attempts to address this problem by training agents on a dataset without environment interactions \citep{agarwal2019optimistic,gulcehre2020rl}.
This enables researchers to train policies with multiple algorithms and different hyperparameter settings \citep{paine2020hyperparameter} on a single dataset.
The best policy can then be chosen with off-policy policy evaluation (OPE) approaches \citep{Precup2000,li2011unbiased} (\autoref{fig:aops}, left). 
These techniques attempt to estimate the expected agent performance by relying on the same pre-recorded dataset \citep{jiang2016doubly,liu2018breaking,le2019batch,yang2020off}.

Unfortunately, OPE estimates are not yet precise enough to be relied upon for choosing which agent to deploy. 
Moreover, OPE is intrinsically limited by the pre-recorded dataset: 
When there is a distribution shift between a trained policy and a behavioural policy, {\it i.e.}, when the trained policy behaves very differently from the data collection policy, OPE is imprecise \citep{levine2020offline}. 
In many applications, however, a subset of policies can be evaluated \emph{online} prior to final deployment \citep{gulcehre2020rl,fu2020d4rl}, even though this process can be hard and laborious.
For example, in robotics a small subset of offline RL policies can be tested on the real hardware.
Another example is recommender systems, which allow for restricted policy evaluation via A/B tests on a small fraction of user traffic. 

In this paper, we introduce the \emph{active offline policy selection} problem.
The goal is to identify the best policy where we benefit from both offline evaluation with logged data and online evaluation with limited interaction (\autoref{fig:aops}, right).
In our approach we advance a solution based on Bayesian optimization (BO). 
It entails learning a Gaussian process (GP) surrogate function that maps policies to their expected returns. 
Then, we build on the GP statistics to construct an acquisition function to decide which policy to test next.
To make BO successful in this problem setting, our approach has two key features.
First, we incorporate existing OPE estimates as additional noisy observations (\autoref{fig:gp}).
This allows us to warm start online policy evaluation and to overcome the difficulties of GP hyper-parameter optimisation at the start. 
Second, we model correlation between policies through a kernel function based on actions that the policies take in the same states of the environment (\autoref{fig:kernel}).
This makes our method data efficient \citep{hoffman2014correlation} as the information about the performance of one policy informs us about the performance of similar behaving policies without costly execution in the environment. 
It is particularly valuable when the number of candidate policies is large (or even larger than the interaction budget).

The contributions of this paper are as follows.
\begin{enumerate}
    \item We introduce the \emph{active offline policy selection} problem. This problem is important in practice, but so far it has not been studied in the literature to the best of our knowledge.
    \item We build a BO solution with an extended observation model to integrate both OPE estimates and interaction data. The use of OPE estimates enables us to warm-start learning a GP.
    \item We propose a novel GP kernel to capture the dependency between the policies through the actions that they take.
    As a result, our method infers the value of one policy from those of similar policies.
\end{enumerate}

The rest of this paper is organized as follows. 
Section~\ref{sec:background} gives background on offline policy evaluation and selection.
In \autoref{sec:method}, we propose a GP model for policy value estimation, explain how we approach the sequential decision making with BO and introduce a novel kernel based on the similarity of policy outputs.
Section \ref{sec:relwork} discusses the related work.
Section \ref{sec:experiments} shows that active policy evaluation can improve upon the OPE after merely a few interactions, thanks to the kernel that ensures data-efficiency.
Additionally, our method works reliably with OPEs of varying quality and it scales well with the growing number of candidate policies. Finally, we demonstrate our method to be efficient in the application of real-world robotics.
The paper is concluded with a discussion in \autoref{sec:conclusion}.

\subsection{Off-policy policy evaluation and selection}
\label{sec:background}
We consider a Markov decision process (MDP) defined by a tuple $(S, A, T, R, d_0, \gamma)$, with state space $S$, action space $A$, transition distribution $T(s'|s, a)$, reward function $R(s, a)$, initial state distribution $d_0(s)$, and discount factor $\gamma \in [0, 1]$.
A policy $\pi(a|s)$ maps from $S$ to $A$. The value of a policy $\pi$ is measured by the expected sum of discounted rewards:
\begin{equation}
    \mu_{\pi} = \mathbb{E}\left[\sum_{t=0}^\infty \gamma^t R(s_t, a_t)\right], \text{ with } s_0\sim d_0(\cdot), a_t\sim \pi(\cdot|s_t), s_{t+1}\sim T(\cdot|s_t, a_t)\,.
\end{equation}

Typically, estimating the value of a given policy requires executing the policy in the environment many times.
However, practical constraints make it difficult to run many policies of unknown quality in the environment. 
To address this, OPE estimates a value $\hat{\mu}_{\pi}$ using a dataset of trajectories $\calD$ collected by a separate behavior policy $\pi_\beta$. 
In offline policy selection (OPS) the task is to select the policy with the highest value from a set of policies ${\pi_1, \pi_2, \dots, \pi_K}$ given access only to a dataset $\calD$ (\autoref{fig:aops}, left). 
A straightforward approach is to select the policy with the highest OPE estimate ${\hat{\mu}_{\pi_1},\dots,\hat{\mu}_{\pi_K}}$ \cite{fu2021benchmarks}, and alternative approaches can be used depending on the quality metric \citep{yang2020offline}.
\section{Active offline policy selection}
\label{sec:method}

\subsection{Problem definition}
\label{sec:a-ops}

We now formally define the active offline policy selection problem.
Suppose we are given a set of $K$ candidate policies $\pi_k, 1 \leq k \leq K$. Denote by $\mu_k$ the unknown policy value and by $\rho_k$ an OPE estimate computed from an offline dataset. 
At every step $i$, we can choose a policy $k_i$ to execute in the environment once and observe the trajectory with (discounted) episodic return $r_i$, which is a noisy sample of the true policy value: $\mathbb{E}[r_i] = \mu_{k_i}$. 
We would like to find the policy with the highest value, $k^* = \arg\max_k \mu_k$, with as few trajectories (budget) as possible. 
Policy selection algorithm estimates the mean as $m_k$ and recommends a policy that maximizes it: $\hat{k} = \arg\max_{k} m_k$.
Simple regret of the recommended policy measures how close we are to the best policy:
\begin{equation}
\label{eq:simple-regret}
    \mathrm{regret} = \mu_{k^*} - \mu_{\hat{k}}\,.
\end{equation}
To solve this sequential decision problem, we introduce a GP over the policy returns and OPE estimates (\autoref{sec:gp}), and design the GP kernel over policies to model their correlated returns (\autoref{sec:method:kernel}).
Then we use these as part of a BO strategy (\autoref{sec:method:bayesopt}).

\subsection{Gaussian process over policy values}
\label{sec:gp}

Given a limited interaction budget and a large number of policies, it is impossible to obtain accurate estimates of their values by executing them many times. 
Fortunately, similar policies tend to have similar values. 
Thus, propagating the information between them helps to improve the value prediction.
Intuitively, the number of policies that we need to execute to find a good policy should depend on their diversity instead of the total number of candidate policies (see experiments supporting this in  \autoref{fig:ope-num_policies-ablation}).
Formally, we use GP to predict a joint distribution over the policy values. 

\begin{figure}[t] 
    \centering
    \includegraphics[width=0.55\textwidth]{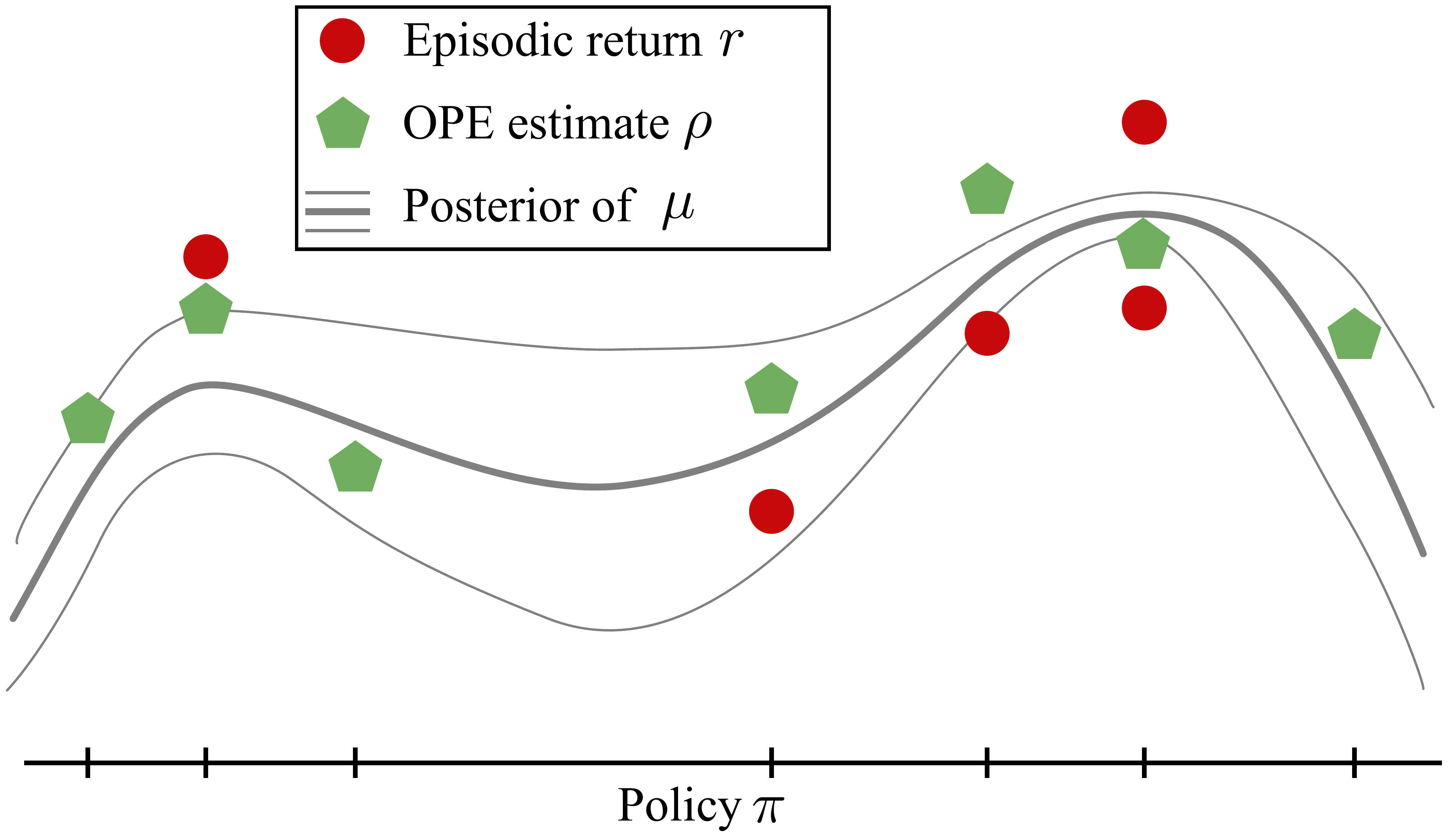}
    \caption{Gaussian process of the policy value function. 
    We observe OPE estimates for all policies and noisy episodic returns for some of them.
    Neighboring policies (measured by kernel $\mathcal{K}$) have similar values, and the posterior variance is lower where there are more observations. 
    Policies are aligned in \num{1}D for visualization purposes, but in practice they reside in a high dimensional space.}
    \label{fig:gp}
\end{figure}

A GP is a stochastic process, that is an indexed collection of random variables, such that every finite collection of them has a multivariate normal distribution \citep{Rasmussen2005}. In our case, we assume that the policy value $\mu$ as a function of policy $\pi$ follows a GP, with a kernel function $\mathcal{K}(\pi_i, \pi_j)$ over policies and mean $m$. 
Specifically, we consider the following generative process for the latent policy value $\mu(\pi)$, OPE estimate $\rho$, and episodic return $r$:
\begin{align}
    \mu(\pi) \sim \GP(m(\cdot), \mathcal{K}(\cdot, \cdot)),\quad
    \rho \sim \gauss(\mu(\pi), \sigma_\rho^2),\quad
    r \sim \gauss(\mu(\pi), \sigma_r^2).
\end{align}
We use a constant mean $m$ without loss of generality. We assume a flat prior for the hyper-parameter $m$, and weakly informative inverse Gamma prior for the variance $\sigma_\rho^2$ and $\sigma_r^2$.
Our probabilistic model is illustrated in \autoref{fig:gp}.

As we have a finite number of policies, the input space is a finite set $\mathcal{X}=\{\pi_1, \dots, \pi_K\}$, where each index corresponds to a candidate policy. The joint distribution of the $K$ policy values reduces to a multivariate normal distribution,
\begin{equation}
\mu_1, \dots, \mu_K \sim \gauss(m \vone_K, \vK), \text{ with }
\vK_{k,k'}=\mathcal{K}(\pi_k, \pi_{k'})\,,
\end{equation}
where $\vone_K$ is a $K$-length vector of ones. For each policy $\pi_k$, we have one noisy observation from an OPE method $\rho_k$, and zero or multiple noisy episodic return observations $r_k^i$. 
As the policies are related through the covariance matrix $\vK$, observing a return sample from one policy updates the posterior distribution of all of them. 
Given the OPE estimates $\vrho = [\rho_1, \dots, \rho_K]$ and $N_k$ return observations $\vr_k = [r_k^1, \dots, r_k^{N_k}]$ for each policy $\pi_k$, the posterior distribution of the mean return is also a Gaussian,
\begin{align}
&\mu_1, \dots, \mu_K | \vrho, \{\vr_k\}_{k} \sim \gauss(\vm, \vSigma), \text{ with } \nonumber\\
&\vm = \vK (\vK + \vLambda)^{-1} \vy,
\quad\quad
\vSigma = \vK - \vK (\vK + \vLambda)^{-1} \vK\nonumber. \\
&\vy_k = \left(\frac{1}{\sigma_\rho^2} + \frac{N_k}{\sigma_r^2}\right)^{-1} \left(\frac{\rho_k}{\sigma_\rho^2} + \frac{1}{\sigma_r^2}\sum_{i=1}^{N_k} r_k^i\right) ,
\quad\quad
\vLambda_{kk} = \left(\frac{1}{\sigma_\rho^2} + \frac{N_k}{\sigma_r^2}\right)^{-1},
\label{eq:gp_arm_post}
\end{align}
where $\vLambda$ is a diagonal matrix with diagonal entries defined above.

\subsection{Kernel}
\label{sec:method:kernel}

\begin{figure}[t!]
     \centering
     \begin{subfigure}[t]{0.4\textwidth}
         \centering
         \includegraphics[width=1\textwidth]{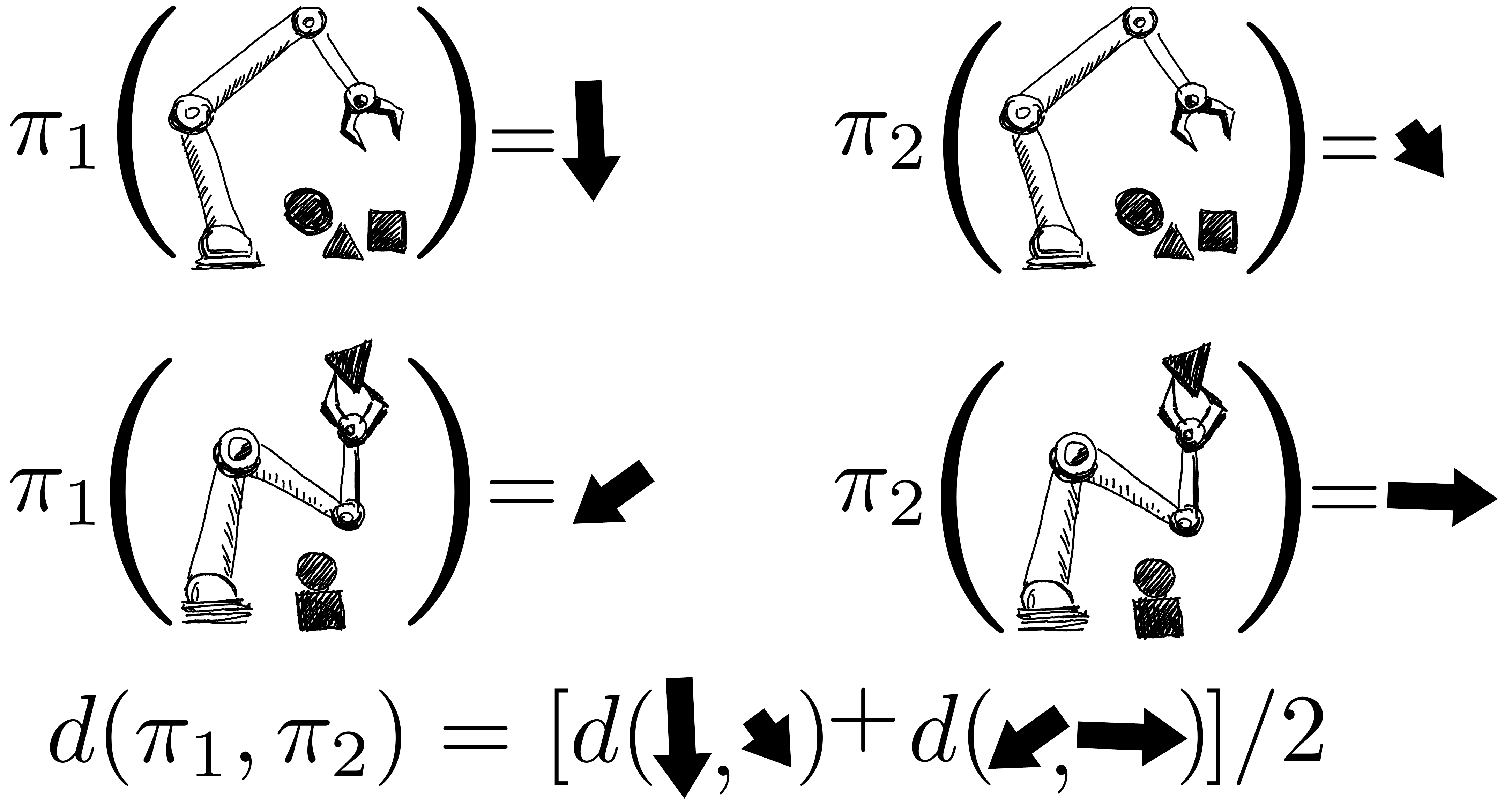}
     \end{subfigure}
     \hfill
     \begin{subfigure}[t]{0.20\textwidth}
         \centering
         \includegraphics[width=1\textwidth]{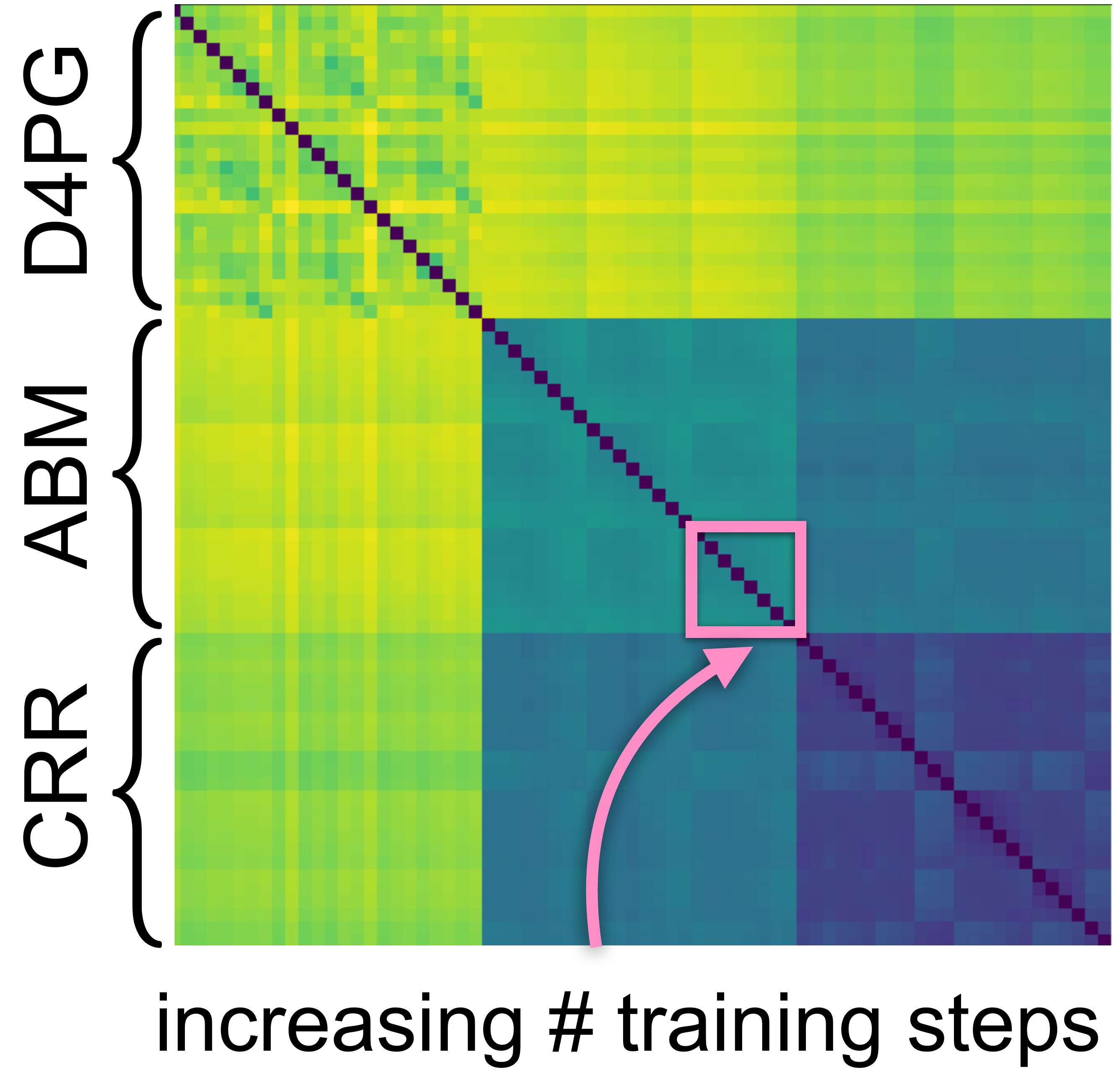}
     \end{subfigure}
     \hfill
     \begin{subfigure}[t]{0.35\textwidth}
         \centering
         \includegraphics[width=1\textwidth]{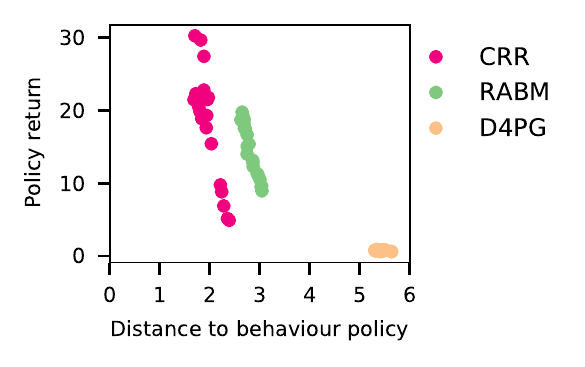}
     \end{subfigure}
        \caption{Left: The distance between two policies $\pi_1$ and $\pi_2$ is computed as an average distance across pairs of action vectors that are selected by each policy on a set of states. Middle: Pairwise distance matrix between policies on \texttt{humanoid-run} task. High distance is shown in yellow and low distance in dark blue. The $3\times 3$ block structure of the matrix reflects the $3$ types of training algorithm. Smaller blocks correspond to the increasing number of training steps with the same hyperparameters. 
        Right: Policy value ($\mu_\pi)$ versus action distance to the behavior policy $\bar{d}(\pi, \pi_\beta)$. Policies with the same training algorithm are in the same color.}
        \label{fig:kernel}
\end{figure}

A key component of the GP model is the kernel that measures our belief about the policy correlation.
To obtain a kernel we make a key assumption: 
\emph{Policies that are similar in the actions that they take, yield similar returns}.
Then, our insight is to measure the distance between the policies through the actions that each of them takes on a fixed set of states from the offline dataset (\autoref{fig:kernel}, left). 

Formally, to measure the similarity between $\pi_1$ and $\pi_2$, we obtain the policy actions at a given state $s$ and measure the distance between these two action vectors $d(\pi_1(s), \pi_2(s))$ \footnote{In this work, we use Euclidean distance for deterministic policies with continuous actions and Hamming distance for discrete actions. An extension to stochastic policies is straightforward.}. We then define the distance between the two policies as the average distance over the states of the offline dataset $\mathcal{D}$,
\begin{equation}
\bar{d}(\pi_1, \pi_2) = \mathbb{E}_{s \sim \mathcal{D}}\ d(\pi_1(s), \pi_2(s)) \,.
\end{equation}
In practice we approximate $\bar{d}$ using a randomly sampled subset of states $\mathcal{D}'\subset \mathcal{D}$.

An example of a pairwise distance matrix for a set of policies is shown in \autoref{fig:kernel}, middle.
Rows and columns are the policies which are arranged first by the type of the training algorithm, then by hyperparameters and finally by the training steps (none of which is observed by our method). 
Note the block structure, which reflects the different nature of training algorithms. 
Smaller blocks of increasing policy distances correspond to policies with more training steps and the same hyperparameters.

Additionally, \autoref{fig:kernel}, right shows an example of policy distances to the behaviour policy.
The values of policies produced by the same algorithm change gradually as a function of this distance. 
It provides indication that our kernel construction is informative for predicting the value of a policy.

Finally, we compute Mat\'ern $1/2$ kernel as: $\mathcal{K}(\pi_1, \pi_2) = \sigma_k^2 \exp (-\bar{d}(\pi_1, \pi_2) / l)$, where $\sigma$ and $l$ are the trainable variance and length-scale hyperparameters.
This is a popular kernel choice \cite{shahriari2015taking}.
As the distance metric is the average distance over states, using it in this kernel is equivalent to multiplication of kernels on these states and it is a valid positive semi-definite matrix.

\subsection{Active offline policy selection with Bayesian optimization}
\label{sec:method:bayesopt}

Under a GP formulation, we can employ BO~\citep{shahriari2015taking} to search for the best policy efficiently.
BO optimizes an unknown function $\mu(\pi)$ with $\pi\in\calX$ using a limited number of function queries that render noisy observations.
BO has been used successfully in many applications including experimental design \citep{azimi2012hybrid}, robotics \citep{lizotte2007automatic,Martinez-Cantin:2009}, hyper-parameter optimization \citep{Snoek:2012,Hutter:smac,chen2018bayesian}, preference learning and interactive machine learning \cite{Brochu:2007,Brochu:2010}.
The key component of a BO algorithm is an acquisition function $u_i(\pi)$ that balances exploration and exploitation when selecting a query. Widely used acquisition functions include upper confidence bounds (UCB), expected improvement (EI) and epsilon-greedy.
Here, we simply use the UCB acquisition function~\citep{srinivas2010gaussian,kaufmann2012bayesian}.\footnote{Experiments with other acquisition functions can be found in the supplementary materials.} Specifically, at every step $i$, we compute the score for a policy $\pi_k$ using
\begin{align}
    u_i(k) = \vm_k + \sqrt{\beta_i \vSigma_{kk}},
    \label{eq:aquisition}
\end{align}
where $\beta_i$ is a constant depending on $i$. We then choose the policy with the highest score, $k_i = \arg\max_k u_i(k)$, to execute next. 
After observing a new return $r_{k_i}$, we update $\vm$ and $\vSigma$ in \autoref{eq:gp_arm_post}. 
We also update the hyper-parameters ($m$, $\sigma_\rho^2$, $\sigma_r^2$, $\sigma_k^2$ and $l$) with the maximum a posteriori estimate (see details in the appendix).

Estimating the hyper-parameters with limited data is challenging in practice. 
Fortunately, in the active offline policy selection formulation, we can take advantage of the OPE estimates as prior observations to fit the hyper-parameters before launching the interactive optimization process. 
In some ablations where OPE is not available we instead execute \num{5} randomly sampled policies before fitting the hyperparameters.

\section{Related work}
\label{sec:relwork}
Offline RL becomes increasingly popular due to its ability to leverage offline data to learn policies for RL in the real world. 
As a results, several benchmark datasets were recently announced \citep{gulcehre2020rl,fu2020d4rl}. 
Offline RL methods include policy-constraint approaches that regularize the learned policy to stay close to the behavior policy \citep{wang2020critic, fujimoto2019off}, 
value-based approaches that encourage more conservative estimates, either through regularization or uncertainty \citep{kumar2020conservative,gulcehre2021regularized}, 
model-based approaches \citep{yu2020mopo,kidambi2020morel}, and 
standard RL methods used in off-policy RL \citep{barth2018distributed}.

OPE is studied extensively across many domains \citep{li2010contextual, theocharous2015personalized, kalashnikov2018qt, nie2019learning}. 
It includes methods that use importance sampling \citep{precup2000eligibility} or stationary state distribution \citep{liu2018breaking, nachum2019dualdice, uehara2019minimax}, value function \citep{sutton2009fast,migliavacca2010fitted,sutton2016emphatic}, or learned transition models \citep{zhang2021autoregressive}, as well as methods that combine these approaches \citep{munos2016safe, dudik2011doubly, jiang2016doubly, mrdr}.
OPS is a related practical problem that receives increasing interest \citep{paine2020hyperparameter, yang2020offline, fu2021benchmarks}. 
Our work has two notable distinctions in contrast to these prior works: 1) we consider similarities between policies when making value predictions, and 2) we focus on active setting, where a small number of evaluations in the environment can help to quickly improve the prediction of the best policy.

Similar to the way the kernel is computed based on policy's actions, Policy Evaluation Networks \citep{harb2020policy} work considers a network fingerprinting that learns a policy embedding through the outputs of the policy on probing states. 
The principal difference to our approach is that we only consider a very limited online policy evaluation budget (in the order of \num{100}) while network fingerprinting needs many policy evaluations to train the policy embeddings (in the order of \num{100000}). 

Applying BO in a finite set of indices reduces the problem to a Bayesian multiarmed bandit problem.
Exploring the shared structure among arms using Bayesian models has been studied in the literature \citep{srinivas2010gaussian,krause2012contextual,hoffman2014correlation}. 
Our work builds upon the Bayesian bandit model to address the problem of policy selection additionally exploiting the offline dataset.

Active learning is used in RL for better exploration, learning reward or world  models  \citep{epshteyn2008active,Akiyama2010,daniel2014active,krueger2020active, ball2020readypolicyone}. 
It has not been studied for the OPE or OPS problems to the best of our knowledge.
Related to the policy representation through actions in the kernel, classifiers in active learning can be also characterised by their predictions on a subset of datapoints \citep{konyushkova19discovering}. 
Furthermore, diversity in policies was measured through the action distribution for sampled states in population-based setting~\citep{parkerholder2020effective}.

\section{Experiments}
\label{sec:experiments}

\subsection{Datasets and environments}

\begin{figure}[t!]
    \centering
    \includegraphics[width=\textwidth]{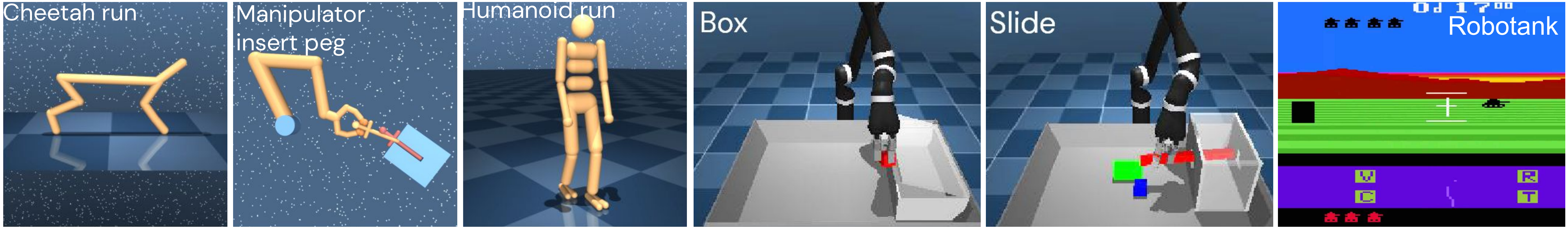}
    \caption{Environment used in the experiments. 9 environments from the \dmcontrol{} domain (3 examples on the left), 4 from the \mpg{} domain (2 examples in the middle), and 3 from Atari domain (1 example on the right).}
    \label{fig:environments}
\end{figure}

In our experiments we consider three domains (\autoref{fig:environments}) with different properties. 
Two domains are standard benchmarks in RL for continuous and discrete control and another one is motivated by the challenges of evaluating control policies in robotics.
For each domain we describe the environments, datasets for offline RL and OPE, and sets of policies below. More details are provided in \autoref{sec:domain_details}.

\paragraph{DeepMind Control Suite (\dmcontrol{})} 
This is a standard set of continuous control environments~\citep{tassa2020dmcontrol}.
It contains \num{9} tasks of locomotion and manipulation, some of which are shown in \autoref{fig:environments}, left.
The observations are the state of the MPD, including joint angles and velocities. 
The dimensionality and semantics of the continuous actions vary across tasks. 
We use the offline RL policies from DOPE \citep{fu2021benchmarks} (a benchmark for OPE and OPS) that are trained using the dataset from the RL Unplugged benchmark~\citep{gulcehre2020rl}. 
For each task, there are up to \num{96} policies with various algorithms (D4PG~\citep{barth2018distributed}, ABM~\citep{siegel2020keep}, and CRR~\citep{wang2020critic}), hyperparameters, and number of training steps.

\paragraph{Manipulation Playground (\mpg{})} 
This is a simulated robotics environment. 
The task is to control a Kinova Jaco arm with $9$ degrees of freedom. 
Joint velocity control of $6$ arm and $3$ hand joints is used. 
The policies are learnt from proprioception and image input from frontal and in-hand cameras of $64 \times 64$ dimensions.
We perform \num{4} manipulation tasks within a $20 \times 20$ cm basket (\autoref{fig:environments}, middle).
The dataset is generated by running an online RL algorithm as in~\citep{wang2020critic} and the policies are generated as in~\citep{paine2020hyperparameter}. For each task, there are up to \num{256} policies trained with various algorithms (including BC, D4PG~\citep{barth2018distributed}, and CRR~\citep{wang2020critic}), and hyperparameters.

\paragraph{Atari}
This is a popular benchmark with discrete actions in online and offline RL~\citep{gulcehre2020rl}.
For each of \num{3} games we trained \num{210} policies on an offline dataset~\citep{gulcehre2020rl} from high dimensional pixel observations with six offline RL methods, including Double DQN~\citep{van2016deep}, CQL~\citep{kumar2020conservative}, BCQ~\citep{fujimoto2019off}, REM~\citep{agarwal2019optimistic} and BVE~\citep{gulcehre2021regularized}.
As the range of returns is quite different from the other two domains, we scale the discounted returns to have approximately the same mean and variance based on the offline RL dataset.

We obtain OPE estimates for each policy by running Fitted Q-Evaluation (FQE)~\citep{migliavacca2010fitted, le2019batch} on the offline RL data. FQE is the default OPE method used in our experiments unless stated otherwise.

\subsection{Experimental setting and implementation}
\label{sec:exp-setting}
To evaluate the policy selection procedure in each experiment we randomly select a subset of $K$ policies out of all trained policies. 
We set $K=50$ for dm-control and $K=200$ for MPG and Atari. 
Then, we repeat each experiment \num{100} times, and report the average results and the standard deviation of the mean estimate.
As a performance metric we monitor the simple regret (\autoref{eq:simple-regret}) as a function of the number of executed episodes.
The algorithms are progressively evaluated when each trajectory is added.
To make the results comparable across domains, we first compute the gap between the best and the worst performing policy in each environment across all the policies. 
Then, we normalise the scores in each experiment by dividing them by this gap. 
The results by environment are discussed in the appendix.

\paragraph{Independent policy model}
To study the contribution of modeling the policy correlation with GP, we consider a baseline model with independently distributed value $\mu_k$ across policies.
The data generating process for the $k$-th policy follows a hierarchical Gaussian distribution:
\begin{equation}
    r_k^i \overset{iid}{\sim} \gauss(\mu_k, \sigma_{k,r}^2), \forall i, \text{ with }
    \mu_k \sim \gauss(\rho_k, \sigma_{k,\rho}^2)\,,
    \label{eq:ind_arm_model}
\end{equation}
where we assume a weakly informative inverse Gamma prior for variance $\sigma_{k,r}^2$ and $\sigma_{k,\rho}^2$ as in GP. We refer to this model as independent policy model (\textsc{Ind}).
Such model can be used in combination with any policy sampling strategy. 

All online and offline RL algorithms, as well as offline policy evaluation algorithms in this work are implemented with Acme~\citep{hoffman2020acme} and Reverb~\citep{cassirer2021reverb}, and run on GPUs on an internal cluster.
We implement \textsc{GP} and \textsc{Ind} together with all BO algorithms using the same TensorFlow~\citep{tensorflow2015-whitepaper} codebase, and run all the policy selection experiments using CPUs.
Details about the tasks and policies, hyperparameter values, and additional BO algorithms and ablations are presented in the appendix.
\subsection{Comparison to the existing methods}
\label{sec:exp-results}

In our main experiments we would like to verify that the proposed method performs better than two strategies commonly employed in practice (see \autoref{sec:intro}).
To this end, we compare:

\setenumerate{noitemsep,topsep=0pt,parsep=0pt,partopsep=0pt}
\begin{enumerate}
    \item \textbf{\ours{}}: Active-Offline Policy Selection, our proposed method that uses OPE estimates to bootstrap GP learning and performs UCB policy sampling as discussed in \autoref{sec:method}.
    \item \textbf{\OPE{}}: standard offline policy selection technique that selects the policy with the highest OPE estimate \citep{paine2020hyperparameter}.
    \item \textbf{\IndUniform{}}: standard online selection where we execute policies uniformly and select the one with the highest return belief (estimated with independent policy model \textsc{Ind}).
\end{enumerate}

\begin{figure}[t!]
    \centering
    \includegraphics{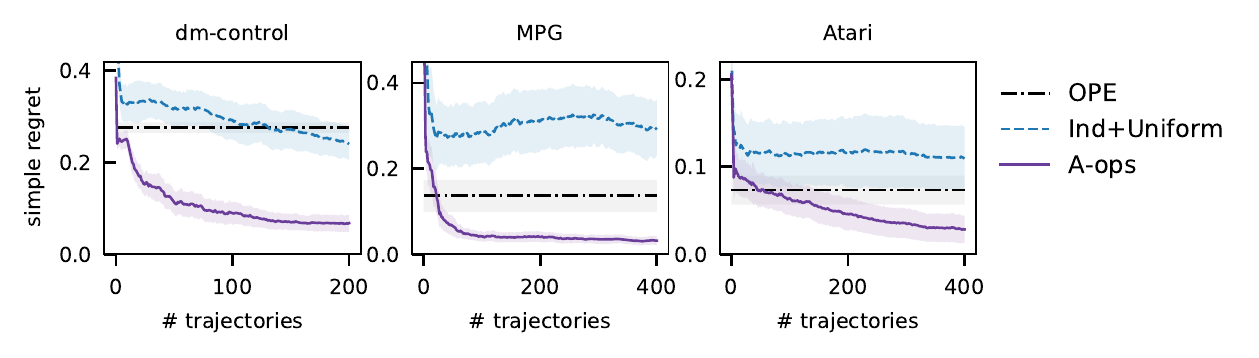}
    \caption{Comparison of \ours{} approach, offline policy selection with \OPE{}, and online policy selection with \IndUniform{} on \dmcontrol{}, \mpg{} and Atari environments. The results are averaged across \num{9}, \num{4} and \num{3} tasks and \num{100} experiments in each of them. Shaded areas show standard deviation of the mean. Our method \ours{} quickly surpasses the baselines and achieves the lowest regrets. }
    \label{fig:results}
\end{figure}

\autoref{fig:results} presents the simple regret as a function of the number of executed trajectories. 
The results are averaged across \num{9} tasks from dm-control, $4$ tasks from MPG and \num{3} games from Atari. 
Our proposed method \ours{} combines offline and online policy evaluations. 
It quickly outperforms \OPE{} and improves with more trajectories.
\ours{} estimate of the best policy may be erroneous initially due to noisy reward samples (\emph{e.g.}, in \mpg{}), but it quickly updates, thus allowing for significant improvements over OPE only after a few actively selected trajectories.

When comparing offline and online policy selection we notice that for the range of considered budgets, online policy selection \IndUniform{} performs worse than \OPE{}.
It suggests that single trajectory returns are noisier than OPE estimates in these environments (more discussion on this is in the appendix).
Naturally, we expect that given a larger budget the online baseline will eventually surpass its offline counterpart in all domains.
\subsection{Component ablations}
\label{sec:exp-ablations}

We are not aware of other works on active offline policy selection. 
Thus, in this section we investigate alternative solutions that help us to understand the importance of the proposed components of \ours{}. 
We design our ablation studies to test the contribution of the three components of \ours{}:
\begin{enumerate}[label={\arabic*)}]
    \item \emph{OPE}: initialisation of algorithms with or without OPE estimates. 
    \item \emph{active learning}: active policy selection (\textsc{UCB}) or uniform sampling (\textsc{Uniform}).
    \item \emph{policy kernel}: \textsc{GP} model or independent model baseline (\textsc{Ind}).
\end{enumerate}
\autoref{tab:algorithms} summarises the \num{8} possible combinations of the components.
We refer to our full method (\textsc{GP+UCB+OPE}) simply as \ours{}.
We also include the pure offline OPE estimates as before.

\begin{table}[t!]
    \caption{Design of the ablation experiments across three axes: the use of OPE, GP or independent policy model, active policy selection or uniform sampling. We refer to our method that combines all the components as \ours{}.}
    \vspace{2mm}
    \centering
    {\small
    \begin{tabular}{l|cc|cc}
    & \multicolumn{2}{c|}{\bf No OPE} & \multicolumn{2}{c}{\bf With OPE} \\
    \midrule
    & \multicolumn{1}{c|}{\bf Independent} & {\bf GP} & \multicolumn{1}{c|}{\bf Independent} & {\bf GP} \\
    \midrule
    {\bf Uniform} & \IndUniform{} & \GPUniform{} &  \IndUniformOPE{} & \GPUniformOPE{} \\
    {\bf UCB} & \IndUCB{} & \GPUCB{} & \IndUCBOPE{} & \ours{} \\
    \end{tabular}
    }
    \label{tab:algorithms}
\end{table}

\begin{figure}[t!]
    \centering
    \includegraphics[width=1\linewidth]{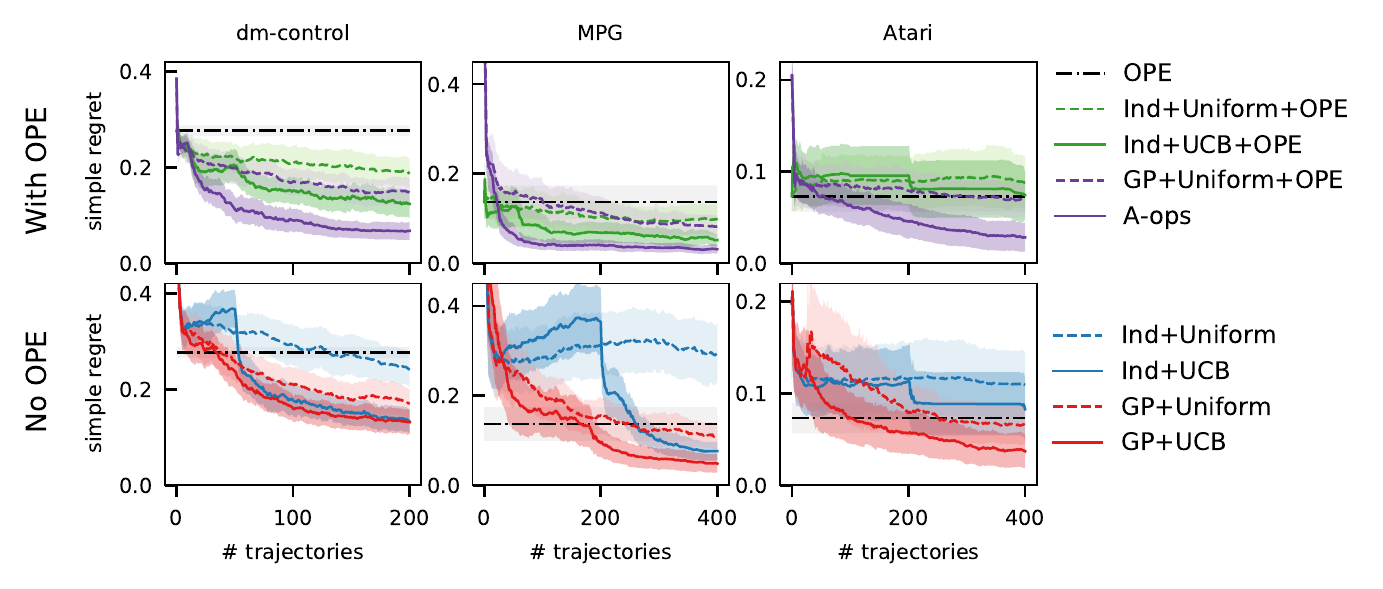}
    \caption{Ablations. In three domains, we ablate each of three components of our approach. 1) The use of OPE (with OPE in the first row and without in the second). 2) The choice of policy model: GP (purple and red) and independent (green and blue). 3) The policy selection strategy: active (dark, solid line) and uniform sampling (lighter, dashed line). 
    In general, active is better than uniform, methods that exploit policy correlation are better than methods that do not, and methods that use an OPE prior are better than those that do not. 
    }
    \label{fig:results_ablations}
\end{figure}

The results with all the baselines in two domains
are presented in \autoref{fig:results_ablations}. 
First of all we observe that \ours{} outperforms all other methods across a wide range of budgets, thus, all the components are crucial for the performance of our method and excluding any of them leads to significantly worse results.
Next, we answer three questions about each component individually.

\paragraph{1. How important is it to incorporate OPE estimates?}
Across all the experiments, incorporating OPE estimates (top row of \autoref{fig:results_ablations}) always considerably improves the results compared to starting policy selection from scratch (bottom row). 
For example, in MPG when \ours{} starts with OPE, it is able to cut the OPE regret by more than half before executing \num{100} trajectories. 
For this budget, \GPUCB{}, which starts from scratch, does not reach the same regret as OPE.

\paragraph{2. How informative is our kernel?}
To understand the value of modelling the correlation between policies through the kernel, we compare the GP method with the independent policy model (\textsc{Ind}).
In \autoref{fig:results_ablations} purple and red lines use {\textsc GP} as the policy model and green and blue use the \textsc{Ind} model.
In the vast majority of settings (\num{11} out of \num{12}) the use of a kernel significantly improves the results and in one setting (\IndUniformOPE{} vs \GPUniformOPE{} in MPG domain) they perform on par. 
We conclude that the kernel is a key ingredient for improving data efficiency.

\paragraph{3. How important is the selection strategy?}
The active selection of policies is generally beneficial for identifying a good policy. 
For this we refer to the results of dashed lines that correspond to \textsc{Uniform} policy selection and solid lines that correspond to the use of {\textsc UCB}.
Using \IndUCB{} yields a high regret in the initial exploration stage, but it improves substantially over \IndUniform{} later. 
Moreover, incorporating OPE estimates and a kernel (resulting in \ours{}) significantly shortens the exploration stage.

\subsection{Results on the real robot}

Evaluation of the offline RL policies is a bottleneck in real world robotic applications because robotic resources are limited and robots cannot be accelerated beyond real time. 
Thus, we believe it is important to carefully design the evaluation procedure.
Here we consider a robotic stacking task, where we train a single policy to operate a robotic arm to stack different object groups, each group containing objects of diverse shapes. It receives a reward of 1 if the stacking goal is achieved in that step. See \cite{lee2021stacking} for detailed task description.
In our experiments we propose to use \ours{} method components in two ways: for selecting the best policy among a set and for selecting the best snapshot of a policy.

\paragraph{Selecting the best policy}
We use \num{36} real robot policies from \cite{lee2021stacking} (skill mastery, Tab.~$2$, $3$). 
Here we treat all \num{5} object groups together as one evaluate setting and one policy evaluation is equivalent to running the policy to generate one stacking trajectory for each group.
The policies are trained with either a sim2real and offline RL approach, both of which do not involve direct interactions with the physical robot. 
We measure the cumulative discounted return to be consistent with the rest of the experiments, while in \cite{lee2021stacking} binary success is used. 
For the kernel, we normalise each action dimension by its variance, but the rest of the settings and hyperparameters is consistent with simulated experiments.
In each experiment we select \num{20} out of \num{36} policies.
The results are reported in \autoref{fig:robotics}.

In \autoref{fig:robotics}, left we compare \ours{} to the strategies that are commonly used in practice: pure online evaluation (\IndUniform{}) and pure offline evaluation with OPE. 
There is a large variance in returns obtained by a single policy and as a result \IndUniform{} reduces the regret slowly. 
The OPE method of our choice (FQE) ranks policies very well (see Appendix \ref{robot-FQE}) and its regret is very low. 
However, after less than \num{200} policy evaluations, \ours{} is able to achieve even lower simple regret.
The full ablations are presented in \autoref{fig:robotics}, middle (strategies which use OPE) and right (strategies without OPE). 
The full \ours{} approach outperforms the other alternatives but the variance of the results is high. 

\paragraph{Selecting the best snapshot}
The Gaussian process that is at the core of \ours{} could be also used for selecting the best checkpoint which is equivalent to deciding on the number of training steps for the policy.
In our experiment, we consider \num{69} checkpoints of the best performing policy from \cite{lee2021stacking}.
Then, we use the results of full policy evaluation of \num{36} policies and make predictions for all the checkpoints. 
We select \num{5} most promising checkpoints and send them for evaluation on the real robot.
The evaluation results are presented in \autoref{tab:best_checkpoint}. 
We show the return for a given number of training steps for the evaluated checkpoints and the existing best policy (shown in blue).
All selected checkpoints slightly improve the results compared to the previous best agent and the improvements are between \num{1.55}\% and \num{4.95}\%, we highlight the best checkpoint in bold.
We conclude that the proposed methodology might be useful for the task of deciding on the number of training steps.

\begin{minipage}{\textwidth}
\vspace{-5mm}
  \begin{minipage}[b]{0.7\textwidth}
    \centering
    \includegraphics[width=1\linewidth]{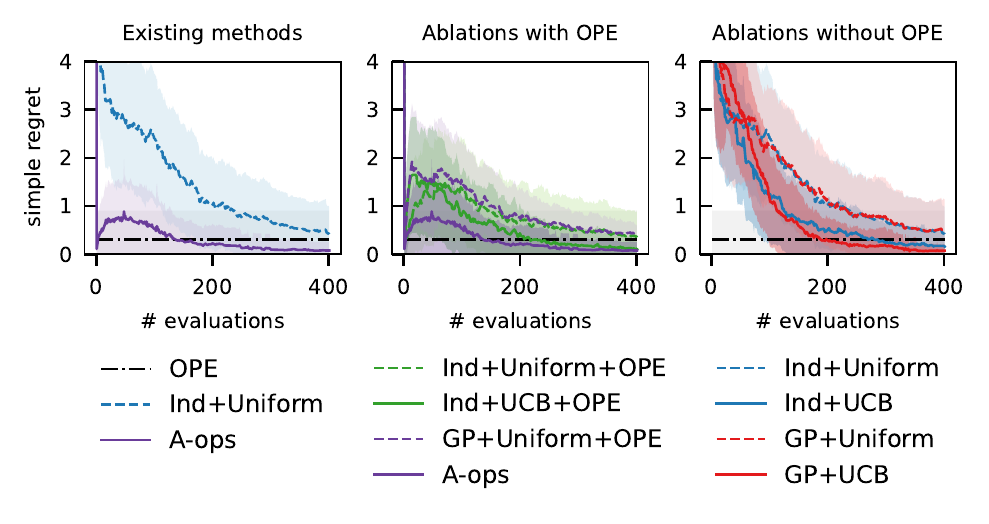}
    \vspace{-8mm}
    \captionof{figure}{\ours{} results on the real robot. 
    The comparison of \ours{} with the commonly used strategies is shown on the left, ablations when using OPE is in the middle and without OPE is on the right. 
    \ours{} is able to reach the best simple regret with a moderate evaluation budget.}
    \label{fig:robotics}
  \end{minipage}
  \hfill
  \begin{minipage}[b]{0.25\textwidth}
    \centering
    {\small
    \begin{tabular}{r|r}
    {\bf training} & {\bf discounted} \\
    {\bf step} & {\bf return} \\
    \midrule
    \num{841000} & $25.58$ \\
    \num{983000} & {\bf 25.81} \\
    \num{1143000} &	$25.33$ \\
    \num{1283000} &	$25.07$ \\
    \num{1303000} &	$24.98$ \\
    \midrule
    \color{Blue} \num{1003000} & \color{Blue} $24.59$ \\
    \end{tabular}
    }
      \captionof{table}{The best current policy is shown in {\color{Blue} blue}, the newly discovered best policy is in {\bf bold}. \ours{} components is promising for identifying the best checkpoint.}
    \label{tab:best_checkpoint}
    \end{minipage}
  \end{minipage}

\subsection{More ablation studies}

\begin{figure}
     \vspace{-.5cm}
     \includegraphics[width=0.5\linewidth]{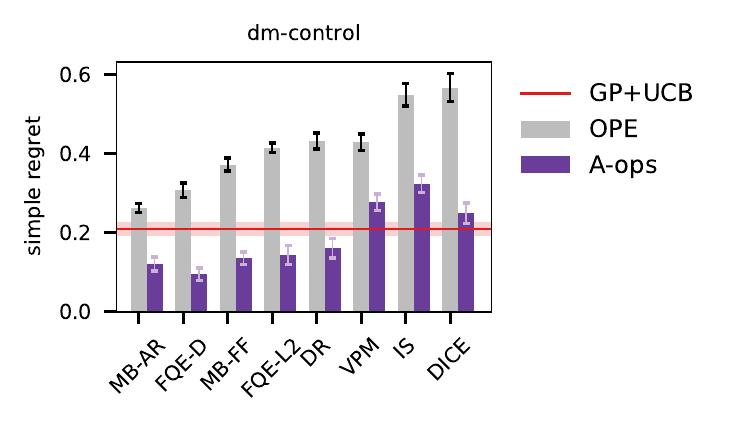}
     \hspace{-0.5cm}
     \includegraphics[width=0.52\linewidth]{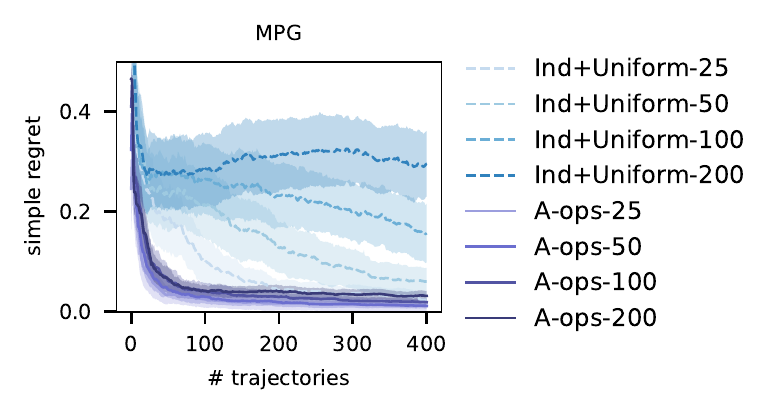}
        \caption{Left: The performance of \ours{} on \dmcontrol{} with varying type of OPE estimate with a budget of 50 trajectories. The various OPE methods are Model-Based OPE with an auto-regressive transition model (MB-AR), Fitted Q Evaluation with distributional Q function and distributional loss (FQE-D), Model-Based OPE with a feedforward transition model (MB-FF), Fitted Q Evaluation with standard Q function and L2 loss (FQE-L2), Double Robust (DR), Variational Power Method (VPM), Importance Sampling (IS), Distribution Correction Estimation (DICE). 
        \ours{} can improve upon the initial OPE estimates in all cases.
        Right: Performance of \ours{} and \IndUniform{} with varying number of policies available for execution. \IndUniform{} is affected a lot by the increase in the number of policies. Our approach \ours{} performs similarly when the number of policies is increase by a factor of almost \num{10}.}
        \label{fig:ope-num_policies-ablation}
\vspace{-5mm}
\end{figure}

Incorporating the OPE estimates is clearly beneficial for policy selection. Here, we further study: 
\textbf{How does the performance depend on the quality of OPE estimates?}
Fig.~\ref{fig:ope-num_policies-ablation}, left compares \ours{} using estimates from different OPE algorithms in DOPE \citep{fu2021benchmarks} on dm-control with a small budget of \num{50} trajectories. \ours{} always improves upon the corresponding OPE estimate, and the performance clearly depends on the quality of the associated OPE estimate.
The performance of \GPUCB{} with \num{50} trajectories that does not rely on any OPE metric is shown as a horizontal line.
Some OPE estimates (e.g., IS, VPM and DICE) have large estimation errors and even a negative correlation with the ranks of policies~\citep{fu2021benchmarks}. 
As expected, \ours{} is negatively impacted by poor OPE estimates and ignoring them could yield better results. 
It should be possible to correct for this to a greater extent with more online observations. For now we follow the recommendation to use the FQE estimates.

\label{sec:ablation_num_of_policy}
In \autoref{sec:gp} we hypothesised that \ours{} method should be data efficient as it would scale with the diversity of policies and not their number. 
We next try to verify it and answer the question: \textbf{How does the policy selection algorithm scale with the growing number of candidate policies?}
In \autoref{fig:ope-num_policies-ablation}, right we show the simple regret for \ours{} and \IndUniform{} in the \mpg{} domain when we sample \num{25}, \num{50}, \num{100} and \num{200} policies. 
We see that when the number of policies is not very high (\num{25}), the algorithm that treats policies independently and samples them at uniform (\IndUniform{}-25) can closely approach the performance of our method \ours{}-25 after \num{100} trajectories. 
However, when the number of policies grows, this small interaction budget is not enough for the naive algorithm: \IndUniform{}-200 significantly degrades its performance. 
At the same time, \ours{}-200 is minimally affected by the increase in the number of policies as the diversity of the policies stays approximately the same. 

\subsection{Qualitative results}

\begin{figure}
    \centering
    \includegraphics[width=1\linewidth]{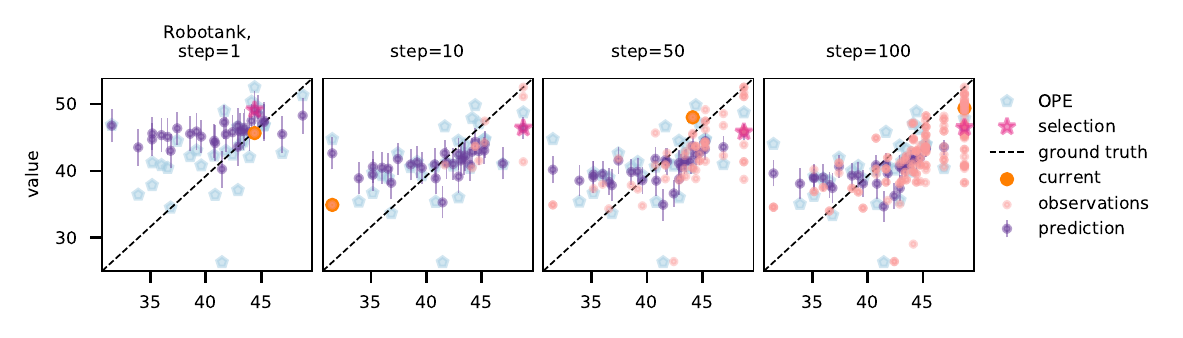}
    \caption{Visualisation of the qualitative performance of \ours{} algorithm on \texttt{Robotank} environment with \num{20} randomly selected policies (ordered by ground truth only for the illustration purpose). From left to right we show the state at \num{1}, \num{10}, \num{50}, \num{100} algorithm iterations. The initial OPE scores are shown in light blue, the policy selected by \ours{} is highlighted with magenta star, the observation of the currently executed policy is shown in orange circle, past observations are in pink and the prediction with its standard deviation is in purple.}
    \label{fig:qualitative}
\end{figure}

Finally, to understand the behaviour of \ours{} method we show its progress for one of the experiments with Robotank game in 
\autoref{fig:qualitative}.
For visualisation purposes, we use only \num{20} policies which are ordered by their ground truth returns (not observed by the method). Perfect value predictions would be aligned along the diagonal line. 
We show the initial \emph{OPE} values, current \emph{selection} of the best policy, \emph{current} observation of the executed policy, past \emph{observations} and the algorithm \emph{prediction} with its standard deviation for steps \num{1}, \num{10}, \num{50}, and \num{100}.
The \ours{} prediction is determined by the OPE observations and past observations of the policy returns. The kernel reinforces smoothness of the function.
We can see that the initial selection is dominated by OPE prediction and the selected policy is quite far from the best policy. 
With more online observations, the predictions improve. 
Our algorithm is data-efficient because 1) it samples more promising policies more frequently (more observations with higher scores), 2) the knowledge from one policy is propagated to the policies related by the kernel (predictions for some policies are updated without any new observations of their values).

\section{Conclusions}
\label{sec:conclusion}

In this paper, we introduced the active offline policy selection problem. 
We proposed a BO solution that integrates OPE estimates with evaluations obtained by interacting with the environment. 
We also designed a novel kernel that models correlation over policies. 
Experiments on three control domains proved that our solution is very effective. 
Furthermore, our experiments ablated the different components of the solution, highlighting the value of modelling correlation via the kernel, active selection, and incorporation of OPE estimates in the observation model.
Finally, we have also demonstrated that our approach is efficient on the real robot.

\paragraph{Limitations and future work}
The \ours{} approach could be of use in many safety critical applications. This work does not deal with this question explicitly as we assume safety constraints to be implemented directly on the hardware. It means that if an unsafe policy is attempted, the safety controller would terminate the policy before reaching the end of an episode, resulting in low reward and thus discouraging \ours{} from trying this policy again.
Alternatively, we could alleviate this limitation by applying constrained BO techniques~\citep{sui2015safe} to search for the best policy without violating safety constraints.

In this work, we explore the policy structure based on the offline dataset, and use every new trajectory only to obtain a noisy observation of policy values. 
This allows a clean Bayesian optimization formulation. 
Going forward, more information could be extracted from the new trajectories, for example, to further improve OPE estimates, GP kernel, or policies themselves. 
However, using this information would make the solution much more complex and we may need to take the additional computational cost into account. 
Finally, to construct a better policy kernel, we may investigate how to measure the policy similarity on the most informative subset of states in the offline dataset.

\clearpage
\section*{Acknowledgements}

We would like to thank Bobak Shahriari for the useful discussions and feedback on the paper. 
Yuxiang Zhou, Konstantinos Bousmalis, and Jost Tobias Springenberg consulted us on the real robot setup and helped us to evaluate the policies on the robot for which we are very grateful.
{\small
\bibliographystyle{plain}
\bibliography{string,refs}

\begin{thebibliography}{10}

\bibitem{tensorflow2015-whitepaper}
Mart\'{\i}n Abadi, Ashish Agarwal, Paul Barham, Eugene Brevdo, Zhifeng Chen,
  Craig Citro, Greg~S. Corrado, Andy Davis, Jeffrey Dean, Matthieu Devin,
  et~al.
\newblock {TensorFlow}: Large-scale machine learning on heterogeneous systems,
  2015.
\newblock Software available from tensorflow.org.

\bibitem{agarwal2019optimistic}
Rishabh Agarwal, Dale Schuurmans, and Mohammad Norouzi.
\newblock An optimistic perspective on offline reinforcement learning.
\newblock In {\em International Conference on Machine Learning}, 2020.

\bibitem{Akiyama2010}
Takayuki Akiyama, Hirotaka Hachiya, and Masashi Sugiyama.
\newblock Efficient exploration through active learning for value function
  approximation in reinforcement learning.
\newblock {\em Neural Networks}, 2010.

\bibitem{audibert2010best}
Jean-Yves Audibert, S{\'e}bastien Bubeck, and R{\'e}mi Munos.
\newblock Best arm identification in multi-armed bandits.
\newblock In {\em Conference on Computational Learning Theory}, 2010.

\bibitem{azimi2012hybrid}
Javad Azimi, Ali Jalali, and Xiaoli~Zhang Fern.
\newblock Hybrid batch bayesian optimization.
\newblock In {\em International Conference on Machine Learning}, 2012.

\bibitem{ball2020readypolicyone}
Philip Ball, Jack Parker-Holder, Aldo Pacchiano, Krzysztof Choromanski, and
  Stephen Roberts.
\newblock Ready policy one: World building through active learning.
\newblock In {\em International Conference on Machine Learning}, 2020.

\bibitem{barth2018distributed}
Gabriel Barth-Maron, Matthew~W Hoffman, David Budden, Will Dabney, Dan Horgan,
  TB~Dhruva, Alistair Muldal, Nicolas Heess, and Timothy Lillicrap.
\newblock Distributed distributional deterministic policy gradients.
\newblock In {\em International Conference for Learning Representations}, 2018.

\bibitem{Brochu:2010}
E.~Brochu, T.~Brochu, and N.~de~Freitas.
\newblock A {Bayesian} interactive optimization approach to procedural
  animation design.
\newblock In {\em ACM SIGGRAPH/Eurographics Symposium on Computer Animation},
  2010.

\bibitem{Brochu:2007}
E.~Brochu, N.~de~Freitas, and A.~Ghosh.
\newblock Active preference learning with discrete choice data.
\newblock In {\em Advances in Neural Information Processing Systems}, 2007.

\bibitem{cabi2020sketchy}
Serkan Cabi, Sergio G{\'o}mez~Colmenarejo, Alexander Novikov, Ksenia
  Konyushkova, Scott Reed, Rae Jeong, Konrad Zolna, Yusuf Aytar, David Budden,
  Mel Vecerik, et~al.
\newblock Scaling data-driven robotics with reward sketching and batch
  reinforcement learning.
\newblock In {\em Robotics: Science and Systems Conference}, 2020.

\bibitem{cassirer2021reverb}
Albin Cassirer, Gabriel Barth-Maron, Eugene Brevdo, Sabela Ramos, Toby Boyd,
  Thibault Sottiaux, and Manuel Kroiss.
\newblock Reverb: A framework for experience replay.
\newblock {\em arXiv preprint arXiv:2102.04736}, 2021.

\bibitem{chen2018bayesian}
Yutian Chen, Aja Huang, Ziyu Wang, Ioannis Antonoglou, Julian Schrittwieser,
  David Silver, and Nando de~Freitas.
\newblock Bayesian optimization in {AlphaGo}.
\newblock {\em arXiv preprint arXiv:1812.06855}, 2018.

\bibitem{dabney2018implicit}
Will Dabney, Georg Ostrovski, David Silver, and R\'{e}mi Munos.
\newblock Implicit quantile networks for distributional reinforcement learning.
\newblock In {\em International Conference on Machine Learning}, 2018.

\bibitem{daniel2014active}
Christian Daniel, Malte Viering, Jan Metz, Oliver Kroemer, and Jan Peters.
\newblock Active reward learning.
\newblock In {\em Robotics: Science and Systems Conference}, 2014.

\bibitem{dudik2011doubly}
Miroslav Dud{\'{\i}}k, John Langford, and Lihong Li.
\newblock Doubly robust policy evaluation and learning.
\newblock {\em arXiv preprint arXiv:1103.4601}, 2011.

\bibitem{challenges}
Gabriel Dulac{-}Arnold, Daniel~J. Mankowitz, and Todd Hester.
\newblock Challenges of real-world reinforcement learning.
\newblock {\em arXiv preprint arXiv:1904.12901}, 2019.

\bibitem{epshteyn2008active}
Arkady Epshteyn, Adam Vogel, and Gerald DeJong.
\newblock Active reinforcement learning.
\newblock In {\em International Conference on Machine Learning}, 2008.

\bibitem{mrdr}
Mehrdad Farajtabar, Yinlam Chow, and Mohammad Ghavamzadeh.
\newblock More robust doubly robust off-policy evaluation.
\newblock In {\em International Conference on Machine Learning}, 2018.

\bibitem{fu2020d4rl}
Justin Fu, Aviral Kumar, Ofir Nachum, George Tucker, and Sergey Levine.
\newblock {D4RL}: Datasets for deep data-driven reinforcement learning.
\newblock {\em arXiv preprint arXiv:2004.07219}, 2020.

\bibitem{fu2021benchmarks}
Justin Fu, Mohammad Norouzi, Ofir Nachum, George Tucker, ziyu wang, Alexander
  Novikov, Mengjiao Yang, Michael~R Zhang, Yutian Chen, Aviral Kumar, Cosmin
  Paduraru, Sergey Levine, and Thomas Paine.
\newblock Benchmarks for deep off-policy evaluation.
\newblock In {\em International Conference for Learning Representations}, 2021.

\bibitem{fujimoto2019off}
Scott Fujimoto, David Meger, and Doina Precup.
\newblock Off-policy deep reinforcement learning without exploration.
\newblock In {\em International Conference on Machine Learning}, 2019.

\bibitem{gulcehre2021regularized}
Caglar Gulcehre, Sergio~G{\'o}mez Colmenarejo, Ziyu Wang, Jakub Sygnowski,
  Thomas Paine, Konrad Zolna, Yutian Chen, Matthew Hoffman, Razvan Pascanu, and
  Nando de~Freitas.
\newblock Regularized behavior value estimation.
\newblock {\em arXiv preprint arXiv:2103.09575}, 2021.

\bibitem{gulcehre2020rl}
Caglar Gulcehre, Ziyu Wang, Alexander Novikov, Tom~Le Paine, Sergio~Gomez
  Colmenarejo, Konrad Zolna, Rishabh Agarwal, Josh Merel, Daniel Mankowitz,
  Cosmin Paduraru, et~al.
\newblock {RL} {U}nplugged: {A} suite of benchmarks for offline reinforcement
  learning.
\newblock {\em arXiv preprint arXiv:2006.13888}, 2020.

\bibitem{harb2020policy}
Jean Harb, Tom Schaul, Doina Precup, and Pierre-Luc Bacon.
\newblock Policy evaluation networks.
\newblock {\em arXiv preprint arXiv:2002.11833}, 2020.

\bibitem{hoffman2020acme}
Matt Hoffman, Bobak Shahriari, John Aslanides, Gabriel Barth-Maron, Feryal
  Behbahani, Tamara Norman, Abbas Abdolmaleki, Albin Cassirer, Fan Yang, Kate
  Baumli, et~al.
\newblock Acme: A research framework for distributed reinforcement learning.
\newblock {\em arXiv preprint arXiv:2006.00979}, 2020.

\bibitem{hoffman2014correlation}
Matthew Hoffman, Bobak Shahriari, and Nando de~Freitas.
\newblock On correlation and budget constraints in model-based bandit
  optimization with application to automatic machine learning.
\newblock In {\em Artificial Intelligence and Statistics}, 2014.

\bibitem{Hutter:smac}
F.~Hutter, H.~H. Hoos, and K.~Leyton-Brown.
\newblock Sequential model-based optimization for general algorithm
  configuration.
\newblock In {\em Learning and Intelligent Optimization}, 2011.

\bibitem{jiang2016doubly}
Nan Jiang and Lihong Li.
\newblock Doubly robust off-policy value evaluation for reinforcement learning.
\newblock In {\em International Conference on Machine Learning}, 2016.

\bibitem{kalashnikov2018qt}
Dmitry Kalashnikov, Alex Irpan, Peter Pastor, Julian Ibarz, Alexander Herzog,
  Eric Jang, Deirdre Quillen, Ethan Holly, Mrinal Kalakrishnan, Vincent
  Vanhoucke, et~al.
\newblock {QT-Opt}: Scalable deep reinforcement learning for vision-based
  robotic manipulation.
\newblock In {\em Conference on Robot Learning}, 2018.

\bibitem{kaufmann2012bayesian}
Emilie Kaufmann, Olivier Capp{\'e}, and Aur{\'e}lien Garivier.
\newblock On {B}ayesian upper confidence bounds for bandit problems.
\newblock In {\em Artificial intelligence and statistics}, 2012.

\bibitem{kidambi2020morel}
Rahul Kidambi, Aravind Rajeswaran, Praneeth Netrapalli, and Thorsten Joachims.
\newblock Morel: Model-based offline reinforcement learning.
\newblock {\em arXiv preprint arXiv:2005.05951}, 2020.

\bibitem{konyushkova19discovering}
Ksenia Konyushkova, Raphael Sznitman, and Pascal Fua.
\newblock Discovering general-purpose active learning strategies.
\newblock {\em arXiv preprint arXiv:1810.04114}, 2019.

\bibitem{krause2012contextual}
Andreas Krause and Cheng~S Ong.
\newblock Contextual gaussian process bandit optimization.
\newblock {\em Advances in Neural Information Processing Systems}, 2012.

\bibitem{krueger2020active}
David Krueger, Jan Leike, Owain Evans, and John Salvatier.
\newblock Active reinforcement learning: Observing rewards at a cost.
\newblock {\em arXiv preprint arXiv:2011.06709}, 2020.

\bibitem{kumar2020conservative}
Aviral Kumar, Aurick Zhou, George Tucker, and Sergey Levine.
\newblock Conservative {Q}-learning for offline reinforcement learning.
\newblock {\em arXiv preprint arXiv:2006.04779}, 2020.

\bibitem{le2019batch}
Hoang Le, Cameron Voloshin, and Yisong Yue.
\newblock Batch policy learning under constraints.
\newblock In {\em International Conference on Machine Learning}, 2019.

\bibitem{lee2021stacking}
Alex~X. Lee, Coline~Manon Devin, Yuxiang Zhou, Thomas Lampe, Konstantinos
  Bousmalis, Jost~Tobias Springenberg, Arunkumar Byravan, Abbas Abdolmaleki,
  Nimrod Gileadi, David Khosid, Claudio Fantacci, Jose~Enrique Chen, Akhil
  Raju, Rae Jeong, Michael Neunert, Antoine Laurens, Stefano Saliceti, Federico
  Casarini, Martin Riedmiller, Raia Hadsell, and Francesco Nori.
\newblock Beyond pick-and-place: Tackling robotic stacking of diverse shapes.
\newblock In {\em Conference on Robot Learning}, 2021.

\bibitem{levine2016end}
Sergey Levine, Chelsea Finn, Trevor Darrell, and Pieter Abbeel.
\newblock End-to-end training of deep visuomotor policies.
\newblock {\em Journal of Machine Learning Research}, 2016.

\bibitem{levine2020offline}
Sergey Levine, Aviral Kumar, George Tucker, and Justin Fu.
\newblock Offline reinforcement learning: Tutorial, review, and perspectives on
  open problems.
\newblock {\em arXiv preprint arXiv:2005.01643}, 2020.

\bibitem{li2010contextual}
Lihong Li, Wei Chu, John Langford, and Robert~E Schapire.
\newblock A contextual-bandit approach to personalized news article
  recommendation.
\newblock In {\em World Wide Web}, 2010.

\bibitem{li2011unbiased}
Lihong Li, Wei Chu, John Langford, and Xuanhui Wang.
\newblock Unbiased offline evaluation of contextual-bandit-based news article
  recommendation algorithms.
\newblock In {\em ACM international conference on Web search and data mining},
  2011.

\bibitem{liu2018breaking}
Qiang Liu, Lihong Li, Ziyang Tang, and Dengyong Zhou.
\newblock Breaking the curse of horizon: Infinite-horizon off-policy
  estimation.
\newblock In {\em Advances in Neural Information Processing Systems}, 2018.

\bibitem{lizotte2007automatic}
Daniel~J Lizotte, Tao Wang, Michael~H Bowling, and Dale Schuurmans.
\newblock Automatic gait optimization with gaussian process regression.
\newblock In {\em International Joint Conference on Artificial Intelligence},
  2007.

\bibitem{Martinez-Cantin:2009}
R.~{Martinez--Cantin}, N.~{de Freitas}, E.~Brochu, J.~Castellanos, and
  A.~Doucet.
\newblock A {Bayesian} exploration-exploitation approach for optimal online
  sensing and planning with a visually guided mobile robot.
\newblock {\em Autonomous Robots}, 2009.

\bibitem{migliavacca2010fitted}
Martino Migliavacca, Alessio Pecorino, Matteo Pirotta, Marcello Restelli, and
  Andrea Bonarini.
\newblock Fitted policy search: Direct policy search using a batch
  reinforcement learning approach.
\newblock In {\em Workshop on Evolutionary and Reinforcement Learning for
  Autonomous Robot Systems}, 2010.

\bibitem{chips}
Azalia Mirhoseini, Anna Goldie, Mustafa Yazgan, Joe Jiang, Ebrahim Songhori,
  Shen Wang, Young-Joon Lee, Eric Johnson, Omkar Pathak, Sungmin Bae, et~al.
\newblock Chip placement with deep reinforcement learning.
\newblock {\em arXiv preprint arXiv:2004.10746}, 2020.

\bibitem{dqn}
Volodymyr Mnih, Koray Kavukcuoglu, David Silver, Alex Graves, Ioannis
  Antonoglou, Daan Wierstra, and Martin Riedmiller.
\newblock Playing {A}tari with deep reinforcement learning.
\newblock {\em arXiv preprint arXiv:1312.5602}, 2013.

\bibitem{munos2016safe}
Remi Munos, Tom Stepleton, Anna Harutyunyan, and Marc Bellemare.
\newblock Safe and efficient off-policy reinforcement learning.
\newblock In {\em Advances in Neural Information Processing Systems}, 2016.

\bibitem{nachum2019dualdice}
Ofir Nachum, Yinlam Chow, Bo~Dai, and Lihong Li.
\newblock Dualdice: Behavior-agnostic estimation of discounted stationary
  distribution corrections.
\newblock In {\em Advances in Neural Information Processing Systems}, 2019.

\bibitem{nie2019learning}
Xinkun Nie, Emma Brunskill, and Stefan Wager.
\newblock Learning when-to-treat policies.
\newblock {\em arXiv preprint arXiv:1905.09751}, 2019.

\bibitem{paine2020hyperparameter}
Tom~Le Paine, Cosmin Paduraru, Andrea Michi, Caglar Gulcehre, Konrad Zolna,
  Alexander Novikov, Ziyu Wang, and Nando de~Freitas.
\newblock Hyperparameter selection for offline reinforcement learning.
\newblock {\em arXiv preprint arXiv:2007.09055}, 2020.

\bibitem{parkerholder2020effective}
Jack Parker-Holder, Aldo Pacchiano, Krzysztof~M Choromanski, and Stephen~J
  Roberts.
\newblock Effective diversity in population based reinforcement learning.
\newblock In {\em Advances in Neural Information Processing Systems}, 2020.

\bibitem{precup2000eligibility}
Doina Precup.
\newblock Eligibility traces for off-policy policy evaluation.
\newblock {\em Computer Science Department Faculty Publication Series}, 2000.

\bibitem{Precup2000}
Doina Precup, Richard~S. Sutton, and Satinder~P. Singh.
\newblock Eligibility traces for off-policy policy evaluation.
\newblock In {\em International Conference on Machine Learning}, 2000.

\bibitem{Rasmussen2005}
Carl~Edward Rasmussen and Christopher K.~I. Williams.
\newblock {\em Gaussian Processes for Machine Learning (Adaptive Computation
  and Machine Learning)}.
\newblock The MIT Press, 2005.

\bibitem{muzero}
Julian Schrittwieser, Ioannis Antonoglou, Thomas Hubert, Karen Simonyan,
  Laurent Sifre, Simon Schmitt, Arthur Guez, Edward Lockhart, Demis Hassabis,
  Thore Graepel, et~al.
\newblock Mastering {A}tari, {G}o, {C}hess and {S}hogi by {P}lanning with a
  {L}earned {M}odel.
\newblock {\em Nature}, 2020.

\bibitem{shahriari2015taking}
Bobak Shahriari, Kevin Swersky, Ziyu Wang, Ryan~P Adams, and Nando De~Freitas.
\newblock Taking the human out of the loop: A review of {Bayesian}
  optimization.
\newblock {\em Proceedings of the IEEE}, 2015.

\bibitem{siegel2020keep}
Noah Siegel, Jost~Tobias Springenberg, Felix Berkenkamp, Abbas Abdolmaleki,
  Michael Neunert, Thomas Lampe, Roland Hafner, Nicolas Heess, and Martin
  Riedmiller.
\newblock Keep doing what worked: Behavior modelling priors for offline
  reinforcement learning.
\newblock In {\em International Conference for Learning Representations}, 2020.

\bibitem{alphago}
David Silver, Aja Huang, Chris~J Maddison, Arthur Guez, Laurent Sifre, George
  Van Den~Driessche, Julian Schrittwieser, Ioannis Antonoglou, Veda
  Panneershelvam, Marc Lanctot, et~al.
\newblock Mastering the game of {Go} with deep neural networks and tree search.
\newblock {\em Nature}, 2016.

\bibitem{Snoek:2012}
J.~Snoek, H.~Larochelle, and R.~P. Adams.
\newblock Practical {Bayesian} optimization of machine learning algorithms.
\newblock In {\em Advances in Neural Information Processing Systems}, 2012.

\bibitem{srinivas2010gaussian}
Niranjan Srinivas, Andreas Krause, Sham Kakade, and Matthias Seeger.
\newblock {G}aussian process optimization in the bandit setting: No regret and
  experimental design.
\newblock In {\em International Conference on Machine Learning}, 2010.

\bibitem{srinivas2009gaussian}
Niranjan Srinivas, Andreas Krause, Sham~M Kakade, and Matthias Seeger.
\newblock Gaussian process optimization in the bandit setting: No regret and
  experimental design.
\newblock {\em arXiv preprint arXiv:0912.3995}, 2009.

\bibitem{sui2015safe}
Yanan Sui, Alkis Gotovos, Joel Burdick, and Andreas Krause.
\newblock Safe exploration for optimization with gaussian processes.
\newblock In {\em International Conference on Machine Learning}, 2015.

\bibitem{sutton2009fast}
Richard~S Sutton, Hamid~Reza Maei, Doina Precup, Shalabh Bhatnagar, David
  Silver, Csaba Szepesv{\'a}ri, and Eric Wiewiora.
\newblock Fast gradient-descent methods for temporal-difference learning with
  linear function approximation.
\newblock In {\em International Conference on Machine Learning}, 2009.

\bibitem{sutton2016emphatic}
Richard~S Sutton, A~Rupam Mahmood, and Martha White.
\newblock An emphatic approach to the problem of off-policy temporal-difference
  learning.
\newblock {\em Journal of Machine Learning Research}, 2016.

\bibitem{tassa2020dmcontrol}
Yuval Tassa, Saran Tunyasuvunakool, Alistair Muldal, Yotam Doron, Siqi Liu,
  Steven Bohez, Josh Merel, Tom Erez, Timothy Lillicrap, and Nicolas Heess.
\newblock dm\_control: Software and tasks for continuous control.
\newblock {\em arXiv preprint arXiv:2006.12983}, 2020.

\bibitem{theocharous2015personalized}
Georgios Theocharous, Philip~S Thomas, and Mohammad Ghavamzadeh.
\newblock Personalized ad recommendation systems for life-time value
  optimization with guarantees.
\newblock In {\em International Joint Conference on Artificial Intelligence},
  2015.

\bibitem{uehara2019minimax}
Masatoshi Uehara, Jiawei Huang, and Nan Jiang.
\newblock Minimax weight and {Q}-function learning for off-policy evaluation.
\newblock In {\em International Conference on Machine Learning}, 2020.

\bibitem{van2016deep}
Hado Van~Hasselt, Arthur Guez, and David Silver.
\newblock Deep reinforcement learning with double {Q}-learning.
\newblock In {\em National Conference on Artificial Intelligence (AAAI)}, 2016.

\bibitem{voloshin2019empirical}
Cameron Voloshin, Hoang~M Le, Nan Jiang, and Yisong Yue.
\newblock Empirical study of off-policy policy evaluation for reinforcement
  learning.
\newblock {\em arXiv preprint arXiv:1911.06854}, 2019.

\bibitem{wang2020critic}
Ziyu Wang, Alexander Novikov, Konrad {\.Z}o{\l}na, Jost~Tobias Springenberg,
  Scott Reed, Bobak Shahriari, Noah Siegel, Josh Merel, Caglar Gulcehre,
  Nicolas Heess, and Nando de~Freitas.
\newblock Critic regularized regression.
\newblock In {\em Advances in Neural Information Processing Systems}, 2020.

\bibitem{yang2020offline}
Mengjiao Yang, Bo~Dai, Ofir Nachum, George Tucker, and Dale Schuurmans.
\newblock Offline policy selection under uncertainty.
\newblock {\em arXiv preprint arXiv:2012.06919}, 2020.

\bibitem{yang2020off}
Mengjiao Yang, Ofir Nachum, Bo~Dai, Lihong Li, and Dale Schuurmans.
\newblock Off-policy evaluation via the regularized lagrangian.
\newblock {\em arXiv preprint arXiv:2007.03438}, 2020.

\bibitem{yu2020mopo}
Tianhe Yu, Garrett Thomas, Lantao Yu, Stefano Ermon, James Zou, Sergey Levine,
  Chelsea Finn, and Tengyu Ma.
\newblock {MOPO}: Model-based offline policy optimization.
\newblock {\em arXiv preprint arXiv:2005.13239}, 2020.

\bibitem{zhang2021autoregressive}
Michael~R Zhang, Tom~Le Paine, Ofir Nachum, Cosmin Paduraru, George Tucker,
  Ziyu Wang, and Mohammad Norouzi.
\newblock Autoregressive dynamics models for offline policy evaluation and
  optimization.
\newblock {\em arXiv preprint arXiv:2104.13877}, 2021.

\end{thebibliography}
}

\clearpage

\appendix

\section{Algorithmic details}
\label{sec:algorithmic_details}

\subsection{GP model}
Following the data generative process in Sec.~\ref{sec:gp}, the hierarchical probabilistic model of all the latent policy values $\mu$, OPE estimates $\rho$, and episodic returns $r$\footnote{We assume Gaussian distribution for the noise model, but if the form of return noise distribution is known in advance, it can replace the Gaussian noise model and approximate GP inference can be performed.} at the given $K$ policies $\{\pi_k\}_{k=1}^K$ is given as follows:
\begin{align}
    \mu_1, \dots, \mu_K &\sim \gauss(m \vone_K, \vK) \nonumber\\
    \rho_k &\overset{i.i.d.}{\sim} \gauss(\mu_k, \sigma_\rho^2), \forall k \in \{1, \dots, K\} \nonumber\\
    r_k^i &\overset{i.i.d.}{\sim} \gauss(\mu_k, \sigma_r^2), \forall i \in \{1, \dots, N_k\} \nonumber\\
    \sigma_\rho^2 &\sim \IG(\alpha, \beta)\nonumber\\
    \sigma_r^2 &\sim \IG(\alpha, \beta)\,,
\end{align}
where the mean $m$ is a constant hyper-parameter and the covariance $\vK(\pi_k, \pi_k')=\mathcal{K}(\pi_k, \pi_{k'})$. The kernel function is defined as
\begin{align}
    \mathcal{K}(\pi_k, \pi_{k'}) &= \sigma_k^2 \exp (-\bar{d}(\pi_k, \pi_{k'}) / l)\nonumber\\
    \bar{d}(\pi_k, \pi_{k'}) &= \mathbb{E}_{s \sim \mathcal{D}'}\ d(\pi_k(s), \pi_{k'}(s))\nonumber\\
    \sigma_k^2 &\sim \IG(\alpha, \beta)\,,
\end{align}
with $\mathcal{D}'$ being a fixed set of states that is randomly subsampled from the entire offline dataset $\mathcal{D}$. $\sigma_k^2$ and $l$ are hyper-parameters. We assume an uninformative prior for $l$.

The posterior distribution of $[\mu_1, \dots, \mu_K]$ given all the observation and hyper-parameters is also a multivariate normal distribution given in \autoref{eq:gp_arm_post}. The log-marginal likelihood of all the hyper-parameters with $\mu_k$'s integrated out is as follows,
\begin{align}
    &\log P(\vrho, \{\vr_k\}_{k}, \sigma_\rho^2, \sigma_r^2, \sigma_k^2 | m, l )\nonumber\\
    &= -\frac{1}{2}\log|\vSigma| -\frac{1}{2} \vy^T \vSigma \vy
      -\frac{1}{2}\left(K\log(\sigma_\rho^2 ) + n \log(\sigma_r^2)
      - \sum_{k=1}^K \log(\vLambda_{kk}) \right) \nonumber\\
    &\quad -\frac{1}{2}\sum_{k=1}^K \left(
     \frac{\rho_k^2}{\sigma_\rho^2} + \frac{1}{\sigma_r^2}\sum_{i=1}^{N_k} (r_k^i)^2 - \frac{\vy_k^2}{\vLambda_{kk}}\right)
     + \log \IG(\sigma_\rho^2)
     + \log \IG(\sigma_r^2)
     + \log \IG(\sigma_k^2)
     + \text{const.}\,,
     \label{eq:gp_arm_marginal_llh}
\end{align}
where $n = \sum_k N_k$ and $\text{const.}$ is a constant.

\subsection{Independent policy value baseline model (\textsc{Ind})}
In Sec.~\ref{sec:exp-setting}, we consider a baseline model with independent policy value distributions as follows,
\begin{align}
    r_k^i &\overset{i.i.d.}{\sim} \gauss(\mu_k, \sigma_{k,r}^2), \forall i, \text{ with }\nonumber\\
    \mu_k &\overset{i.i.d.}{\sim} \gauss(\rho_k, \sigma_{k,\rho}^2), \forall k\nonumber\\
    \sigma_{k,\rho}^2 &\overset{i.i.d.}{\sim} \IG(\alpha, \beta), \forall k\nonumber\\
    \sigma_{k,r}^2 &\overset{i.i.d.}{\sim} \IG(\alpha, \beta), \forall k .
\end{align}

Since the model is i.i.d. across policies, we omit the index $k$ for ease of notation. The posterior distribution of $\mu$ given $\vr = [r^1, \dots, r^N]$, $\sigma_r^2$ and $\sigma_\rho^2$ is then given as
\begin{align}
    \mu &\sim \gauss(m, s^2)\nonumber\\
    m &= \frac{\sum_{i=1}^{N} r^i + (\sigma_r^2 / \sigma_\rho^2) \rho}{N + (\sigma_r^2 / \sigma_\rho^2)}\nonumber\\
    s^2 &= \frac{\sigma_r^2}{N + (\sigma_r^2 / \sigma_\rho^2)}\,.
    \label{eq:ind_arm_post}
\end{align}
The log-marginal likelihood with $\mu$ integrated out can be derived as
\begin{align}
    &\log P(\rho, \vr, \sigma_\rho^2, \sigma_r^2 |)\nonumber\\
    &= -\frac{1}{2}\left(\log(\sigma_\rho^2) + N\log(\sigma_r^2) - \log(s^2)\right)
       -\frac{1}{2}\left(\frac{\rho^2}{\sigma_\rho^2} + \frac{1}{\sigma_r^2}\sum_{i=1}^N (r^i)^2 - \frac{y^2}{s^2}\right) \nonumber\\
    &\quad + \log \IG(\sigma_\rho^2)
     + \log \IG(\sigma_r^2)
     + \text{const.}\,,
    \label{eq:ind_arm_marginal_llh}
\end{align}
where $\text{const.}$ is a constant.

\subsection{Hyperparameter fitting}
We fit the hyperparameters in both models whenever a new observation is obtained by searching for the maximum a posteriori (MAP) estimate using a few steps of gradient descent on the log-marginal likelihood.

Fitting the hyper-parameters at the beginning with few observations is challenging.
When we do not use OPE to warm start GP inference such as in \GPUCB{} and \GPUniform{}, we start by randomly selecting the first \num{5} policies to execute. Besides, it is important in this case to apply a prior for hyperparameters to prevent variance from shrinking towards zero. When OPE is available, we find neither of these techniques is required for a robust estimation and we can start fitting the hyperparameters of GP from the first step.

\subsection{Hyperparameter values}
The initial values of the hyperparameters of GP and our method in general are listed next. 
We set the GP hyperparameters based on the statistics about the policy returns from the offline dataset.
The same hyperparameters are used for all the domains. 
The discounted returns in Atari domain are very different in scale from the other domains.
To ensure that the same hyperparameters are applicable to all domains,  we rescale returns in Atari to the same range of means and variances as other domains based on the statistics from the datasets for offline RL.
\begin{itemize}
    \item We find using a constant value $\sqrt{\beta_i}=5$ in GP-UCB works well across different budgets, Bayesian models, and environments, although a more sophisticated scheduling \citep{srinivas2009gaussian,audibert2010best,hoffman2014correlation} could further improve the results.
    \item The hyperparameters of the prior for \textsc{Ind} were set to: $\alpha=1, \beta=1000$.
    \item Observation noise variance of GP was set to the initial value of $1000$ or prior with $\alpha=1, \beta=200$ when the prior is used. Initial value of the offset $m$ was $0$.
    \item For GP optimisation we used Adam optimiser with learning rate $0.001$ and $1000$ optimisation steps per observation update. $\beta_1=0.9, \beta_2=0.999$.
    \item To compensate to the non-zero mean in the GP, we use a constant component in the kernel with trainable variance parameter initially set to $10$. When the prior is used, we set $\alpha=1, \beta=200$ in all the experiments.
\end{itemize}

\subsection{Kernel selection}
In our preliminary experiments, we experimented with a few ways to compute the kernel over policies. 
To quickly select promising ideas, we tested them in a simple regression task where we split all policies with their true returns into training and testing sets and predict the returns of the test policies based on the training policies. 
Then, we select the kernel that gives the highest log likelihood and lowest squared prediction error.
We choose the best type of kernel using 3 \dmcontrol{} environments (\verb|cartpole_swingup|, \verb|fish_swim| and \verb|manipulator_insert_ball|) and apply the same kernel to all tasks of both domains in the experiments. 
The most important variations of the kernel that we studied are enumerated below. We highlight the variant that was used in the final experiments.
\begin{itemize}
    \item As a subset $\mathcal{D}'$ of states for computing policy's actions, we select \num{1000} states. We tried to use 1) \emph{random states}, 2) initial states, 3) states with high rewards, 4) states with low rewards, 5) states in the trajectories with high returns, 6) states in the trajectories with low returns. Several options looked promising, including 1, 2, 4. At the end we chose the simplest option that yielded good results (1).
    \item As a way to construct a kernel from the policy actions on several states, we considered: 1) \emph{multiplication of kernels on separate states}, 2) concatenating action predictions into a single feature vector. We chose kernel multiplication because we found in our experiments that concatenating the action in a single vector makes the distances not very informative.  This happens because the dimensionality of the policy representation becomes very high.
    \item After obtaining the distances matrix for the policies, we compute the kernel. We tried 1) RBF, 2) Mat\'ern $5/2$, 3) Mat\'ern $3/2$, 4) \emph{Mat\'ern $1/2$} kernels. While the difference between these versions was not large, the less smooth kernel Mat\'ern $1/2$ performed slightly better and reflected our intuition that the return of policies trained with different algorithms can change quickly (\autoref{fig:kernel}, right).
    
\end{itemize}

\subsection{Bayesian optimization algorithm}
The pseudo-code of the Bayesian optimization algorithm for both models is presented in \autoref{alg:algo}.

\begin{algorithm}[tbh!]
\SetAlgoLined
\KwData{Environment $\mathrm{env}$, set of policies $\{\pi_k\}$, $1\leq k \leq K$, number of initial random samples $n_\mathrm{init}$, total number of steps $n$, (optional) OPE estimates $\{\rho_k\}$}
\KwResult{Final recommendation $\hat{k}$}
 \For{$i: 1 \rightarrow n$}{
  \eIf{$i\leq n_\mathrm{init}$}{
  Sample a policy $k_i$ uniformly from $\{1, 2, \dots, K\}$
  }
  {
  Fit hyper-parameters by maximizing the log-marginal likelihood (\autoref{eq:gp_arm_marginal_llh} or \autoref{eq:ind_arm_marginal_llh}) with all the return samples $r_i$ and OPE estimates $\rho_k$\;
  Update the posterior of $\vmu$ using \autoref{eq:gp_arm_post} or \autoref{eq:ind_arm_post}\;
  Compute the UCB acquisition function $u_i(k)$ for every policy $k$ using \autoref{eq:aquisition}\;
  Choose a policy $k_i = \arg\max_k u_i(k)$\;
  }
  Apply policy $\pi_{k_i}$ in $\mathrm{env}$ for one episode and receive the discounted episodic return $r_i$\;
 }
 Maximize the log-marginal likelihood (\autoref{eq:gp_arm_marginal_llh} or \autoref{eq:ind_arm_marginal_llh}) wrt the hyper-parameters using Adam\;
  Update the posterior of $\vmu$ using \autoref{eq:gp_arm_post} or \autoref{eq:ind_arm_post}\;
 \KwRet{$\hat{k}=\arg\max_k \vm_k$ with $\vm_k$ being the posterior mean of $k$-th policy}
 \caption{UCB algorithm for active offline policy selection}
 \label{alg:algo}
\end{algorithm}
\section{Domains details}
\label{sec:domain_details}

In this section, we present the statistics about our domains of interest: dm-control, MPG and Atari. 
We describe how we obtain the policies and present the statistics about the returns. 

\begin{figure}[t]
    \centering
    \includegraphics[width=\textwidth]{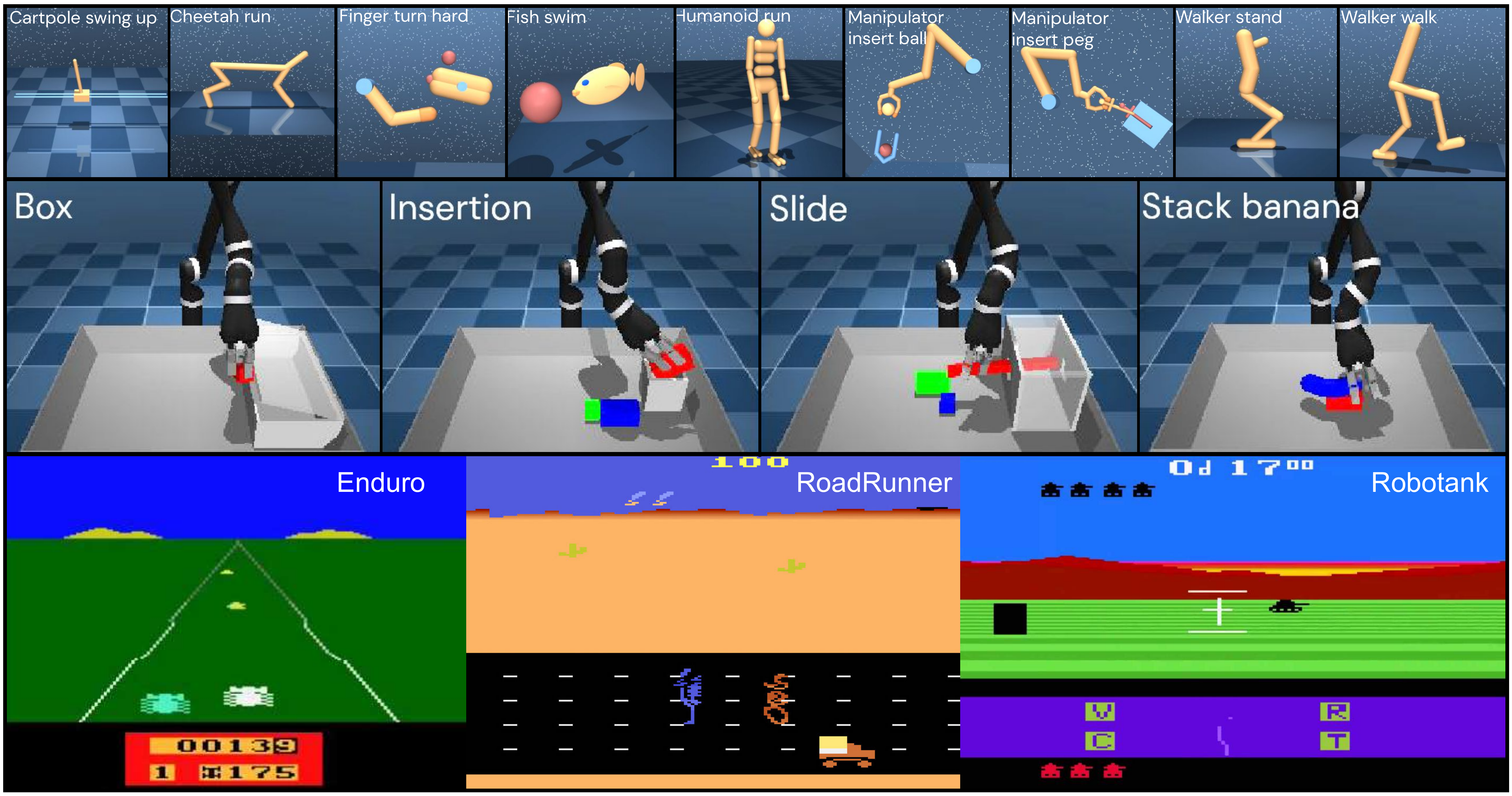}
    \caption{Environments used in the experiments. Top: \num{9} environments from the \dmcontrol{} domain. Middle: \num{4} environments from the \mpg{} domain. Bottom: \num{3} game environments from Atari domain.}
    \label{fig:all-environments}
\end{figure}

\subsection{Dm-control}
\label{sec:dm_control_details}
In \dmcontrol{} domain we use policies from the DOPE benchmark \citep{fu2021benchmarks}, which results in a diverse set of policies with varying algorithms (D4PG, ABM, CRR), hyperparameters like learning rate, and number of training steps, giving up to 96 policies per task, and a diverse range of returns. There are \num{9} environments as shown in the top row of \autoref{fig:all-environments}: \verb|cartpole_swingup|, \verb|cheetah_run|, \verb|finger_turn_hard|, \verb|fish_swim|, \verb|humanoid_run|, \verb|manipulator_insert_ball|, \verb|manipulator_insert_peg|, \verb|walker_stand|, \verb|walker_walk|.
There is a significant gap between the scores of the badly-performing and well-performing policies. 
\autoref{tab:dmcontrolstats} shows the standard deviation, minimum and maximum returns of the policies from the dataset.

\begin{table}[t!]
    \caption{Statistics about the policies in \dmcontrol{}: the return of the worst policy, of the best policy and standard deviation of the policies returns.}
    \vspace{2mm}
    \centering
    {\small
    \begin{tabular}{l|r|r|r}
    task & stdev & min return & max return \\
    \midrule
    \verb|cartpole_swingup| & \num{6.488} & \num{0.451} & \num{31.504} \\
    \verb|cheetah_run| & \num{5.504} & \num{6.933} & \num{36.291} \\
    \verb|finger_turn_hard| & \num{14.055} & \num{8.613} & \num{60.971} \\
    \verb|fish_swim| & \num{3.208} & \num{8.097}  & \num{18.755} \\
    \verb|humanoid_run| & \num{8.993} & \num{0.566} & \num{30.117} \\
    \verb|manipulator_insert_ball| & \num{2.825} & \num{3.668} & \num{13.801} \\
    \verb|manipulator_insert_peg| & \num{2.306} & \num{4.668} & \num{11.249} \\
    \verb|walker_stand| & \num{13.173} & \num{39.180} & \num{85.533} \\
    \verb|walker_walk| & \num{13.973} & \num{11.950} & \num{74.719} \\
    \end{tabular}
    }
    \label{tab:dmcontrolstats}
\end{table}

\begin{figure}[t!]
    \begin{subfigure}[b]{0.25\textwidth}
        \includegraphics[width=1\linewidth]{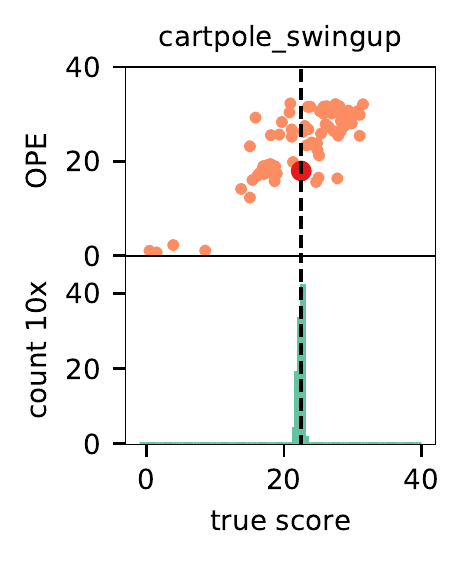}
    \end{subfigure}
    ~
    \begin{subfigure}[b]{0.25\textwidth}
        \includegraphics[width=1\linewidth]{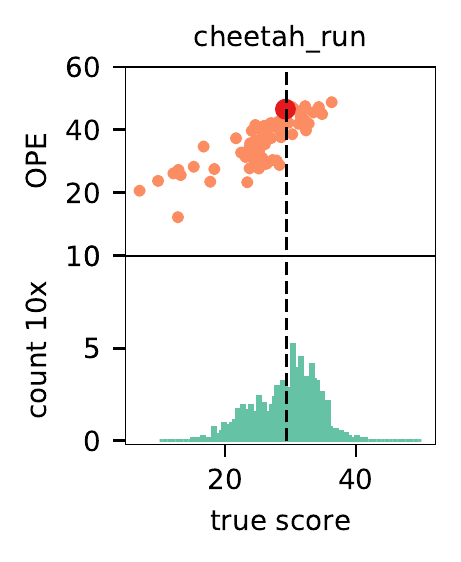}
    \end{subfigure}
    ~
    \begin{subfigure}[b]{0.25\textwidth}
        \includegraphics[width=1\linewidth]{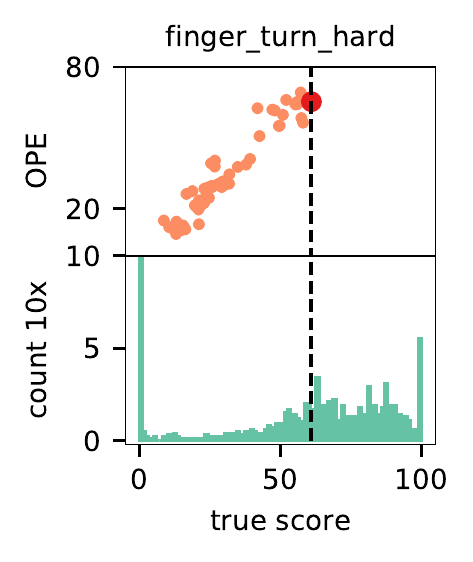}
    \end{subfigure}
    \\
    \begin{subfigure}[b]{0.25\textwidth}
        \includegraphics[width=1\linewidth]{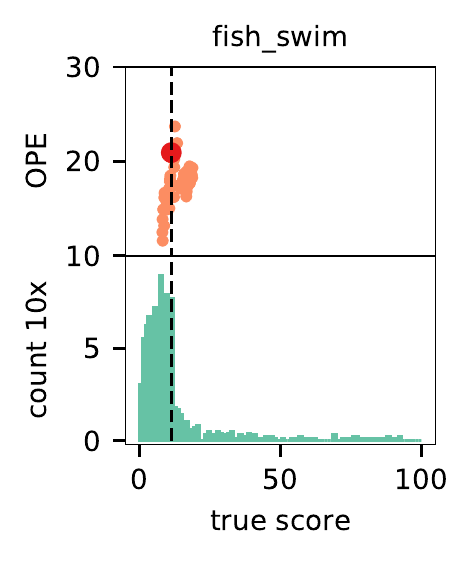}
    \end{subfigure}
    ~
    \begin{subfigure}[b]{0.25\textwidth}
        \includegraphics[width=1\linewidth]{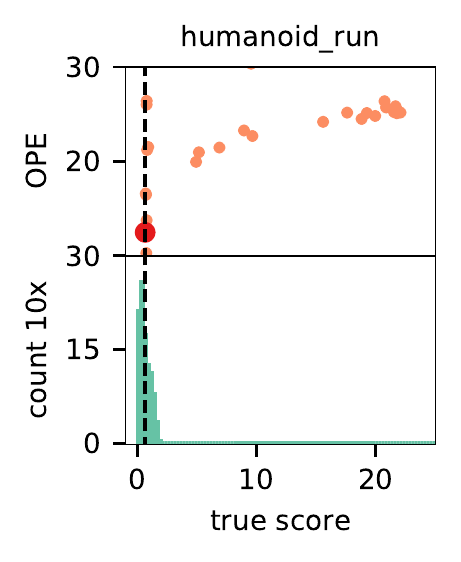}
    \end{subfigure}
    ~
    \begin{subfigure}[b]{0.25\textwidth}
        \includegraphics[width=1\linewidth]{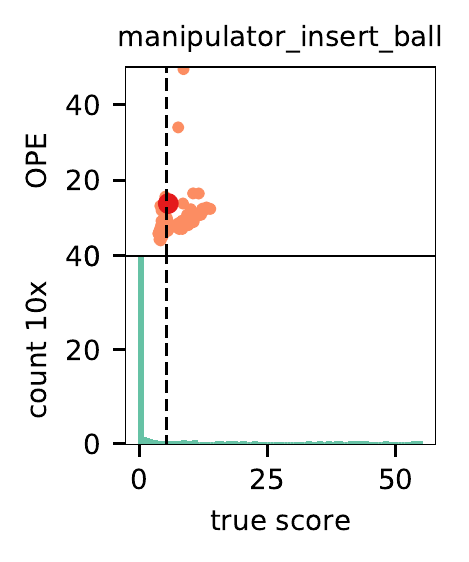}
    \end{subfigure}
    \\
    \begin{subfigure}[b]{0.25\textwidth}
        \includegraphics[width=1\linewidth]{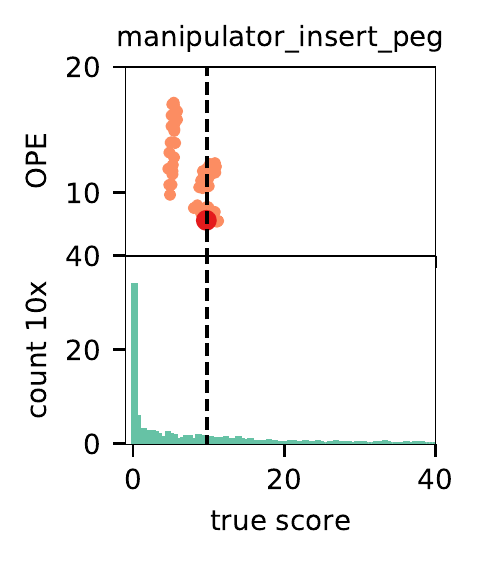}
    \end{subfigure}
    ~
    \begin{subfigure}[b]{0.25\textwidth}
        \includegraphics[width=1\linewidth]{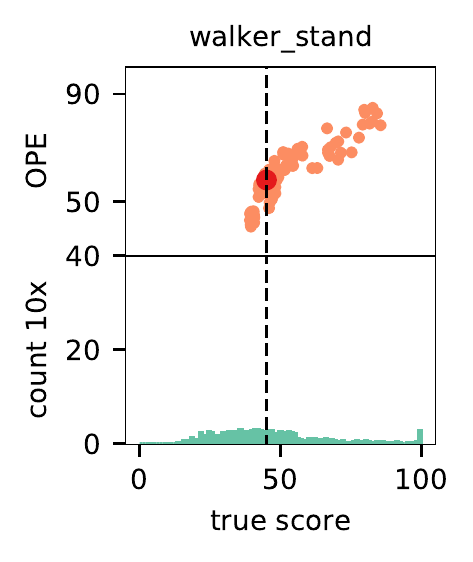}
    \end{subfigure}
    ~
    \begin{subfigure}[b]{0.25\textwidth}
        \includegraphics[width=1\linewidth]{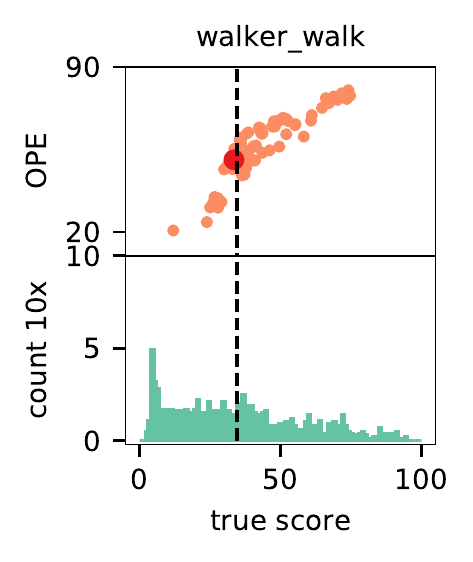}
    \end{subfigure}
    \\
    \caption{For each environment in \dmcontrol{} we show 1) the distribution of policy OPE metrics as a function of true scores at the top (in orange), and 2) the histogram of the episodic returns for a single randomly selected policy (in green) at the bottom. The randomly selected policy is highlighted with a red dot on the top and it is connected by a dashed line to the mean values in the histogram.}
    \label{fig:stats-dmcontrol}
\end{figure}

The success of the active policy selection strategy in each experiment depends on the relation between the variance of the returns of the policies in the dataset and the variance of the returns that a single policy renders. 
To understand this relationship better in \autoref{fig:stats-dmcontrol} we show for each environment: 1) the distribution of policy OPE metrics as a function of true scores at the top, and 2) the histogram of the episodic returns for a single randomly selected policy (highlighted and connected by a dashed line) at the bottom. 
We observe that in some cases (for example, in \verb|finger_turn_hard|, \verb|fish_swim|, \verb|walker_walk|) the variance of episodic returns is greater than the variance of the policies true scores. 
This is particularly challenging for any policy selection that involves the online component.
Furthermore, we also observe that in some cases OPE scores are not well correlated with the true scores (for example, \verb|fish_swim|, \verb|humanoid_run|, \verb|manipulator_insert_ball|, \verb|manipulator_insert_peg|). 
In this case, on one hand, relying on OPE scores makes policy selection more challenging, but on the other hand, it leaves the biggest potential for improvement by involving the online evaluation. 

\subsection{MPG}
\label{sec:mpg_details}
To generate policies for the environments in \mpg{} domain we follow the protocol in \citep{paine2020hyperparameter}. 
For each task, we generate up to 256 policies trained with various algorithms (BC, D4PG, and CRR), and hyperparameters including number of hidden units (\num{64}, \num{1024}), number of blocks (\num{1}, \num{5}), learning rate (\num{1e-3}, \num{1e-5}), learning steps (\num{50}k, \num{250}k, \num{25}k), and for CRR the beta parameter (\num{0.1}, \num{10}). This ensures there is a diverse range of policies that span the range of possible returns.
There are four tasks in \mpg{} domain: \verb|box|, \verb|insertion|, \verb|slide|, and \verb|stack_banana| as shown in the middle row of \autoref{fig:all-environments}.
We include MPG as a challenging domain where policies are learnt from high dimensional images and training online policy takes thousands of trajectories (around \num{8000}) even when human demonstrations are included. 

The policies in this domain also exhibit a significant gap and large variance in the scores of the badly-performing and well-performing policies as it is demonstrated by the statistics in \autoref{tab:mpgstats}. 
As this is a challenging domain, in all the environments, the worst policy achieves \num{0} (lowest possible) regret.

\begin{figure}[t!]
    \begin{subfigure}[b]{0.25\textwidth}
        \includegraphics[width=1\linewidth]{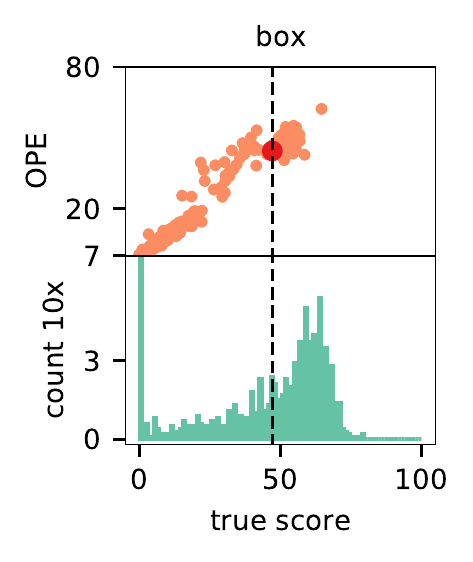}
    \end{subfigure}
    ~
    \begin{subfigure}[b]{0.25\textwidth}
        \includegraphics[width=1\linewidth]{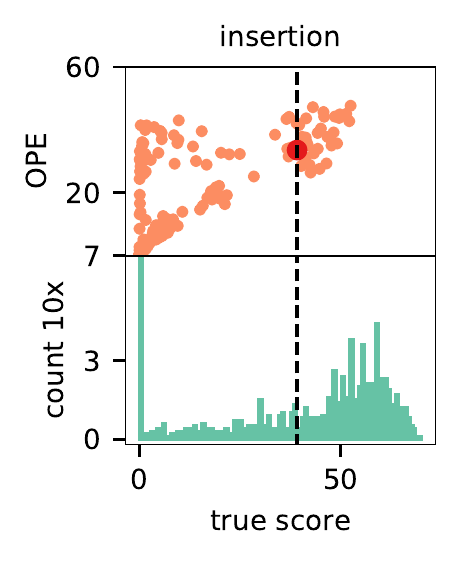}
    \end{subfigure}
    ~
    \begin{subfigure}[b]{0.25\textwidth}
        \includegraphics[width=1\linewidth]{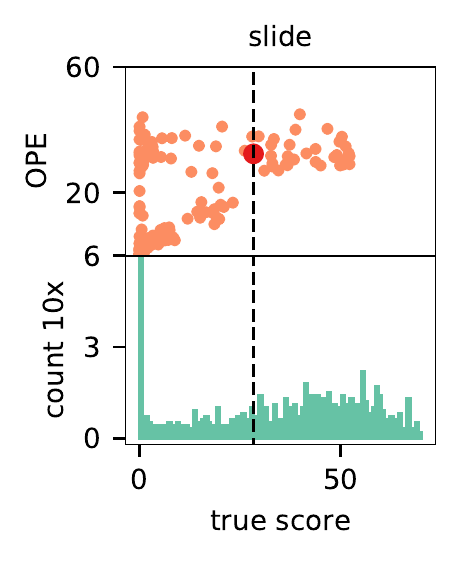}
    \end{subfigure}
    \\
    \begin{subfigure}[b]{0.25\textwidth}
        \includegraphics[width=1\linewidth]{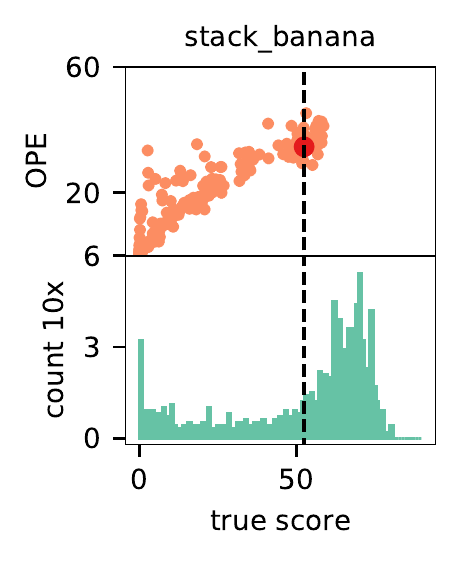}
    \end{subfigure}
    \\
    \caption{For each environment in \mpg{} we show 1) the distribution of policy OPE metrics as a function of true scores at the top (in orange), and 2) the histogram of the episodic returns for a single randomly selected policy (in green) at the bottom. The randomly selected policy is highlighted with a red dot on the top and it is connected by a dashed line to the mean values in the histogram.}
    \label{fig:stats-mpg}
\end{figure}

In \autoref{fig:stats-mpg} we study again the relationship between the distribution of policy OPE metrics and true scores, as well as the distribution of the episodic returns for a random policy. 
In all environments, the variance of episodic returns is comparable or even greater than the variance of the policies true scores. 
In some cases, like in \verb|box| and \verb|stack_banana| the OPE scores are almost perfectly correlated with the true returns, which makes the OPE baseline very hard to beat.
Even when OPE scores are not perfectly correlated with the true returns, like in the \verb|insertion| environment, it may still be very informative for the task of selecting the \emph{best} policy. 
There is the biggest potential for improvement over OPE in \verb|slide| environment. 
Also, notice that many policies have a peak value in the histogram at value \num{0} (including policies with high average return). 
This property makes the sparse online evaluation particularly challenging.

\begin{table}[t!]
    \caption{Statistics about the policies in \mpg{}: the return of the worst policy, of the best policy and standard deviation of the policies returns.}
    \vspace{2mm}
    \centering
    {\small
    \begin{tabular}{l|r|r|r}
    task & stdev & min return & max return \\
    \midrule
    \verb|box| & \num{19.711} & \num{0.0} & \num{64.58} \\
    \verb|insertion| & \num{15.808} & \num{0.0} & \num{52.42} \\
    \verb|slide| & \num{14.622} & \num{0.0} & \num{52.17} \\
    \verb|stack_banana| & \num{18.420} & \num{0.0}  & \num{58.74} \\
    \end{tabular}
    }
    \label{tab:mpgstats}
\end{table}

\subsection{Atari}
\label{sec:atari_details}
We choose Atari in order to test our method on an established discrete control offline RL domain where training an online policy from pixels is challenging and requires thousands of trajectories.
We follow the protocol in \cite{gulcehre2020rl} to train \num{210} policies with different architectures and hyperparameter for each of three Atari games: \verb|Enduro|, \verb|RoadRunner|, and \verb|Robotank| as shown in~\autoref{fig:all-environments}, bottom row.
\verb|Enduro| is previously used as an OPE benchmark~\citep{voloshin2019empirical}. 
We include \verb|RoadRunner| and \verb|Robotank| to cover a range of difficulty and reward scales presented in the Atari benchmark.
The policies cover a diverse range of methods and this results in the most diverse set of policies for the experiments in this paper: DDQN~\citep{van2016deep}, BCQ~\citep{fujimoto2019off}, REM~\citep{agarwal2019optimistic}, IQN~\citep{dabney2018implicit}, CQL~\citep{kumar2020conservative}, and BVE~\citep{gulcehre2021regularized}.
We use grid search over the following hyperparameters: learning rate for all algorithm (\num{3e-5}, \num{1e-4}, \num{3e-4}), the pessimism regularizer in CQL (\num{1e-3}, \num{1e-2}, \num{1e-1}, \num{1}, \num{2}), the threshold parameter in BCQ (\num{0.1}, \num{0.25}, \num{0.5}, \num{0.75}). 
Each algorithm with a hyperparameter setting is trained on \num{5} datasets generated by DQN snapshots trained with five different seeds as described in \citep{agarwal2019optimistic}. 

\begin{figure}[t!]
    \begin{subfigure}[b]{0.25\textwidth}
        \includegraphics[width=1\linewidth]{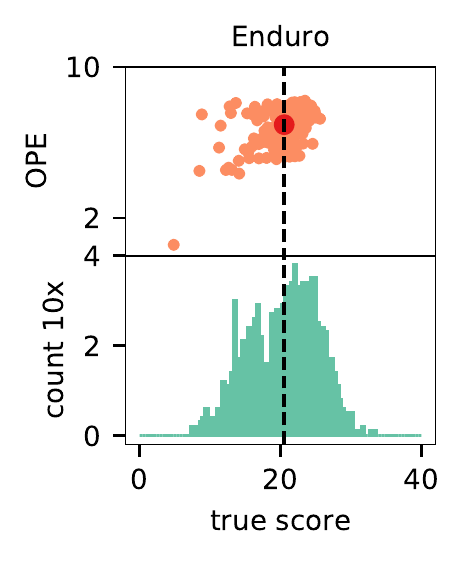}
    \end{subfigure}
    \begin{subfigure}[b]{0.25\textwidth}
        \includegraphics[width=1\linewidth]{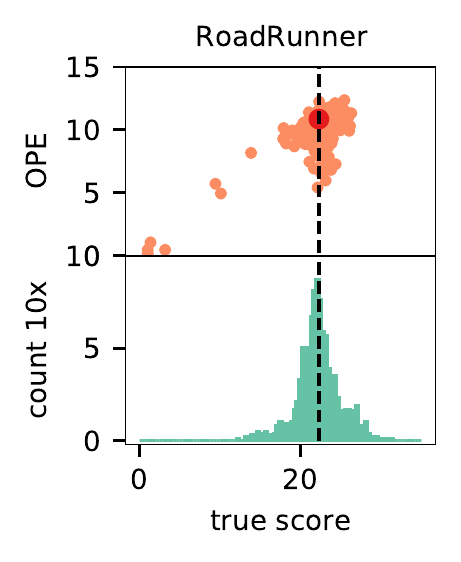}
    \end{subfigure}
    \begin{subfigure}[b]{0.25\textwidth}
        \includegraphics[width=1\linewidth]{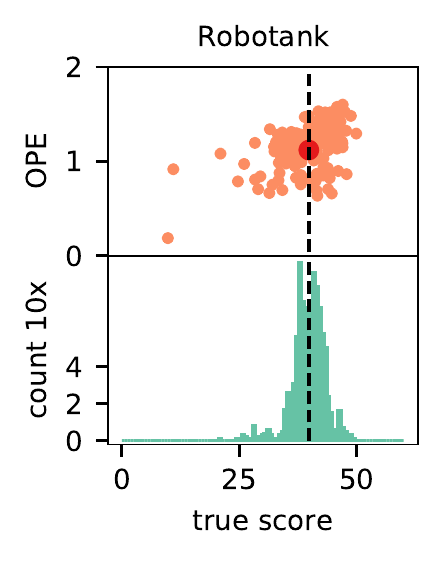}
    \end{subfigure}

    \caption{For each environment in Atari we show 1) the distribution of policy OPE metrics as a function of true scores at the top (in orange), and 2) the histogram of the episodic returns for a single randomly selected policy (in green) at the bottom. The randomly selected policy is highlighted with a red dot on the top and it is connected by a dashed line to the mean values in the histogram.}
    \label{fig:stats-atari}
\end{figure}

As in other domains, there is a significant gap between the scores of the badly-performing and well-performing policies as indicated in the \autoref{tab:ataristats}. 
In \autoref{fig:stats-atari} we show the relationship between the distribution of policy OPE metrics and true scores, as well as the distribution of the episodic returns for one of the policies.
The OPE scores are correlated, but the correlation is far from perfect and there is space for improvement by including the online trajectories.

\begin{table}[t!]
    \caption{Statistics about the policies in Atari: the return of the worst policy, of the best policy and standard deviation of the policies returns.}
    \vspace{2mm}
    \centering
    {\small
    \begin{tabular}{l|r|r|r}
    task & stdev & min return & max return \\
    \midrule
    \verb|Enduro| & \num{3.198} & \num{4.848} & \num{25.617} \\
    \verb|Robotank| & \num{5.509} & \num{9.786} & \num{49.898} \\
    \verb|RoadRunner| & \num{3.867} & \num{1.012} & \num{26.298} \\
    \end{tabular}
    }
    \label{tab:ataristats}
\end{table}

\section{Results by task}

While \autoref{sec:exp-results} and \autoref{sec:exp-ablations} present the aggregated results across all tasks where each task is normalised by the gap between policies, in this section we present figures for each task separately. 
We discuss the variability of the results across different environments and explain the reasons for this variability.

\subsection{Dm-control}

\begin{figure}[tbph!]
    \begin{subfigure}[b]{0.28\textwidth}
        \includegraphics[width=1\linewidth,left]{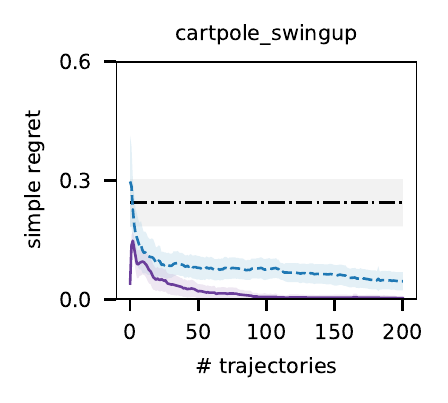}
    \end{subfigure}
    \begin{subfigure}[b]{0.28\textwidth}
        \includegraphics[width=1\linewidth,left]{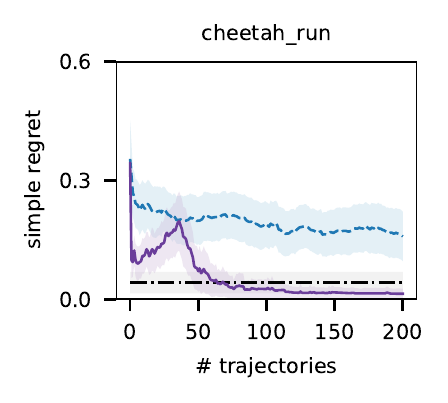}
    \end{subfigure}
    \begin{subfigure}[b]{0.28\textwidth}
        \includegraphics[width=1\linewidth,left]{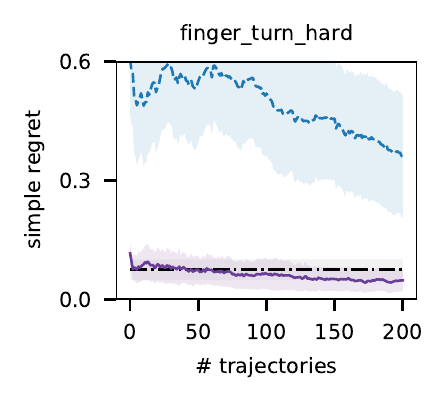}
    \end{subfigure}
    \\
    \begin{subfigure}[b]{0.28\textwidth}
        \includegraphics[width=1\linewidth,left]{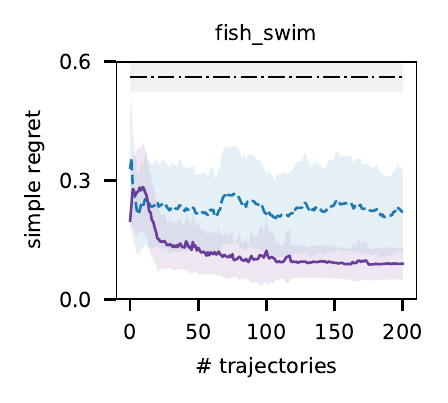}
    \end{subfigure}
    \begin{subfigure}[b]{0.28\textwidth}
        \includegraphics[width=1\linewidth,left]{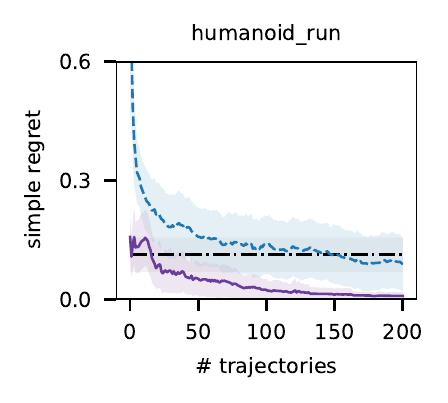}
    \end{subfigure}
    \begin{subfigure}[b]{0.28\textwidth}
        \includegraphics[width=1\linewidth,left]{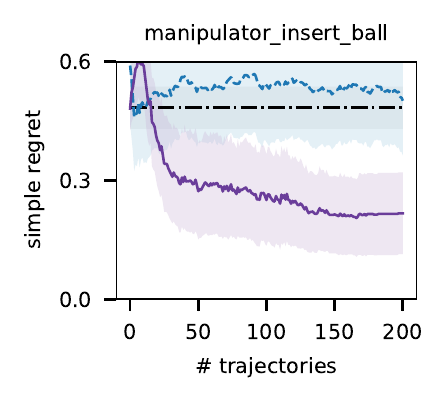}
    \end{subfigure}
    \\
    \begin{subfigure}[b]{0.28\textwidth}
        \includegraphics[width=1\linewidth,left]{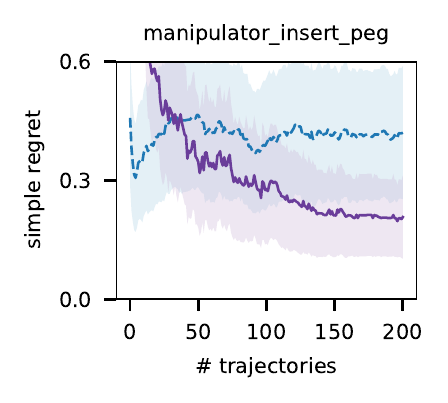}
    \end{subfigure}
    \begin{subfigure}[b]{0.28\textwidth}
        \includegraphics[width=1\linewidth,left]{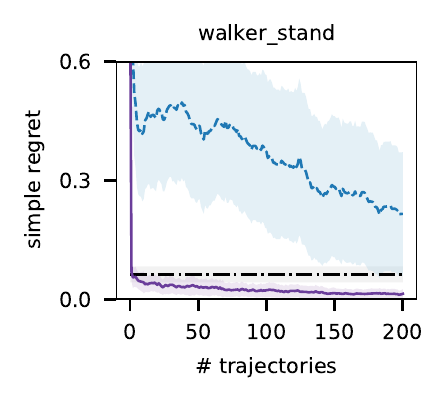}
    \end{subfigure}
    \begin{subfigure}[b]{0.28\textwidth}
        \includegraphics[width=1.55\linewidth,left]{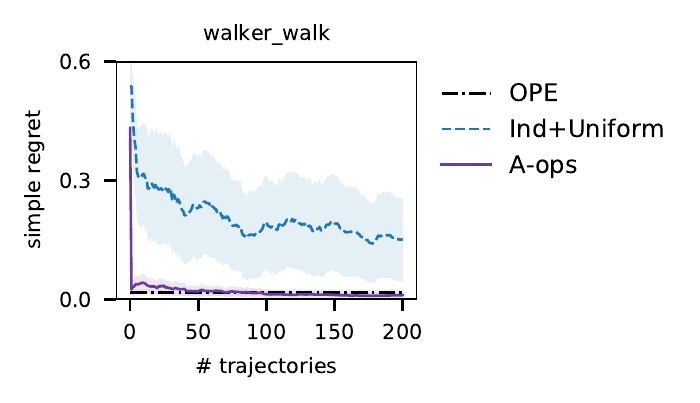}
    \end{subfigure}
    \caption{Comparison of \ours{} approach, offline policy selection with \OPE{}, and online policy selection with \IndUniform{} on each environment in \dmcontrol{} domain. The results are averaged across \num{100} experiments. We do not show OPE performance in \texttt{manipulator}\_\texttt{insert}\_\texttt{peg} in order to respect the y-axis limits. The value is \num{0.88} as D4PG policies are highly overestimated. Shaded areas show standard deviation of the mean. Our method \ours{} quickly surpasses the baselines and achieves the lowest regrets. }
    \label{fig:results-by-task}
\end{figure}

Similar to \autoref{fig:results}, \autoref{fig:results-by-task} shows the comparison between our proposed method \ours{}, completely offline policy selection with OPE and completely online selection with \IndUniform{} on each of \num{9} task from \dmcontrol{}. 
Depending on the quality of the initial OPE values and the variance of the policy returns (see \autoref{fig:stats-dmcontrol}), \ours{} may take different number of trajectories before it outperforms all the baselines, but usually it only takes a few steps.
Eventually, \ours{} reaches the best performance with the limited budget.

\begin{figure}[tbph!]
    \begin{subfigure}[b]{0.28\textwidth}
        \includegraphics[width=1\linewidth]{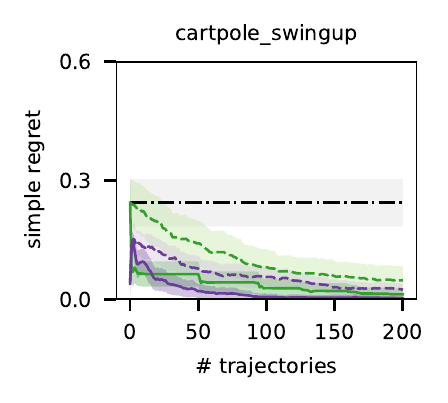}
    \end{subfigure}
    \hspace{0em}
    \begin{subfigure}[b]{0.28\textwidth}
        \includegraphics[width=1\linewidth]{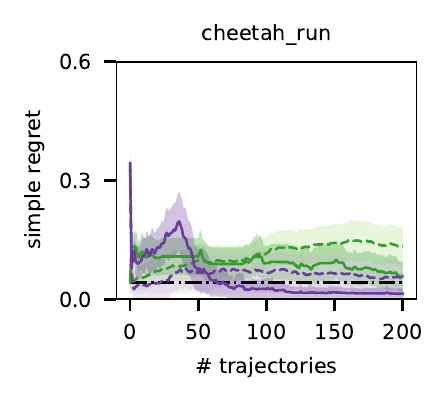}
    \end{subfigure}
    \hspace{0em}
    \begin{subfigure}[b]{0.28\textwidth}
        \includegraphics[width=1\linewidth]{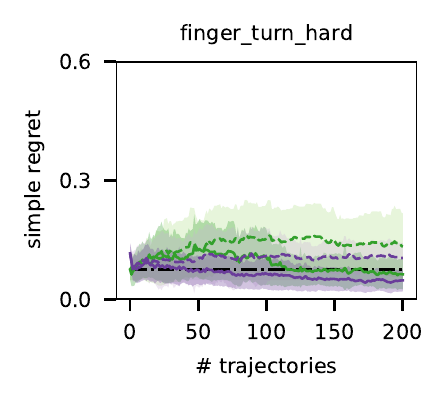}
    \end{subfigure}
    \\
    \begin{subfigure}[b]{0.28\textwidth}
        \includegraphics[width=1\linewidth]{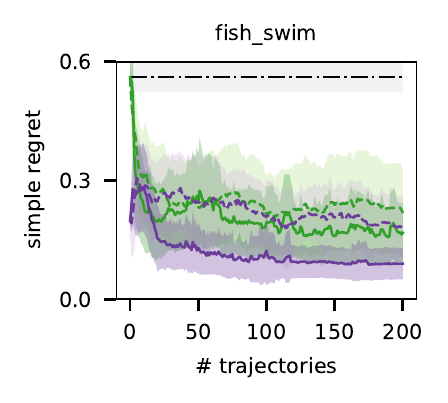}
    \end{subfigure}
    \hspace{0em}
    \begin{subfigure}[b]{0.28\textwidth}
        \includegraphics[width=1\linewidth]{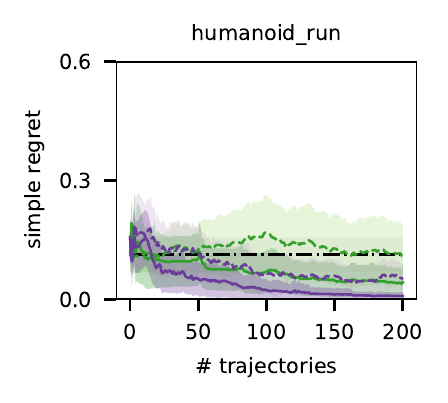}
    \end{subfigure}
    \hspace{0em}
    \begin{subfigure}[b]{0.28\textwidth}
        \includegraphics[width=1\linewidth]{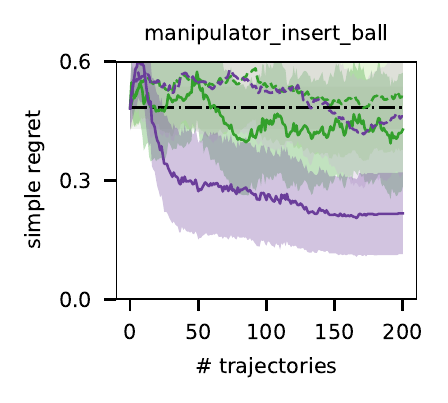}
    \end{subfigure}
    \\
    \begin{subfigure}[b]{0.28\textwidth}
        \includegraphics[width=1\linewidth]{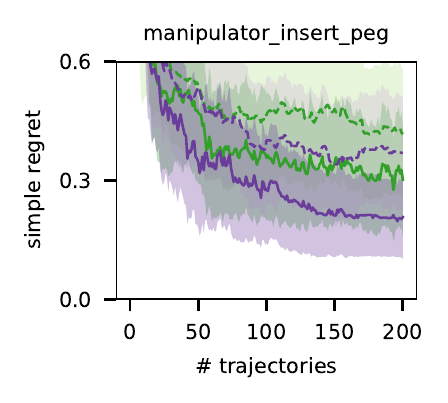}
    \end{subfigure}
    \hspace{0em}
    \begin{subfigure}[b]{0.28\textwidth}
        \includegraphics[width=1\linewidth]{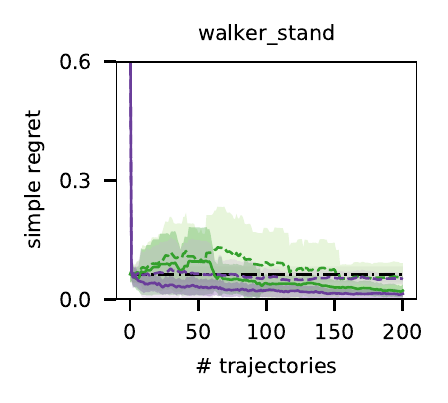}
    \end{subfigure}
    \hspace{0em}
    \begin{subfigure}[b]{0.28\textwidth}
        \includegraphics[width=1.55\linewidth]{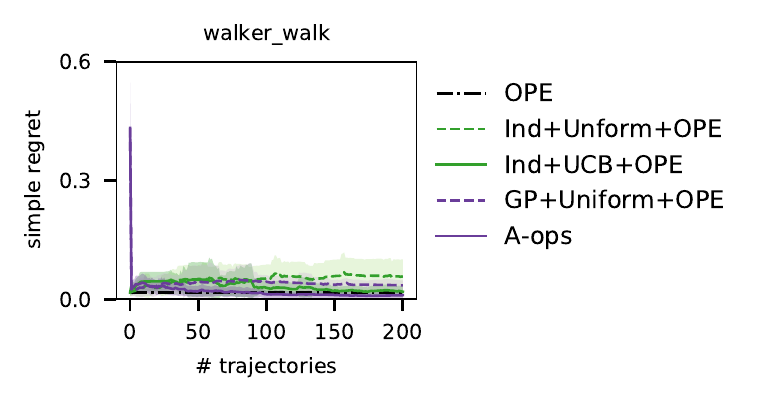}
    \end{subfigure}
    \caption{Ablations. In \num{9} environments of \dmcontrol{} domain, we ablate two components of our approach when using OPE metrics. 1) The choice of policy model: GP (purple) and independent (green). 2) The policy selection strategy: active (dark, solid line) and uniform sampling (bright, dashed line). 
    We do not show OPE performance in \texttt{manipulator}\_\texttt{insert}\_\texttt{peg} in order to respect the y-axis limits. The value is \num{0.88} as D4PG policies are highly overestimated.
    In general, active is better than uniform, methods that exploit policy correlation are better than methods that do not. }
    \label{fig:results_ablations-by-task}
\end{figure}

Similar to \autoref{fig:results_ablations} (first row), \autoref{fig:results_ablations-by-task} shows the contribution of each of the components of the method. 
Our method \ours{} is preferable in all environments across a wide range of interaction budget except for \verb|cheetah_run| with less than \num{50} trajectories. 
Again we observe that modelling correlated policies as in \textsc{GP} performs better than modelling independent policies as in \textsc{Ind}, active policy selection as in \textsc{UCB} is better than uniform policy selection as in \textsc{Uniform}.
In manipulator tasks, no method achieves a regret as low as in other tasks. 
We believe that the main reasons for this are 1) low performance of the initial OPE estimates and 2) the skewed distribution of episodic returns of all policies where most returns are close to \num{0} (see \autoref{fig:stats-dmcontrol}). 

\begin{figure}[ht!]
    \begin{subfigure}[b]{0.28\textwidth}
        \includegraphics[width=1\linewidth]{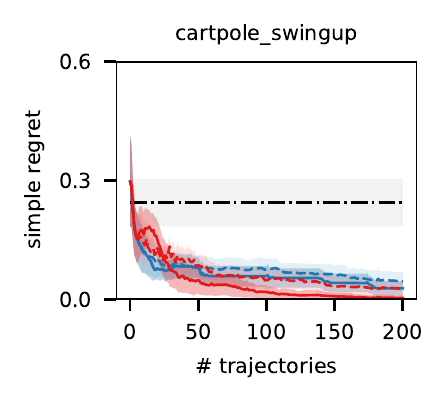}
    \end{subfigure}
    \hspace{0em}
    \begin{subfigure}[b]{0.28\textwidth}
        \includegraphics[width=1\linewidth]{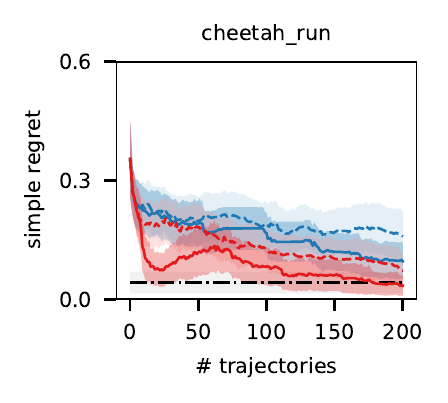}
    \end{subfigure}
    \hspace{0em}
    \begin{subfigure}[b]{0.28\textwidth}
        \includegraphics[width=1\linewidth]{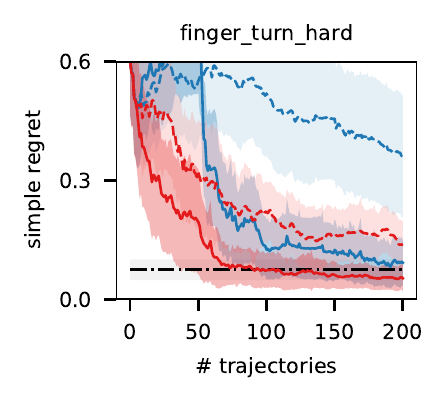}
    \end{subfigure}
    \\
    \begin{subfigure}[b]{0.28\textwidth}
        \includegraphics[width=1\linewidth]{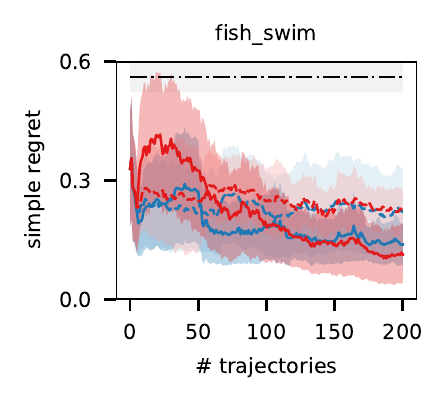}
    \end{subfigure}
    \hspace{0em}
    \begin{subfigure}[b]{0.28\textwidth}
        \includegraphics[width=1\linewidth]{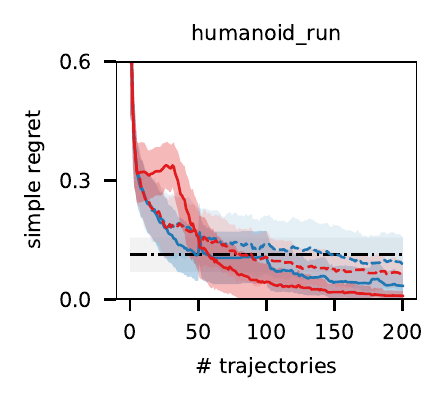}
    \end{subfigure}
    \hspace{0em}
    \begin{subfigure}[b]{0.28\textwidth}
        \includegraphics[width=1\linewidth]{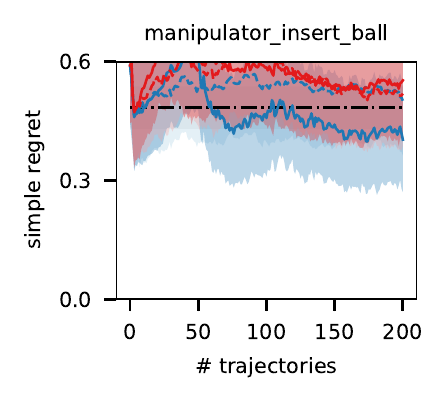}
    \end{subfigure}
    \\
    \begin{subfigure}[b]{0.28\textwidth}
        \includegraphics[width=1\linewidth]{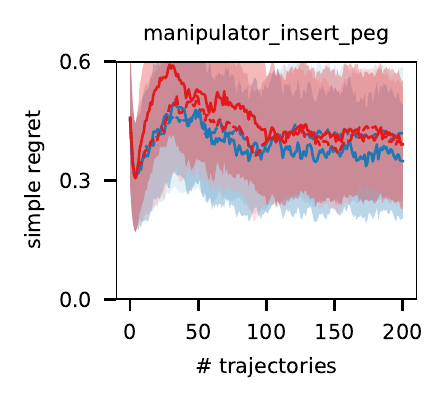}
    \end{subfigure}
    \hspace{0em}
    \begin{subfigure}[b]{0.28\textwidth}
        \includegraphics[width=1\linewidth]{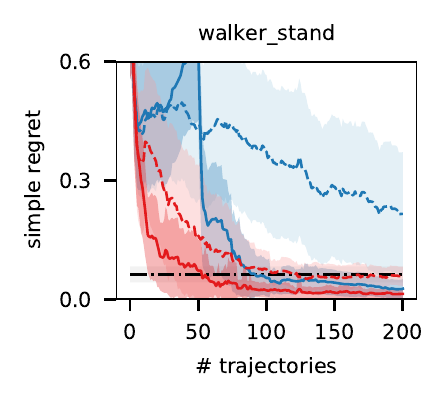}
    \end{subfigure}
    \hspace{0em}
    \begin{subfigure}[b]{0.28\textwidth}
        \includegraphics[width=1.55\linewidth]{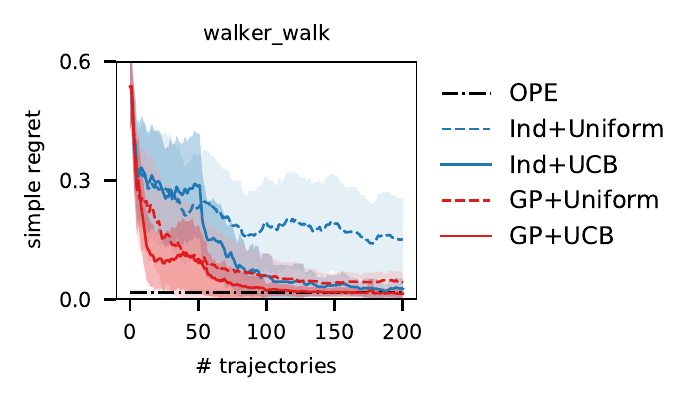}
    \end{subfigure}
    \caption{Ablations. In \num{9} environments of \dmcontrol{} domain, we ablate two components of our approach when NOT using OPE metrics. 1) The choice of policy model: GP (red) and independent (blue). 2) The policy selection strategy: active (dark, solid line) and uniform sampling (bright, dashed line). 
    We do not show OPE performance in \texttt{manipulator}\_\texttt{insert}\_\texttt{peg} in order to respect the y-axis limits. The value is \num{0.88} as D4PG policies are highly overestimated.
    In general, active is better than uniform, methods that exploit policy correlation are better than methods that do not. }
    \label{fig:results_ablations_no_fqe-by-task}
\end{figure}

Similar to \autoref{fig:results_ablations} (second row), \autoref{fig:results_ablations_no_fqe-by-task} shows the contribution of each of the components of the method in case when OPE is not used.
It is clear that the results are significantly worse than when using OPE (\autoref{fig:results_ablations-by-task}) which clearly indicated the benefit of OPE component in \ours{}.
When OPE estimates are not available, the combination of modelling correlated policies as in \textsc{GP} and intelligent policy selection as in \textsc{UCB} gives the best results on average. 
It performs better than the next best method in \num{6} environments, slightly worse in \num{1} and approximately the same in \num{2}.
On average, the \textsc{GP+UCB} strategy is the best when OPE estimates are not available.

\subsection{MPG}
\begin{figure}[tbh!]
    \begin{subfigure}[b]{0.28\textwidth}
        \includegraphics[width=1\linewidth,left]{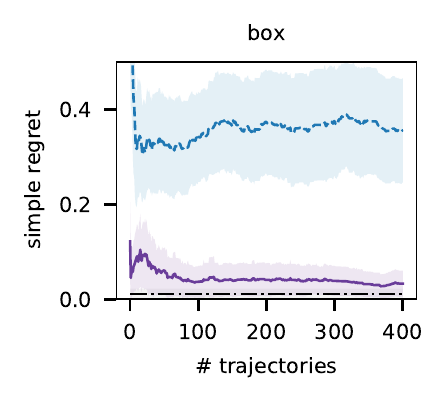}
    \end{subfigure}
    \hspace{0em}
    \begin{subfigure}[b]{0.28\textwidth}
        \includegraphics[width=1\linewidth,left]{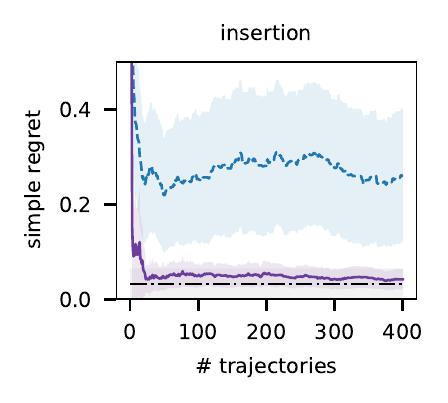}
    \end{subfigure}
    \\
    \begin{subfigure}[b]{0.28\textwidth}
        \includegraphics[width=1\linewidth,left]{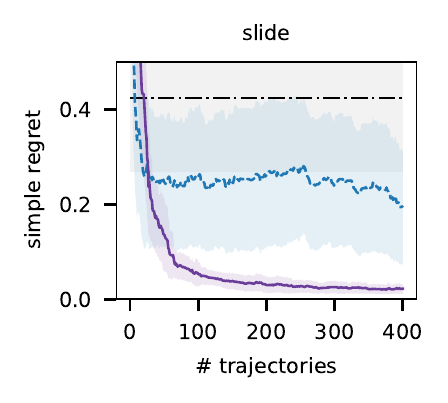}
    \end{subfigure}
    \hspace{0em}
    \begin{subfigure}[b]{0.28\textwidth}
        \includegraphics[width=1.55\linewidth,left]{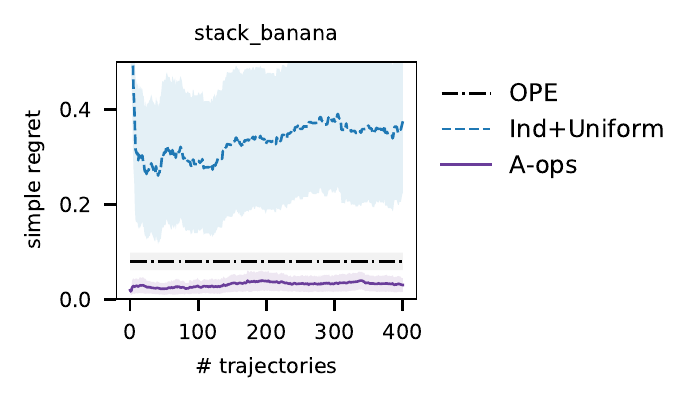}
    \end{subfigure}
    \caption{Comparison of \ours{} approach, offline policy selection with \OPE{}, and online policy selection with \IndUniform{} on each environment in \mpg{} domain. The results are averaged across \num{100} experiments. Shaded areas show standard deviation of the mean. Our method \ours{} quickly surpasses \IndUniform{} baseline. \ours{} achieves the regret close to \num{0} in all environments. }
    \label{fig:results-by-task-mpg}
\end{figure}

Similar to \autoref{fig:results}, \autoref{fig:results-by-task-mpg} shows the comparison between our proposed method \ours{}, completely offline policy selection with OPE and completely online selection with \IndUniform{} on each of \num{4} tasks from \mpg{}. 
OPE performs exceedingly well in \num{3} of \num{4} tasks, getting regret close to zero for \num{2} tasks (see \autoref{fig:stats-mpg} with plot of OPE vs the ground truth). 
Nevertheless we manage to perform about as well or better on all of the tasks: In \num{2} environments, \ours{} only approaches the OPE baseline, but in the other \num{2} environments, \ours{} quickly surpasses OPE.
It makes the most improvement in the \verb|slide| task.
This difference in the performance in different tasks is due to the variability of performance of OPE method and the variance in returns in the online policy executions.
High return variance requires a large number of environment interactions for an online policy evaluation method to provide accurate estimates.
However, the most important observation is that \ours{} achieves a small regret in all environments.

\begin{figure}[tbh!]
    \begin{subfigure}[b]{0.28\textwidth}
        \includegraphics[width=1\linewidth,left]{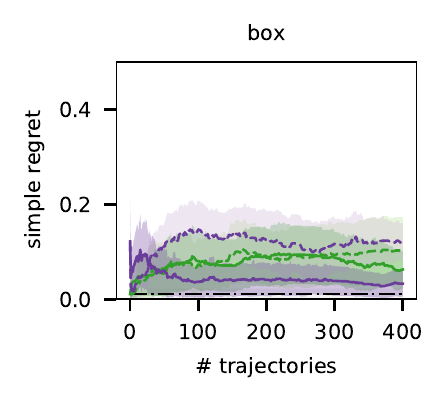}
    \end{subfigure}
    \hspace{0em}
    \begin{subfigure}[b]{0.28\textwidth}
        \includegraphics[width=1\linewidth,left]{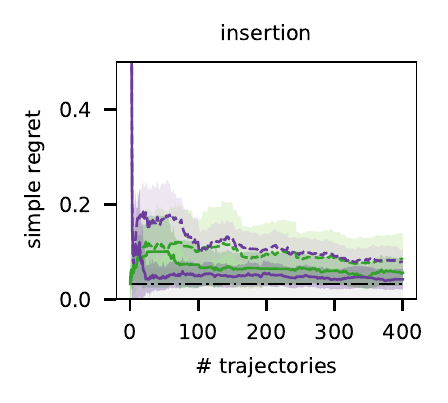}
    \end{subfigure}
    \\
    \begin{subfigure}[b]{0.28\textwidth}
        \includegraphics[width=1\linewidth,left]{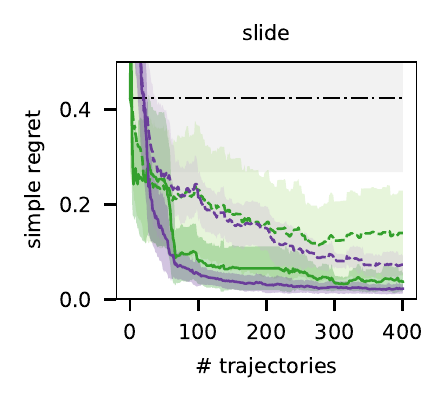}
    \end{subfigure}
    \hspace{0em}
    \begin{subfigure}[b]{0.28\textwidth}
        \includegraphics[width=1.7\linewidth,left]{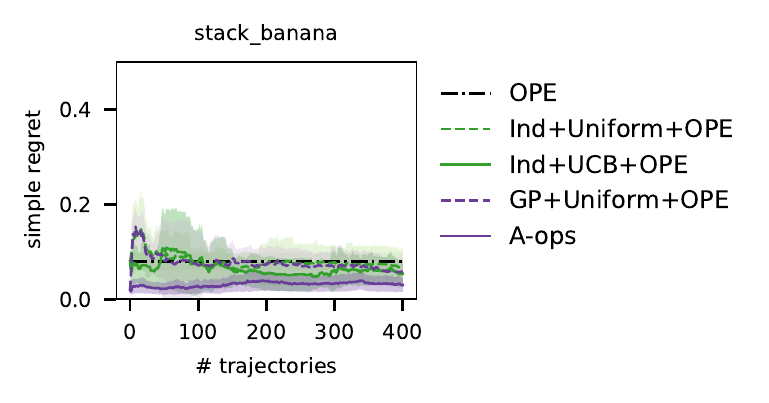}
    \end{subfigure}
    \caption{Ablations. In \num{4} environments of \mpg{} domain, we ablate two components of our approach when using OPE metrics. 1) The choice of policy model: GP (purple) and independent (green). 2) The policy selection strategy: active (dark, solid line) and uniform sampling (bright, dashed line). 
    In general, active is better than uniform, methods that exploit policy correlation are better than methods that do not. }
    \label{fig:ablations-by-task-mpg}
\end{figure}

Similar to \autoref{fig:results_ablations} (first row), \autoref{fig:ablations-by-task-mpg} shows the contribution of each of the components of the method when using OPE. 
Our method \ours{} is preferable in all environments across a wide range of interaction budgets. 
The same observation as before holds: modelling correlated policies as in \textsc{GP} performs better than modelling independent policies as in \textsc{Ind}, active policy selection as in \textsc{UCB} is better than uniform policy selection as in \textsc{Uniform}.

\begin{figure}[tbh!]
    \begin{subfigure}[b]{0.28\textwidth}
        \includegraphics[width=1\linewidth,left]{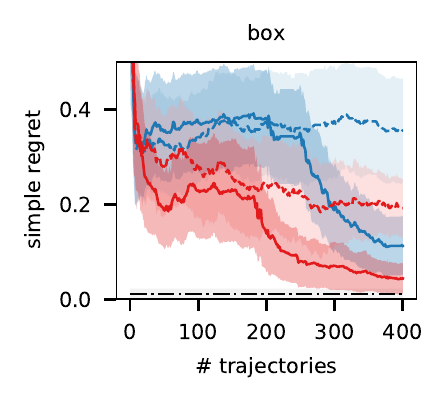}
    \end{subfigure}
    \hspace{0em}
    \begin{subfigure}[b]{0.28\textwidth}
        \includegraphics[width=1\linewidth,left]{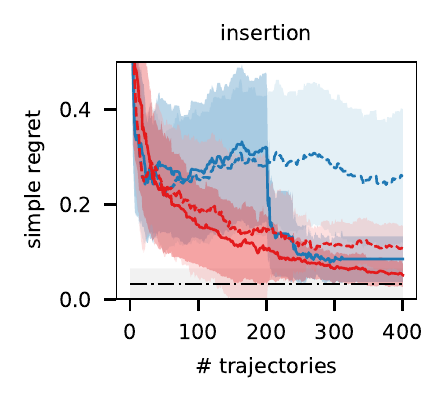}
    \end{subfigure}
    \\
    \begin{subfigure}[b]{0.28\textwidth}
        \includegraphics[width=1\linewidth,left]{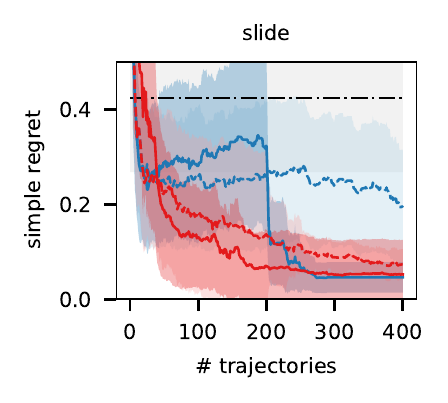}
    \end{subfigure}
    \hspace{0em}
    \begin{subfigure}[b]{0.28\textwidth}
        \includegraphics[width=1.55\linewidth,left]{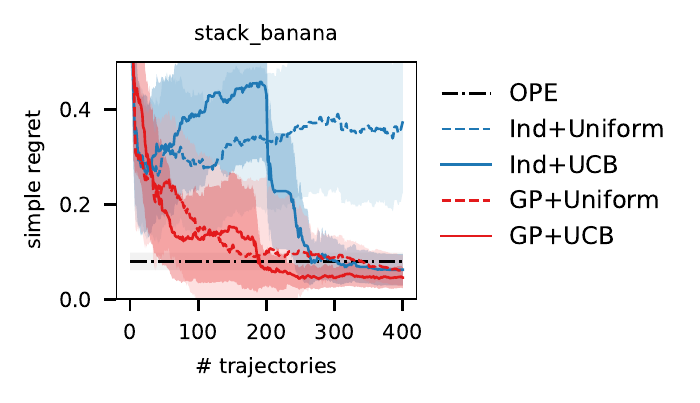}
    \end{subfigure}
    \caption{Ablations. In \num{4} environments of \mpg{} domain, we ablate two components of our approach when NOT using OPE metrics. 1) The choice of policy model: GP (red) and independent (blue). 2) The policy selection strategy: active (dark, solid line) and uniform sampling (bright, dashed line). 
    In general, active is better than uniform, methods that exploit policy correlation are better than methods that do not.}
    \label{fig:ablations-nofqe-by-task-mpg}
\end{figure}

Similar to \autoref{fig:results_ablations} (second row), \autoref{fig:ablations-nofqe-by-task-mpg} shows the contribution of each of the components of the method in case when OPE is not used.
As before, the results are significantly worse than when using OPE (\autoref{fig:ablations-by-task-mpg}).
When OPE estimates are not available, the combination of modelling correlated policies as in \textsc{GP} and intelligent policy selection as in \textsc{UCB} gives the best results in each task. 
Notice the degraded performance of \textsc{Ind+UCB} in the first \num{200} iterations (mostly exploration stage). 
This happens because each policy is treated independently and until each of them is executed (\num{200} policies) the regret is quite high. 
Modelling correlation between the policies as in \textsc{GP} methods helps to alleviate this problem.

\subsection{Atari}

\begin{figure}[t!]
    \begin{subfigure}[b]{0.28\textwidth}
        \includegraphics[width=1\linewidth,left]{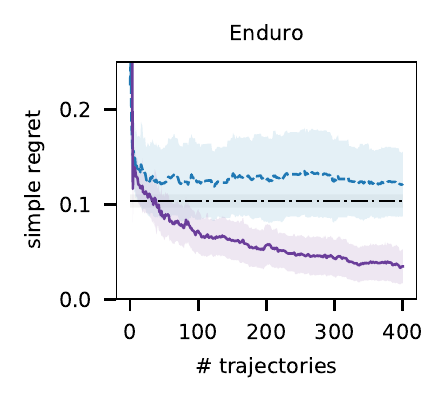}
    \end{subfigure}
    ~
    \begin{subfigure}[b]{0.28\textwidth}
        \includegraphics[width=1\linewidth,left]{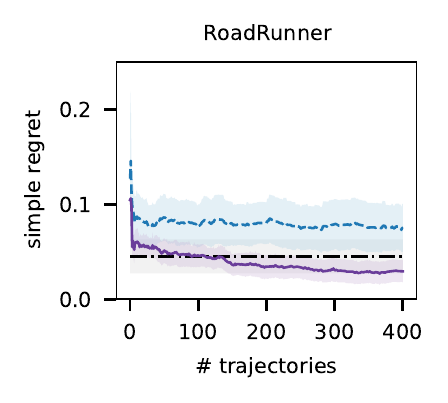}
    \end{subfigure}
    ~
    \begin{subfigure}[b]{0.28\textwidth}
        \includegraphics[width=1.55\linewidth,left]{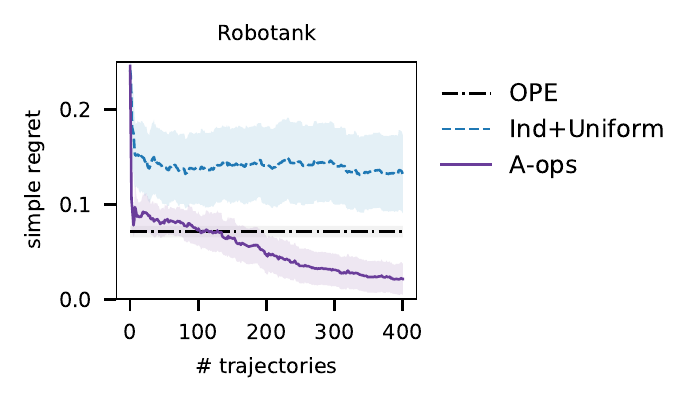}
    \end{subfigure}
    \caption{Comparison of \ours{} approach, offline policy selection with \OPE{}, and online policy selection with \IndUniform{} on each environment in Atari domain. The results are averaged across \num{100} experiments. Shaded areas show standard deviation of the mean. Our method \ours{} quickly surpasses \IndUniform{} baseline. \ours{} achieves the regret close to \num{0} in all environments. }
    \label{fig:atari-results-by-task}
\end{figure}

Similar to \autoref{fig:results}, \autoref{fig:atari-results-by-task} shows the comparison between our proposed method \ours{}, completely offline policy selection with OPE and completely online selection with \IndUniform{} on each of \num{3} Atari games that we consider. 
Due to the variance of the returns of the policies in this domain, it takes a large number of environment interactions for an online policy evaluation method to provide accurate estimates and in all environments offline evaluation is better.
However, \ours{} method outperforms other baselines with only a small amount of environment interactions.

\begin{figure}[t!]
    \begin{subfigure}[b]{0.28\textwidth}
        \includegraphics[width=1\linewidth,left]{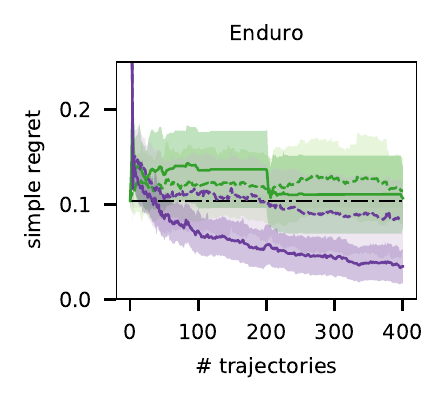}
    \end{subfigure}
    ~
    \begin{subfigure}[b]{0.28\textwidth}
        \includegraphics[width=1\linewidth,left]{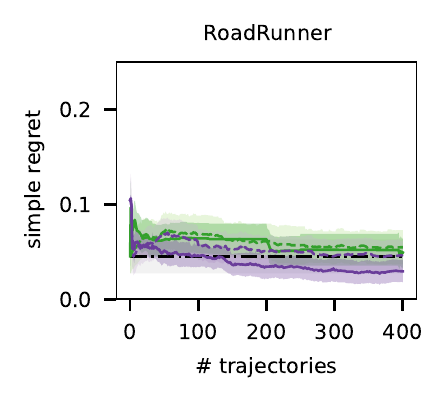}
    \end{subfigure}
    ~
    \begin{subfigure}[b]{0.28\textwidth}
        \includegraphics[width=1.55\linewidth,left]{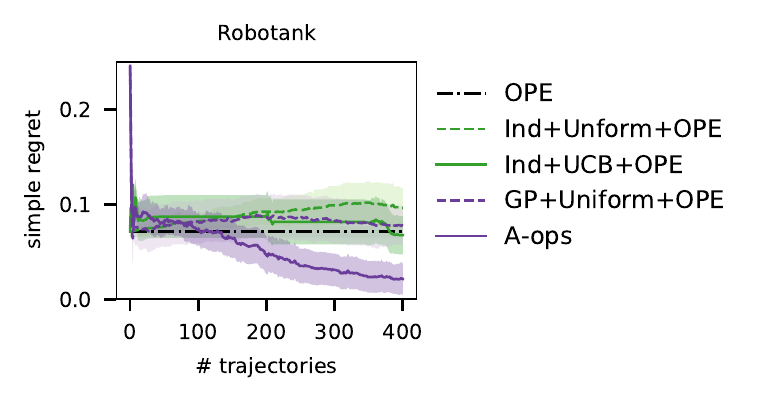}
    \end{subfigure}
    \caption{Ablations. In \num{3} environments of \mpg{} domain, we ablate two components of our approach when using OPE metrics. 1) The choice of policy model: GP (purple) and independent (green). 2) The policy selection strategy: active (dark, solid line) and uniform sampling (bright, dashed line). 
    In general, active is better than uniform, methods that exploit policy correlation are better than methods that do not. }
    \label{fig:atari-fqe-results-by-task}
\end{figure}

Similar to \autoref{fig:results_ablations} (first row), \autoref{fig:atari-fqe-results-by-task} shows the contribution of each of the components of the method. 
Our method \ours{} is preferable in all environments across a wide range of interaction budgets. 
The same observation as before holds: modelling correlated policies as in \textsc{GP} performs better than modelling independent policies as in \textsc{Ind}, active policy selection as in \textsc{UCB} is better than uniform policy selection as in \textsc{Uniform}. 
Notice that the gain of \ours{} is particularly pronounced in this domain. 
We attribute it to the data efficiency of our method which is particularly important given a large diversely performing set of policies.

\begin{figure}[t!]
    \begin{subfigure}[b]{0.28\textwidth}
        \includegraphics[width=1\linewidth,left]{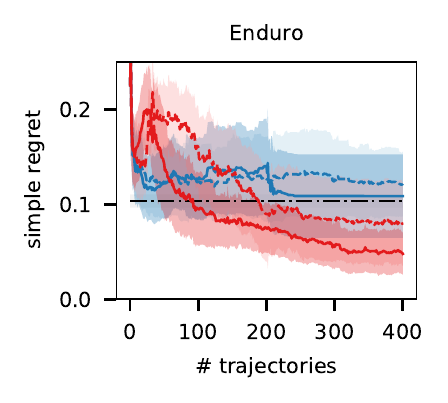}
    \end{subfigure}
    ~
    \begin{subfigure}[b]{0.28\textwidth}
        \includegraphics[width=1\linewidth,left]{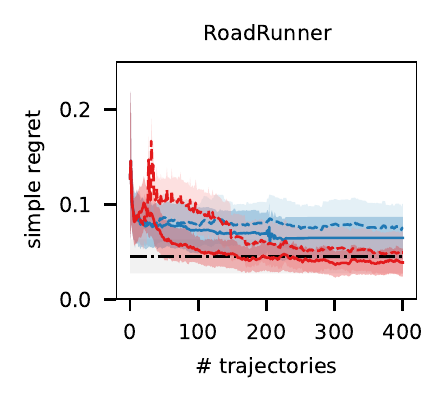}
    \end{subfigure}
    ~
    \begin{subfigure}[b]{0.28\textwidth}
        \includegraphics[width=1.55\linewidth,left]{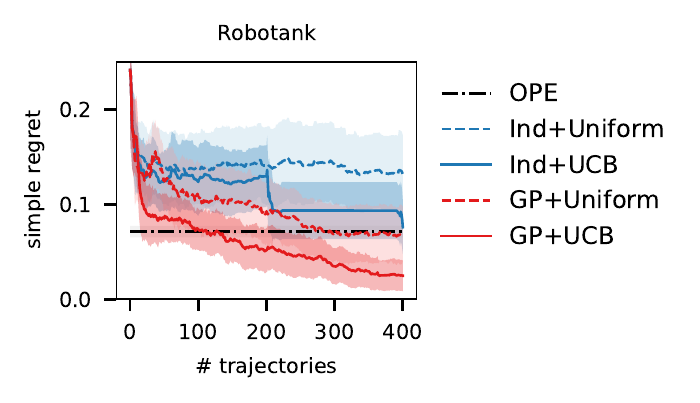}
    \end{subfigure}
    \caption{Ablations. In \num{3} environments of Atari domain, we ablate two components of our approach when NOT using OPE metrics. 1) The choice of policy model: GP (red) and independent (blue). 2) The policy selection strategy: active (dark, solid line) and uniform sampling (bright, dashed line). 
    In general, active is better than uniform, methods that exploit policy correlation are better than methods that do not.}
    \label{fig:atari-nofqeresults-by-task}
\end{figure}

Similar to \autoref{fig:results_ablations} (second row), \autoref{fig:atari-nofqeresults-by-task} shows the contribution of each of the components of the method in case when OPE is not used.
As before, the results are noticeably worse than when using OPE (\autoref{fig:atari-fqe-results-by-task}).
When OPE estimates are not available, the combination of modelling correlated policies as in \textsc{GP} and intelligent policy selection as in \textsc{UCB} gives the best results in each game. 
Notice the degraded performance of \textsc{Ind+UCB} in the first \num{200} iterations (mostly exploration stage). 
This happens because each policy is treated independently and until each of them is executed (\num{200} policies) the regret is quite high. 
Modelling correlation between the policies as in \textsc{GP} methods helps to alleviate this problem and \textsc{GP+UCB} method outperforms all competitors within a small number of trajectories, including OPE method that relies on the offline dataset.

\subsection{Discussion of the results by task}
The performance of the methods varies from task to task. 
We highlight that \ours{} does as well as or better than both offline policy selection (OPE) and online policy selection on \num{9} of \num{9} dm-control suite tasks, in \num{2} of \num{4} MPG tasks and in \num{3} out of \num{3} Atari games. 
The simple regret of \ours{} approaches \num{0} or is very low in \num{7} out of \num{9} dm-control tasks, in \num{4} out of \num{4} MPG tasks and in \num{3} out of \num{3} Atari tasks. 
Thus, in cases when \ours{} regret is not very low, it is still much better than that of the alternative methods, and when \ours{} does not outperform some of the methods, its regret remains very low. 
As a result, despite variability in different tasks, \ours{} is a very reliable method for policy selection.
Another reason for variability across different tasks is the fact that we do not address the question of when to stop active policy selection. 
In some cases, the OPE estimate is so good that there is no need for extra online data (for example, as in some of the \mpg{} environments) and in some cases due to the variance of the returns and low correlation of OPE scores with policy returns, even more extensive data collection could be needed. 
Future work can address this question and derive some recommendations on when to stop policy querying based on statistics in collected data: policy variance, OPE correctness, gap between the policies etc.

In general, the main factors that affect the viability in the \ours{} performance are 1) the distribution of OPE estimates vs ground truth, 2) the variance of returns of a single policy, 3) variance in the returns across available policies. 
As this information is not available prior to conducting the experiments, it is important that our method \ours{} performs reliably even in the challenging conditions.

Note that the efficiency of \ours{} does not depend on the difficulty of training a policy. 
For example, training policies from visual input such as in \mpg{} and Atari domains as well as some challenging tasks in \dmcontrol{} including \verb|humanoid_run|, \verb|manipulator_insert_ball|, \verb|manipulator_insert_peg| require thousands of trajectories to be solved online, but it takes only around \num{100} of trajectories to identify a good policy.

\section{Qualitative results by task}

We illustrate the behaviour of \ours{} by showing its estimates and selection in one experiment in each environment in \autoref{fig:qualitative-all} in the same way as in \autoref{fig:qualitative}.
Each experiment is produced with a fixed random seed and the examples are not curated.
Thus, they include both the examples of success in identification of the best strategy (for example, among others in \verb|cartpole_swingup|, \verb|insertion|, and \verb|Robotank|) and failure (as in, for example, \verb|manipulator_insert_ball|).
For visualisation purposes, we use only \num{20} policies which are ordered by their ground truth returns (not observed by the method) and perfect predictions would be aligned along a diagonal line. 
We show the initial \emph{OPE} values, current \emph{selection} of the algorithm, \emph{current} observation, past \emph{observations} and the algorithm \emph{prediction} with standard deviation for steps \num{1}, \num{10}, \num{50}, and \num{100}.

We can see that the initial selection is dominated by OPE prediction.
Policy selection by OPE  may be quite far from the best policy (for example, as in \verb|manipulator_insert_ball|, \verb|manipulator_insert_peg|, \verb|insertion|, \verb|slide|, \verb|Enduro|), but it may also be very precise leaving no space for further improvement (for example, in \verb|cheetah_run|, \verb|walker_walk|).
Notice that \ours{} prediction may be different from observations because our GP model assumes policies are correlated as defined by the policy kernel. 
An obvious example of this occurs at step $=1$ when there are no online trajectories and only OPE scores are available.
Intuitively, OPE scores are ``smoothed'' by the kernel when making a prediction. 
In practice it means that \ours{} predictions at the very few first interactions might be worse than OPE scores if OPE is highly accurate, but they rapidly improve as more online evaluations are collected and OPE is quickly outperformed.
If we wanted to match the predictions of OPE at the very first iteration, we could perform the GP optimisation for many iterations and without any hyperparameter priors, but this would essentially result in overfitting and would not have any positive effect on the performance of our method with a moderate amount of online observations. 

With more online observations, the prediction by \ours{} improves. 
We clearly see that the selected policies are not uniformly sampled and are biased toward more promising samples (there are usually more pink points towards the right of the figure).
Furthermore, the predictions are interdependent, we can see that the predictions of the return of a particular policy might change even without any new observations of it.

In some experiments the visualisation shows that the predictions by the model are dominated by the variance of the episodic returns at some point during the training. 
For example, it happens in \verb|cheetah_run|, \verb|fish_swim|, \verb|manipulator| tasks, and \verb|insertion| when at some step during the trajectory acquisition, all predictions look rather ``flat'' as GP does not assign very different predictions to them given a small amount of observations, but with high variance.

In some experiments we do observe consistent overestimation (for example, in \verb|cheetah_run|, \verb|humanoid_run|, and \verb|walker| tasks) and underestimation (for example, in \verb|Enduro| and \verb|RoadRunner|) by OPE methods.
However, \ours{} method still performs well and is able to select a good policy for recommendation.
It can recover quickly from this bias thanks to its kernel that brings policies with similar actions together: When a return is observed for one policy, the predictions for several related policies are adjusted automatically.

\begin{figure}[tbph!]
    \includegraphics[width=1\linewidth]{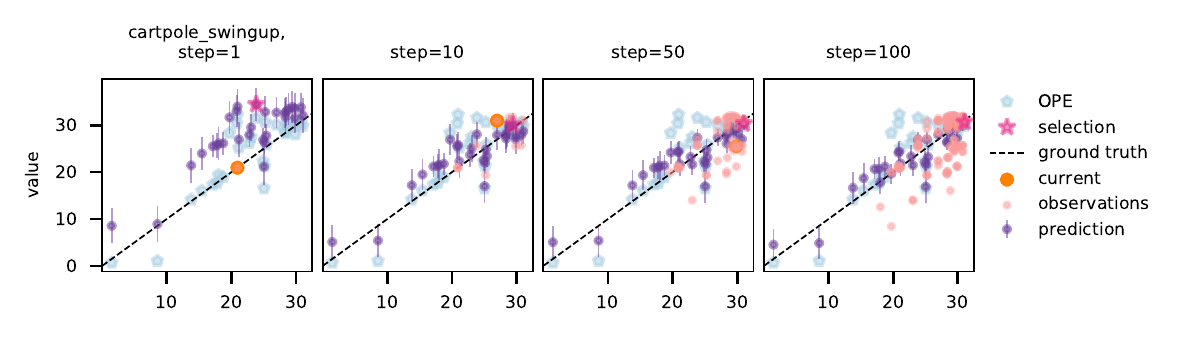}
    \hspace{0em}
    \includegraphics[width=1\linewidth]{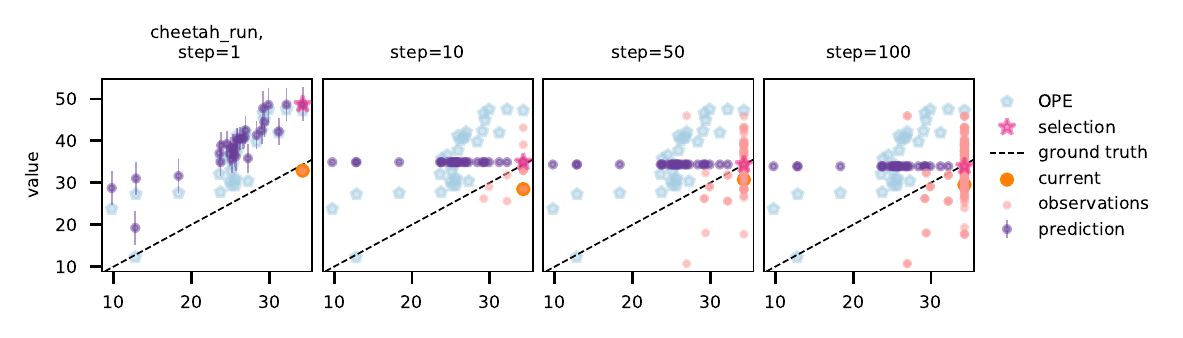}
    \hspace{0em}
    \includegraphics[width=1\linewidth]{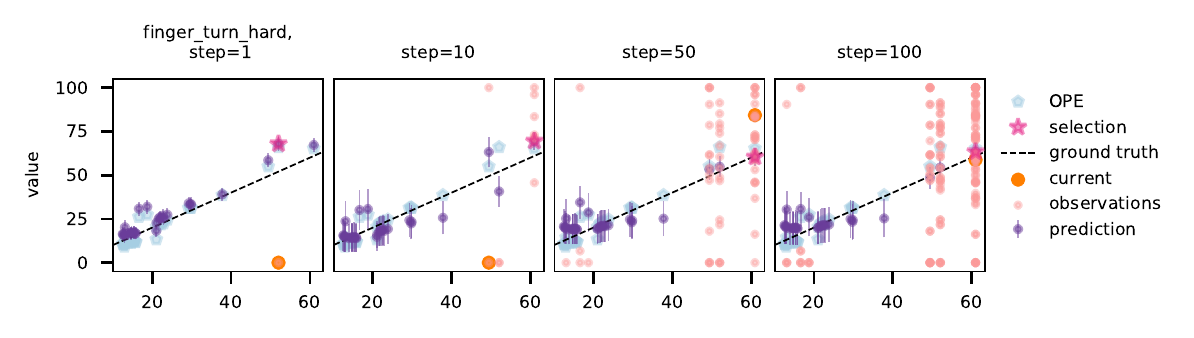}
    \hspace{0em}
    \includegraphics[width=1\linewidth]{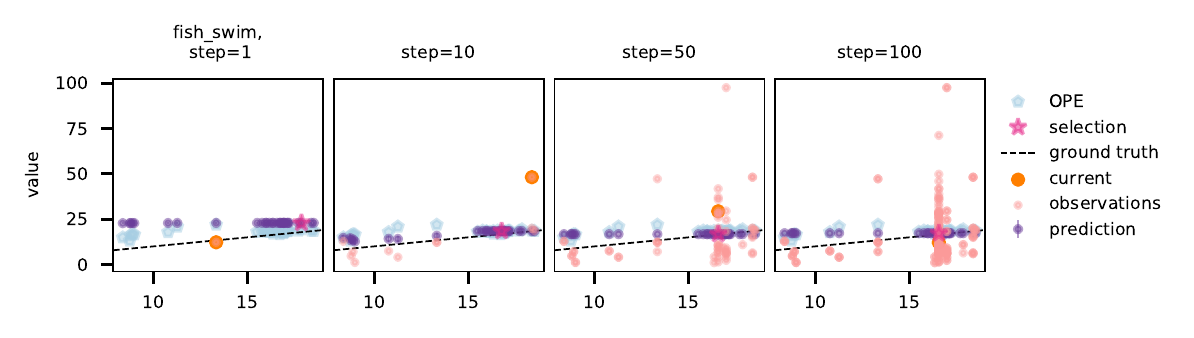}
    \hspace{0em}
    \includegraphics[width=1\linewidth]{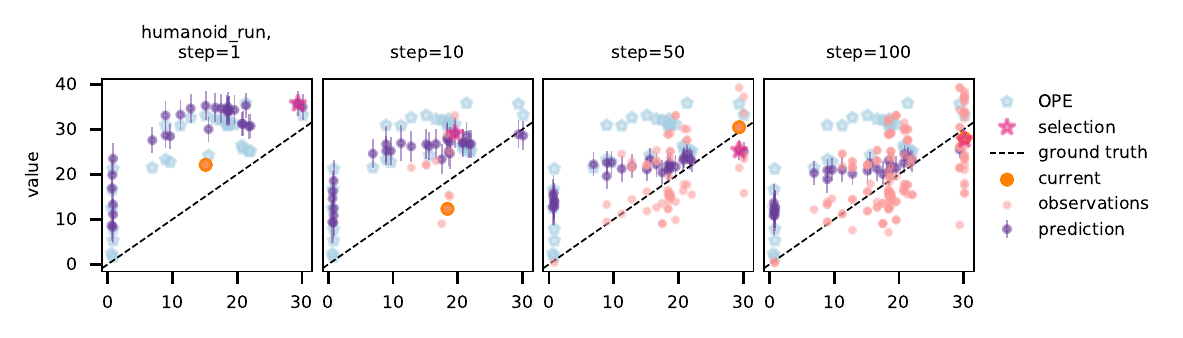}
\end{figure}
\begin{figure}[tbph!]
\ContinuedFloat
    \includegraphics[width=1\linewidth]{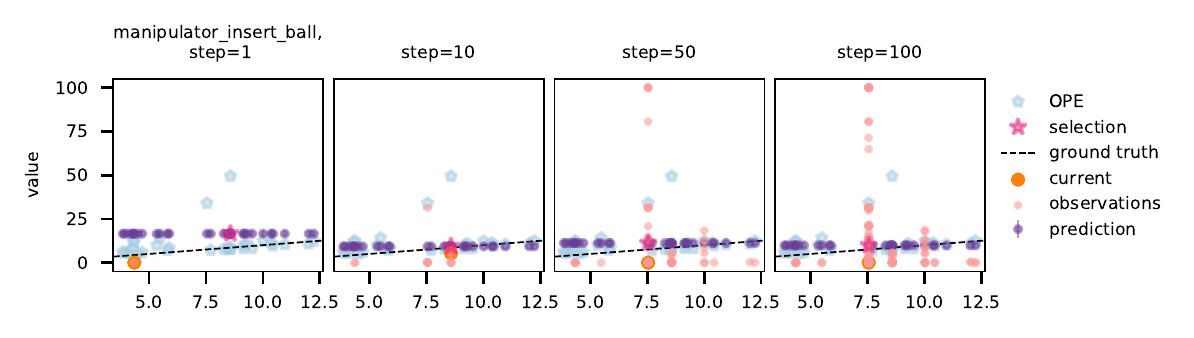}
    \includegraphics[width=1\linewidth]{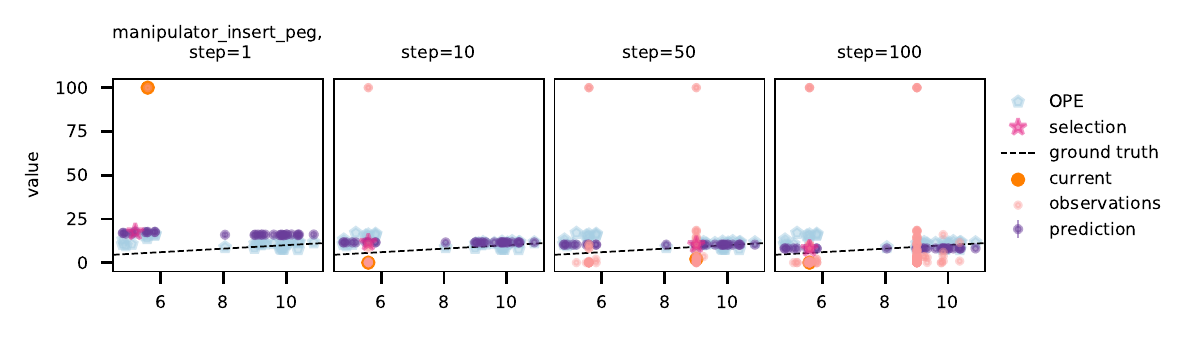}
    \includegraphics[width=1\linewidth]{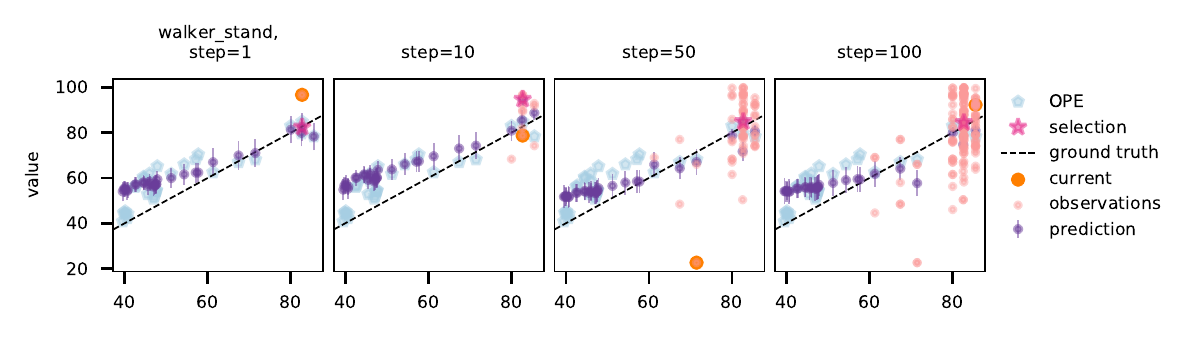}
    \includegraphics[width=1\linewidth]{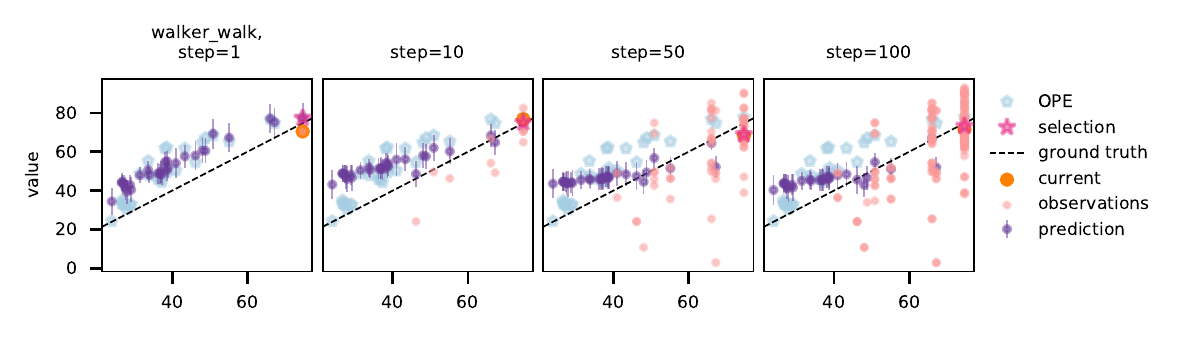}
    \includegraphics[width=1\linewidth]{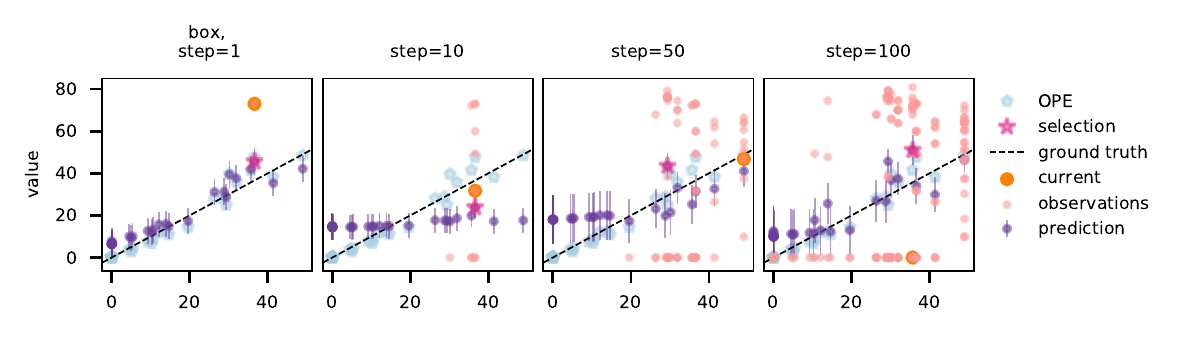}
\end{figure}
\begin{figure}[tbph!]
\ContinuedFloat        
    \includegraphics[width=1\linewidth]{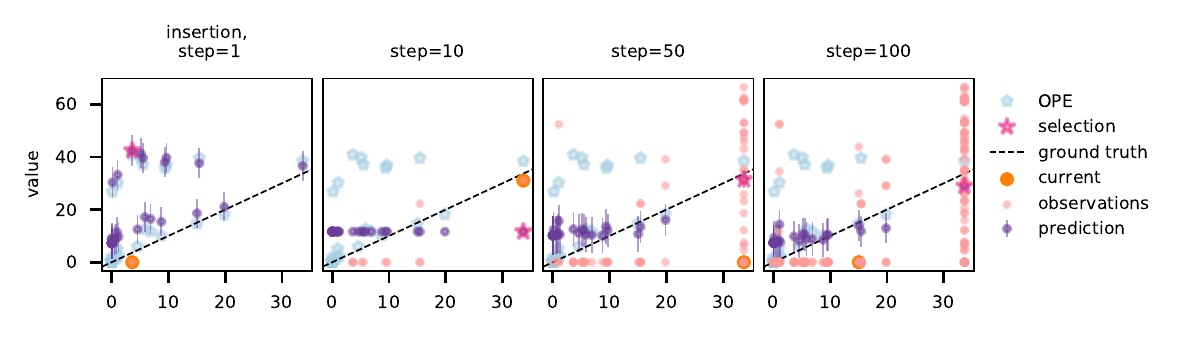}
    \includegraphics[width=1\linewidth]{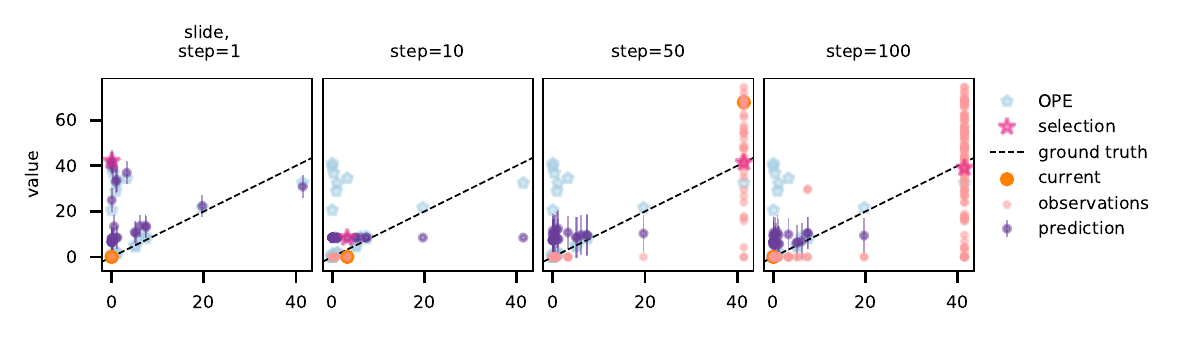}
    \includegraphics[width=1\linewidth]{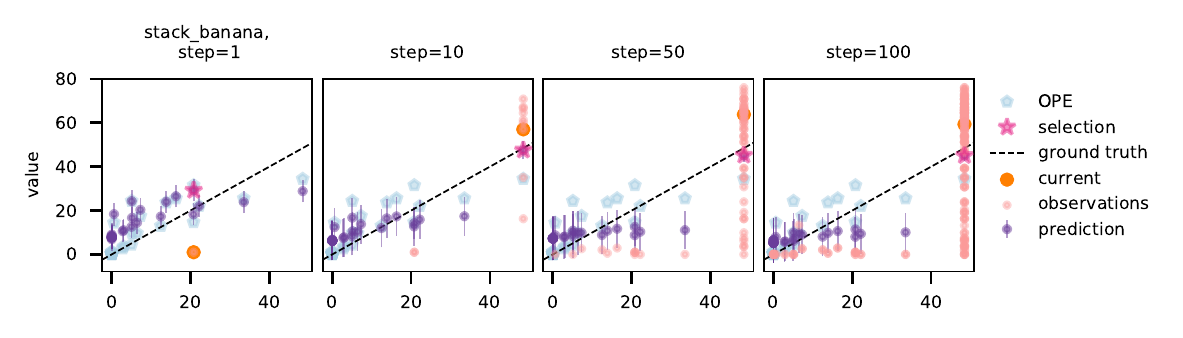}
    \includegraphics[width=1\linewidth]{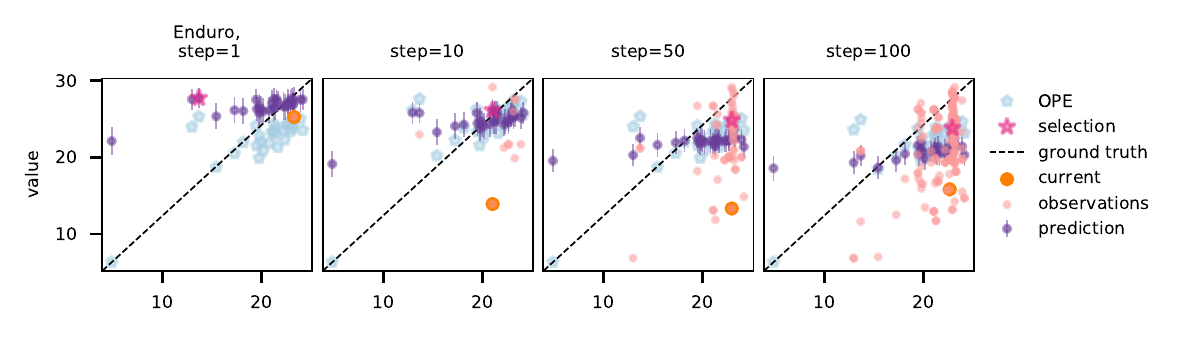}
    \includegraphics[width=1\linewidth]{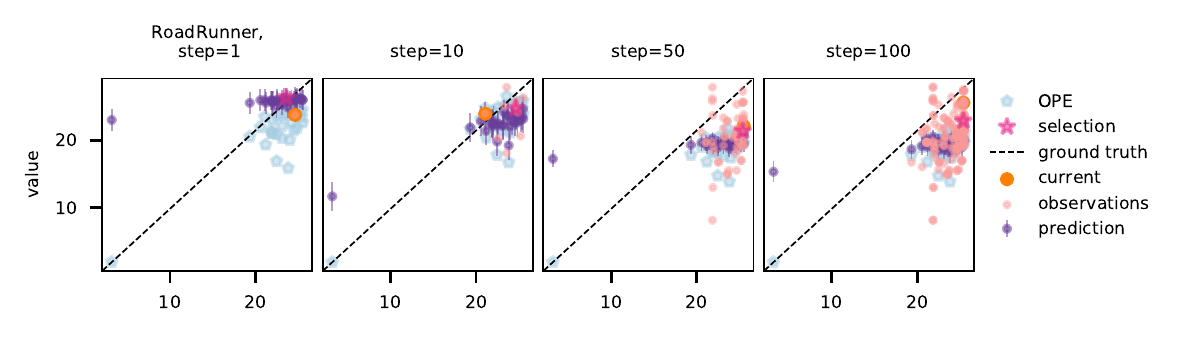}
\end{figure}
\begin{figure}[tbph!]
\ContinuedFloat
    \includegraphics[width=1\linewidth]{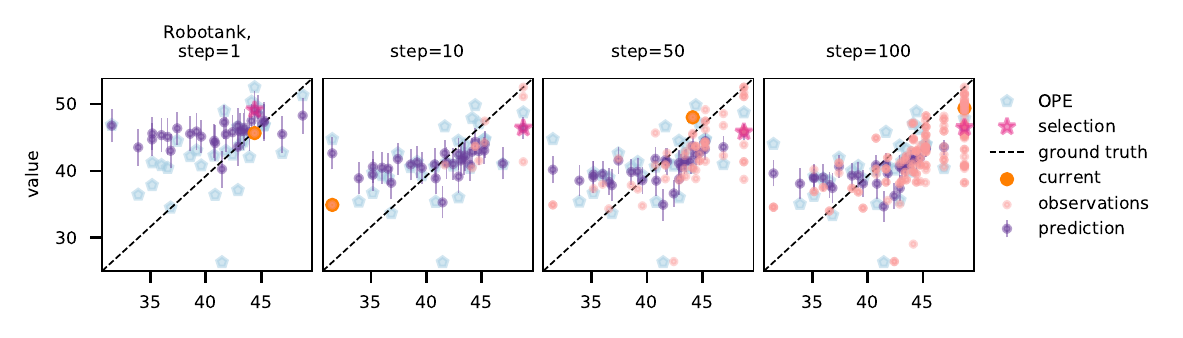}
    \caption{Visualisation of the qualitative performance of \ours{} algorithm on all the environments from \dmcontrol{}, \mpg{} and Atari domains with \num{20} randomly selected policies (ordered by ground truth only for the illustration purpose). From left to right we show the state at \num{1}, \num{10}, \num{50}, \num{100} algorithm iterations. The initial OPE scores are shown in light blue, the policy selected by \ours{} is highlighted with a magenta star, the current observation by the algorithm is shown in orange circle, past observations are in pink and the prediction with its standard deviation is in purple.}
    \label{fig:qualitative-all}
\end{figure}
\section{Additional experiments}
\label{sec:additional_experiments}

Finally, in this section we present additional experiments that demonstrate various aspects of the problem and our proposed method.

\subsection{Ranking as performance measure}

\begin{figure}[t!]
    \begin{subfigure}[b]{0.44\textwidth}
        \includegraphics[width=1\linewidth,left]{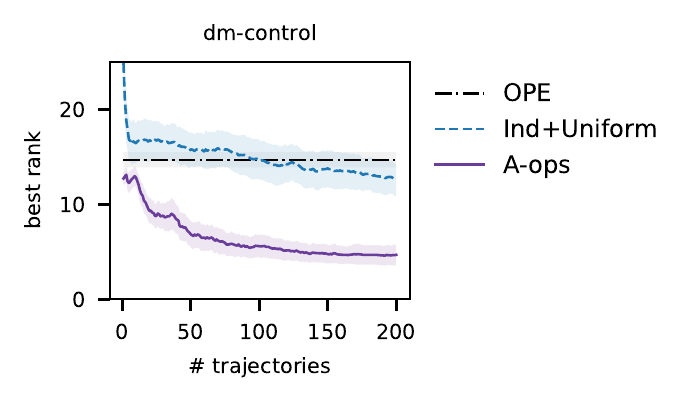}
    \end{subfigure}
    \\
    \begin{subfigure}[b]{0.48\textwidth}
        \includegraphics[width=1\linewidth,left]{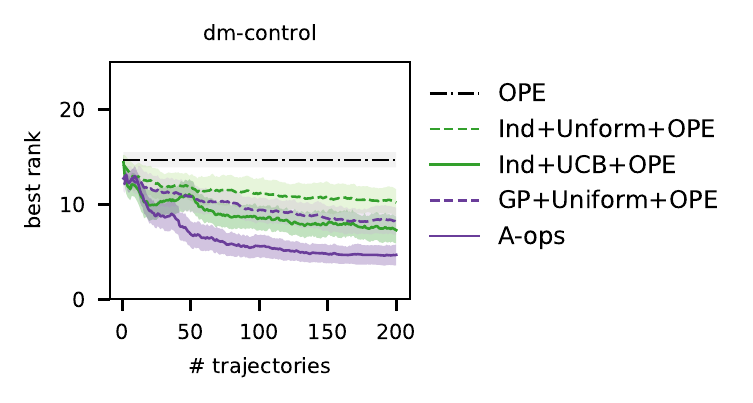}
    \end{subfigure}
    \hspace{0em}
    \begin{subfigure}[b]{0.44\textwidth}
        \includegraphics[width=1\linewidth,left]{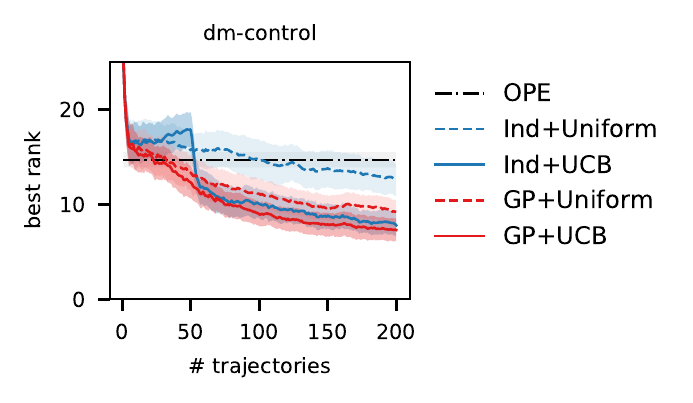}
    \end{subfigure}
    \caption{Ranking performance metric. Aggregated across \num{9} environments of the \dmcontrol{} domain, we show the performance of the policy selection methods in terms of the rank of the selected policy.
    The same observations hold: \ours{} is the best strategy, active is better than uniform, methods that exploit policy correlation are better than methods that do not.}
    \label{fig:ranking}
\end{figure}

We quantified the performance of the policy selection algorithms based on the simple regret so far. 
Other performance metrics could be used, for example, the rank of the selected policy. 
The best rank is \num{1} that means that the best policy was found and the worst rank is $K$, where $K$ is the number of candidate policies.
\autoref{fig:ranking} shows results similar to \autoref{fig:results} and \autoref{fig:results_ablations} for the ranking metric. 
The same observations still hold.

\subsection{Cumulative and minimum regret}

\begin{figure}[t!]
    \includegraphics{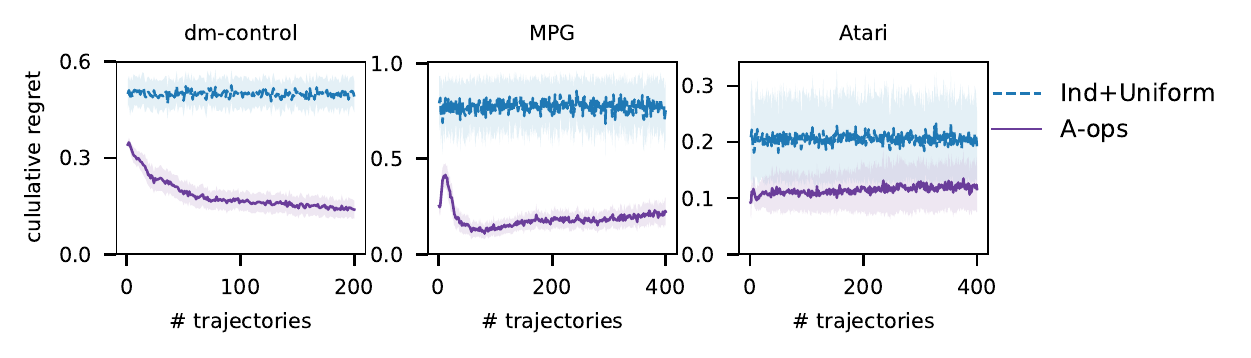}
\caption{Cumulative regret of \ours{} is lower than that of online policy selection with \IndUniform{}.}
    \label{fig:culumative}
\end{figure}

While the goal of \ours{} is to minimise the simple regret and not cumulative regret, it still avoids repetitive execution of most poorly performing policies when employing UCB criterion. 
As a result, \ours{} achieves lower cumulative regret than strategies based on uniform sampling as manifested by \autoref{fig:culumative}. 
Interestingly, towards the end of the experiment the cumulative regret starts growing slightly, that indicates that \ours{} goes back to the exploration phase again once a promising policy is identified.

We also analyse how often during the policy selection procedure \ours{} executes very bad policies. 
For this we compute the percentage of \ours{} (and alternative methods) iterations where the worst policy or one of the \num{10}\% worst policies is executed. 
Our findings are summarized in \autoref{tab:min} for the three most representative algorithms. 
\ours{} strategy ends up querying the worst-performing policies significantly less frequently than other algorithms despite not being optimized for it directly. 
A potential explanation is that the policy kernel allows us to infer which policies are poorly performing in a very data efficient manner.

\begin{table}[t!]
    \caption{The percentage of time that the worst and one of the \num{10}\% worst policies is executed during the policy selection procedure by different algorithms including \ours{}.}
    \vspace{2mm}
    \centering
    {\small
    \begin{tabular}{l|r|r|r}
    method & \dmcontrol{} & \mpg{} & Atari \\
    \midrule
    & \multicolumn{3}{c}{the worst policy} \\
    \midrule
    \IndUniformOPE{} & \num{2}\% & \num{0.52}\% & \num{0.52}\% \\
    \IndUCBOPE{} & \num{1.2}\% & \num{0.26}\% & \num{0.31}\% \\
    \ours{} & \num{0.14}\% & \num{0.007}\%  & \num{0.01}\% \\
    \midrule
     & \multicolumn{3}{c}{\num{10}\% worst policies} \\
    \midrule
    \IndUniformOPE{} & \num{12}\% & \num{10.6}\% & \num{10.5}\% \\
    \IndUCBOPE{} & \num{7.7}\% & \num{5.4}\% & \num{9}\% \\
    \ours{} & \num{1.3}\% & \num{0.17}\%  & \num{1.4}\% \\
    \end{tabular}
    }
    \label{tab:min}
\end{table}

\subsection{Varying amount of OPE estimated}

\begin{figure}
     \centering
     \includegraphics{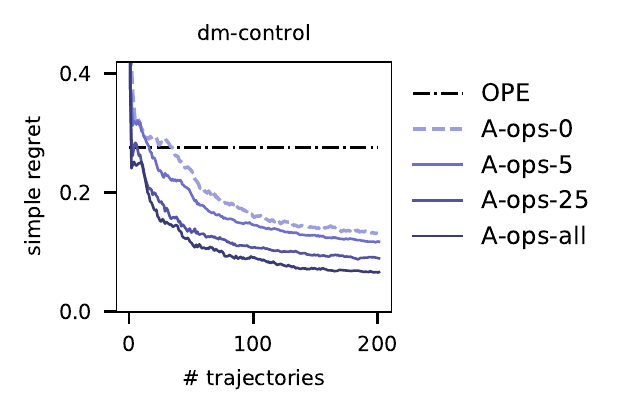}
     \caption{The performance of \ours{} on \dmcontrol{} with varying number of FQE}
     \label{fig:ope-vary_n}
\end{figure}

Computing OPE estimates (like FQE or model-based OPE) can sometimes be expensive.
\autoref{fig:ope-vary_n} depicts the performance of \ours{} by providing OPE estimates for a random subset of $k \in [0, 50]$ policies in \dmcontrol{} environment.
Increasing the number of OPE observations consistently improves the results. 
This result demonstrates that our method \ours{} can seamlessly incorporate all the available information and benefit from the increasing number of initial observations.

\subsection{Effect of the selections strategies}

\begin{figure}[t!]
    \centering
    \includegraphics[width=0.5\linewidth]{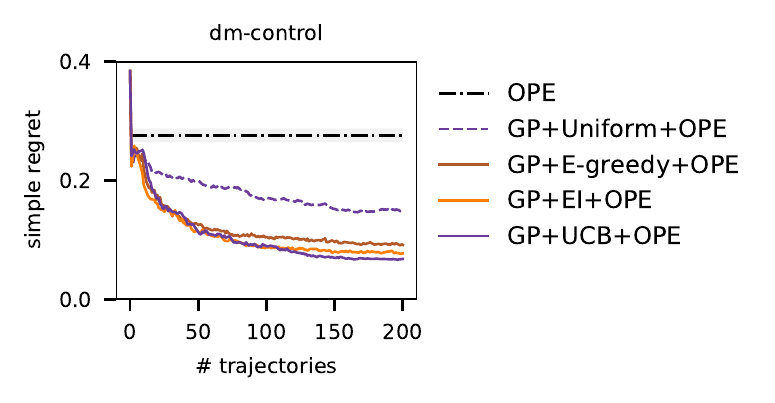}
    \includegraphics[width=0.42\linewidth]{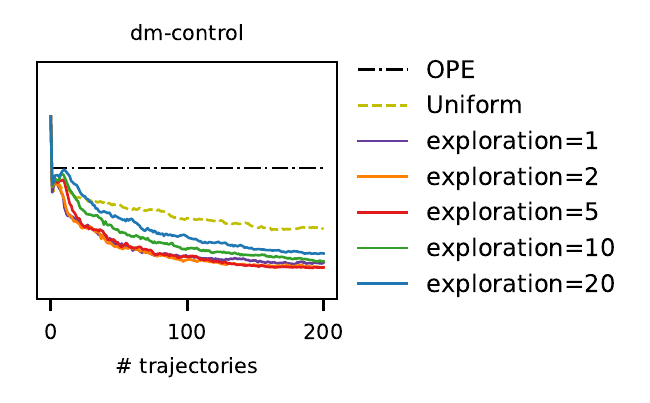}
    \caption{Left: The performance of \ours{} where different values of exploration coefficients are used. The difference between different coefficients is small for a wide range of values and using UCB criterion with any exploration value is better than doing uniform sampling.
    Right: The performance of various selection strategies in conjunction with \textsc{GP+OPE} in \num{9} environments of \dmcontrol{} domain. The difference between uniform and active selection strategies is significant, while the difference between various active selection strategies is small. For better visualization, we do not show the standard deviation, which is comparable to the \ours{} result in \autoref{fig:results_ablations}. }
    \label{fig:different-selection-strategies}
\end{figure}

In the next experiment we study the effect of the choice of Bayesian Optimisation method on the performance of the \ours{} method.
First, we investigate the influence of the exploration coefficient on the performance. 
It turns out that the performance of the \ours{} method is not very sensitive to the choice of exploration coefficient value. 
Our ablation study with various values in \autoref{fig:different-selection-strategies}, right shows that there is no difference in performance for the values between \num{1} and \num{7} and the performance slightly degrades, but still stays better than for the baselines, for the values \num{7}-\num{20}.

Furthermore, we note that the selected best strategy \ours{} as \GPUCBOPE{} is just one realisation of the active policy selection strategy. 
We chose UCB as our strategy as it demonstrated promising results across different environments and domains. 
However, other active selection strategies could be applied as well, for example, epsilon-greedy or expected improvement (EI). 
In \autoref{fig:different-selection-strategies}, left we show the performance of various selection strategies in one domain. 
The choice of a BO strategy makes a considerably smaller difference than the increase of regret when using uniform sampling (\textsc{GP+Uniform+OPE}).

\begin{figure}[t!]
     \centering
     \includegraphics[width=0.3\linewidth]{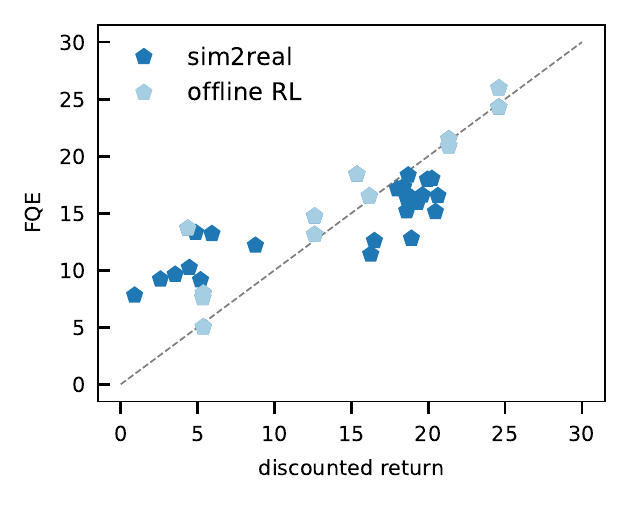}
     \caption{FQE results for sim2real (light blue) and offline RL (dark blue) policies. The correlation of FQE with the ground truth returns is very high.}
     \label{fig:robotics-ope}
\end{figure}

\section{Real-world robotic experiments}
\label{robot-FQE}

For the real-world experiment with robotics data we performed OPE with FQE for the policies trained with sim2real and offline RL. 
To the best of our knowledge, this is the first time this method was applied to this domain. 
We present the results in ~\autoref{fig:robotics-ope} where the ground truth discounted return is plotted against the computed FQE values. 
The policies that were trained with sim2real approach are indicated with light blue and the policies trained on real world data are indicated in dark blue. 
We notice that all predicted FQE values are well correlated with the ground truth returns, the correlation coefficient is \num{0.842}. 
There is no significant difference between the results of offlince policy evaluation with FQE for offline RL and sim2real policies and to the best of our knowledge this is the first demonstration of FQE for sim2real.

\end{document}